\RequirePackage{fix-cm}
\documentclass[twocolumn]{svjour3}          
\smartqed  
\usepackage{graphicx}
\usepackage{natbib}
\usepackage{multirow}
\usepackage{amsmath}
\usepackage{amssymb}
\usepackage{bbm}
\usepackage{bm}
\usepackage{booktabs}
\usepackage{array} 
\usepackage{blindtext}
\usepackage{comment}
\usepackage{tabularx}

\usepackage{array}
\newcolumntype{P}[1]{>{\centering\arraybackslash}p{#1}}

\newcommand{\bfsection}[1]{\vspace*{0.5mm}\noindent\textbf{#1.}}

\usepackage{algorithm}
\usepackage{algorithmic}

\usepackage{subcaption}

\usepackage{xspace}
\usepackage{appendix}
\makeatletter
\DeclareRobustCommand\onedot{\futurelet\@let@token\@onedot}
\def\@onedot{\ifx\@let@token.\else.\null\fi\xspace}

\def\ie{\emph{i.e}\onedot} 
 
 \def\vs{\emph{vs}\onedot}

\makeatother

\usepackage{color, colortbl}
\usepackage[table]{xcolor}
\definecolor{remark}{rgb}{1,.5,0} 
\definecolor{citepcolor}{rgb}{0,0.443,0.737} 
\definecolor{linkcolor}{rgb}{0.956,0.298,0.235} 
\definecolor{gray}{gray}{0.95}
\definecolor{cyan}{rgb}{0.831,0.901,0.945}
\definecolor{mygray}{gray}{.9}

\usepackage{amsmath,amsfonts,bm}

\def\eqref#1{equation~\ref{#1}}

\def\1{\bm{1}}

\def\vg{{\bm{g}}}

\def\vs{{\bm{s}}}

\DeclareMathAlphabet{\mathsfit}{\encodingdefault}{\sfdefault}{m}{sl}
\SetMathAlphabet{\mathsfit}{bold}{\encodingdefault}{\sfdefault}{bx}{n}

\usepackage{multirow}
\usepackage{booktabs}
\usepackage{enumitem}
\usepackage{graphicx}
\usepackage{amsmath}
\usepackage{amssymb}
\usepackage{gensymb}
\usepackage{mathtools}
\usepackage{bm}

\def\vg{{\bm{g}}}

\def\vs{{\bm{s}}}

\usepackage{enumitem}
\setlist[itemize]{topsep={0pt},partopsep={0pt}}

\def\ie{\emph{i.e., }}

\def\vs{\emph{vs. }}

\def\name{\emph{Instant3D}}

\def\tab{Table}
\def\fig{Figure}
\def\sec{Section}
\def\equ{Equation}

\usepackage{epsfig}
\usepackage{subcaption}

\usepackage{url}
\usepackage[misc]{ifsym}

\definecolor{lightgreen}{HTML}{D8ECD1}
\definecolor{edit}{HTML}{C63678}

\usepackage{pifont}

\usepackage{color, colortbl}
\definecolor{citecolor}{HTML}{0071bc}
\definecolor{tabhighlight}{HTML}{e5e5e5}
\usepackage[colorlinks,citecolor=citecolor]{hyperref}

\makeatletter
\renewcommand\paragraph{
  \@startsection{paragraph} 
  {4} 
  {\z@} 
  {.5em \@plus1ex \@minus.2ex} 
  {-.5em} 
  {\normalfont\normalsize\bfseries} 
}
\makeatother
\begin{document}
\sloppy

\title{\name: Instant Text-to-3D Generation 
}


\author{Ming Li \and
        Pan Zhou \and
        Jia-Wei Liu \and
        Jussi Keppo \and
        Min Lin \and
        Shuicheng Yan \and
        Xiangyu Xu\textsuperscript{ \Letter}
}


\institute{Ming Li \at
              Institute of Data Science, National University of Singapore and Sea AI Lab \\
              \email{ming.li@u.nus.edu}
           \and
           Pan Zhou \at
              School of Computing and Information Systems, Singapore Management University \\
              \email{panzhou@smu.edu.sg}
           \and
           Jia-Wei Liu \at
              Show Lab, National University of Singapore \\
              \email{jiawei.liu@u.nus.edu}
           \and
            Jussi Keppo \at
              Business School, National University of Singapore \\
              \email{keppo@nus.edu.sg}
           \and
           Min Lin \at
              Sea AI Lab, Singapore \\
              \email{linmin@sea.com}
            \and
           Shuicheng Yan \at
            Skywork AI, Singapore   \\
              \email{shuicheng.yan@gmail.com}
           \and
           Xiangyu Xu (Project Lead, Corresponding Author)\at
           Xi'an Jiaotong University, China \\
           \email{xuxiangyu2014@gmail.com}
}

\date{Received: date / Accepted: date}

\maketitle

\begin{abstract}
\textcolor{black}{
Text-to-3D generation has attracted much attention from the computer vision community. Existing methods mainly optimize a neural field from scratch for each text prompt, relying on heavy and repetitive training cost which impedes their practical deployment.
In this paper, we propose a novel framework for fast text-to-3D generation, dubbed {\name}.
Once trained, {\name} is able to create a 3D object for an unseen text prompt in less than one second with a single run of a feedforward network. 
We achieve this remarkable speed by devising a new network that directly constructs a 3D triplane from a text prompt.
The core innovation of our {\name} lies in our exploration of strategies to effectively inject text conditions into the network. In particular, we propose to combine three key mechanisms: cross-attention, style injection, and token-to-plane transformation, which collectively ensure precise alignment of the output with the input text.
Furthermore, we propose a simple yet effective activation function, the scaled-sigmoid, to replace the original sigmoid function, which speeds up the training convergence by more than ten times.
Finally, to address the Janus (multi-head) problem in 3D generation, we propose an adaptive Perp-Neg algorithm that can dynamically adjust its concept negation scales according to the severity of the Janus problem during training, effectively reducing the multi-head effect.
Extensive experiments on a wide variety of benchmark datasets demonstrate that the proposed algorithm performs favorably against the state-of-the-art methods both qualitatively and quantitatively, while achieving significantly better efficiency.
The code, data, and models are available at \url{https://github.com/ming1993li/Instant3DCodes}.}

\keywords{Text-to-3D Generation \and Large-Scale Generative Models \and Neural Radiance Fields}
\end{abstract}

\section{Introduction}

\begin{figure}[!t]
    \centering
    \includegraphics[width=0.99\linewidth]{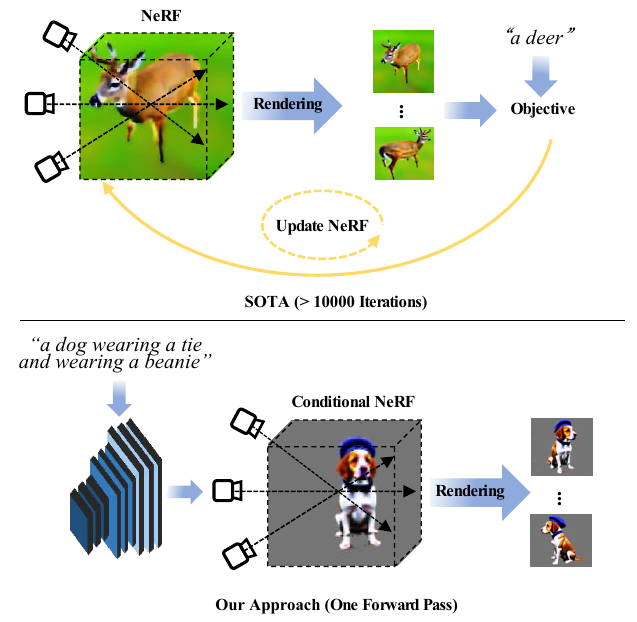}
  \caption{
  Comparison of existing SOTA methods with our proposal. Existing methods optimize a randomly initialized NeRF from scratch for each text prompt, which usually requires more than 10,000 iterations, taking \textbf{hours} to converge. In contrast, our approach is designed to learn a \text{text-conditioned} NeRF, which takes much less training computational cost and has strong generalization ability on new prompts. It is able to generate a conditional 3D representation (triplane) of a 3D object for an unseen text prompt in one single pass of a feedforward network, taking about $\textbf{25ms}$.
  }
  \label{fig:teaser}
\end{figure}

Text-guided 3D content generation has immense potential across diverse applications, such as film production, animation, and virtual reality. 
Recently, this field has witnessed rapid and notable progress, as exemplified by the pioneering works including DreamFusion~\citep{poole2022dreamfusion}, Latent-NeRF~\citep{metzer2022latent}, SJC~\citep{Jain_2022_CVPR}, and ProlificDreamer~\citep{wang2023prolificdreamer}.
These works typically represent a target object with a randomly initialized Neural Radiance Field (NeRF)~\citep{mildenhall2020nerf} and then optimize the NeRF under the guidance of 2D diffusion priors leveraging a pre-trained large-scale text-to-image diffusion model \citep{rombach2022high}.

While achieving impressive results, these works face a critical challenge: they rely on an optimization-based learning paradigm, requiring thousands of iterations to optimize a NeRF for each new input (top of \fig~\ref{fig:teaser}). This incurs substantial computational costs and leads to notably slow response speed for practical text-to-3D systems, often taking hours to process a single input~\citep{poole2022dreamfusion,wang2023prolificdreamer}. 
Moreover, since the NeRFs are separately learned for different text inputs, they are not able to leverage general 3D priors shared across objects.

Note that Point-E \citep{nichol2022point} abandons this optimization paradigm by directly training a diffusion model for 3D point cloud generation.
However, these diffusion models require a large number of iterative diffusion steps, which inevitably leads to low efficiency. Besides, it is trained on millions of private 3D assets, which is not feasible for most researchers. 

In this paper, we propose a novel framework, named \name{}, for fast text-to-3D generation. It is able to generate a realistic 3D object that faithfully aligns with a given text description, all within a single forward pass of a neural network. Remarkably, this generation process takes less than one second (bottom of \fig~\ref{fig:teaser}). 
The proposed \name{} is a feedforward neural network conditioned on the text prompt and produces a triplane representation~\citep{chan2022efficient} encapsulating the desired 3D object. 
Similar to DreamFusion~\citep{poole2022dreamfusion}, we use the Score Distillation Sampling (SDS) loss to train our network, eliminating the need for costly 3D training data.

While conceptually simple, this network is substantially challenging to design in practice. 
The main reason is that the SDS loss provides relatively weak supervision signals, making it difficult to learn an accurate connection between the text condition and the 3D output for a common condition model. 
This stands in contrast to training models that connect 2D images with texts, such as Stable Diffusion~\citep{rombach2022high} and CLIP~\citep{wang2022clip}, where numerous image-text pairs can provide strong supervision signals. 

The challenge of weak supervision motivates us to devise more effective condition mechanisms that can better inject text information into the network, alleviating the difficulty of learning the relationship between the two drastically different modalities: text and 3D.
As the core innovation of \name{}, our proposed solution to this challenge involves the fusion of three modules: cross-attention, style injection, and token-to-plane transformation. The integration of these components collectively ensures the precise alignment of the generated 3D output with the condition text prompt.

Moreover, we also propose a simple yet effective activation function, called scaled-sigmoid, to replace the original sigmoid function in NeRF, which speeds up the network convergence by more than ten times. 

Meanwhile, text-to-3D methods often suffer from severe Janus (multi-head) problems. 
While the Perp-Neg algorithm~\citep{armandpour2023re} can effectively remedy this issue for existing text-to-3D methods, it does not perform well in our context.  
We find that this is mainly due to the different degrees of multi-head effect exhibited by different 3D objects. 
As previous text-to-3D approaches only train a single object at a time, one can easily tune the concept negation scale in the Perp-Neg algorithm to adapt to the specific multi-head level. 
However, in our framework, it is challenging to find a universal, optimum concept negation scale for all various training samples.
To tackle this issue, we present an adaptive variant of the Perp-Neg algorithm which dynamically adjusts its concept negation scales according to the severity of the multi-head effect during training, thereby significantly improving the performance. 

Our approach demonstrates a powerful capacity to generate a high-quality 3D object consistent with a novel text prompt in only \textit{one second}. 
We conduct extensive evaluations of the proposed algorithm on a wide variety of benchmark datasets, showing that the \name{} performs favorably against the state-of-the-art approaches both qualitatively and quantitatively with much improved efficiency.

In summary, our contributions are as follows:
\begin{itemize}[noitemsep,topsep=0pt]
\item We make early explorations of fast text-to-3D generation with a single run of a feedforward neural network. 
\item We propose a novel condition model that can effectively absorb text information, facilitating the establishment of the connection between text and 3D under weak supervision.
\item We present a new activation function, called scaled-sigmoid, which significantly accelerates the training convergence.
\item We present an adaptive Perp-Neg algorithm to better tackle the Janus problem. 
\end{itemize}

\section{Related Works}
\label{sec:related}
\bfsection{Text-to-Image} 
Previously, the research in this field mainly concentrates on generating images belonging to a specific domain or distribution~\citep{reed2016generative, Zhang_2017_ICCV, Xu_2018_CVPR, Qiao_2019_CVPR, Tan_2019_ICCV, Ruan_2021_ICCV}. The literature is dominated by Generative Neural Networks (GANs)~\citep{goodfellow2020generative} and various invariants of GANs are proposed.  
Recently, inspired by large-scale image-text pairs~\citep{schuhmann2021laion, schuhmann2022laion} and generative models, text-to-image generation has shown unprecedented imaginations, \ie synthesizing all sorts of imaginative images corresponding to text prompts. DALL-E~\citep{ramesh2021zero} and CogView~\citep{ding2021cogview} are developed based on an autoregressive architecture~\citep{Esser_2021_CVPR, ramesh2021zero}, while GLIDE~\citep{nichol2021glide} employs a diffusion model~\citep{song2020denoising, ho2020denoising} conditioned on CLIP guidance~\citep{radford2021learning}. DALL-E-2 ~\citep{ramesh2022hierarchical} also leverages diffusion priors to translate CLIP text embeddings to CLIP image embeddings, followed by synthesizing images from them. Imagen~\citep{saharia2022photorealistic} employs a large pre-trained language model to guide the reverse process of a diffusion model in pixel space. It is similar to Stable Diffusion~\citep{rombach2022high}. Differently, the latter is based on a large-scale UNet architecture and works in the vector-quantized discrete latent space, enabling an efficient sampling process.

\bfsection{Text-to-3D} 
Recent methods of text-to-3D generation without 3D supervision target to generate 3D objects corresponding to input prompts only with the guidance from CLIP or text-to-image models~\citep{Sanghi_2022_CVPR, liu2022iss, Jain_2022_CVPR, khalid2022clipmesh, poole2022dreamfusion, lee2022understanding,yi2023progressive}. By leveraging CLIP embeddings to make the generated object closer to the text prompt, DreamFields~\citep{Jain_2022_CVPR} and CLIP-Mesh~\citep{khalid2022clipmesh} trigger the research in this field. They represent 3D objects by NeRFs and spherical meshes, respectively. PureCLIPNeRF~\citep{lee2022understanding} follows their paradigm except for replacing NeRFs or spherical meshes with grid-based representation~\citep{Sun_2022_CVPR_DVGO}. Inspired by pre-trained large-scale text-to-image diffusion models, DreamFusion~\citep{poole2022dreamfusion} proposes a score distillation sampling loss to distill 2D image priors into 3D generation process and achieves much better results than previous works. A concurrent work SJC \citep{wang2023score} presents a similar approach. Following this line, Latent-NeRF \citep{metzer2022latent} proposes to learn 3D representations in the latent space instead of pixel space and incorporate more guidance like sketch shapes into the object generation. To solve the problem of low-diversity introduced by SDS, ProlificDreamer \citep{wang2023prolificdreamer} presents variational score distillation to construct multiple particles for a single prompt input, sampling the optimal 3D representation from the corresponding probabilistic distribution. These works mainly focus on optimizing an implicit 3D representation for a prompt. Given a new prompt, they need to repeat the optimization process, taking more than one hour. In contrast, our \name{} is able to generate a high-quality 3D object for a new prompt in less than one second after training. 

A concurrent work ATT3D~\citep{lorraine2023att3d} follows a similar paradigm to ours, learning a neural network for fast text-to-3D generation. 
Our \name{} differs from ATT3D in that ATT3D employs a straightforward MLP to learn a hash grid~\citep{mueller2022instant}, while we devise a novel decoder architecture with enhanced condition mechanisms and a scaled-sigmoid function to generate triplanes, which significantly improves the results.
In addition, we present an adaptive Perp-Neg algorithm that effectively tackles the Janus problem.
The proposed method achieves higher generation quality and more accurate text-3D alignment than ATT3D, as will be seen later in the experiment section.

\bfsection{3D Priors in Generation}
\textcolor{black}{Recently, there has been another line of research that introduces 3D priors into 3D generation. MVDiffusion~\citep{MVDiffusion} pioneers at this direction by proposing the correspondence-aware attention to maintain cross-view consistency primarily for text-to-panorama generation. MVDream~\citep{shi2023mvdream} designs a multi-view diffusion model with 3D self-attentions. After training on a 3D dataset, the model can provide strong 3D diffusion priors via multi-view SDS. SweetDreamer~\citep{li2023sweetdreamer} tunes a 2D diffusion model to produce canonical coordinate maps for a prompt, conferring 3D geometric consistency in generation. Additionally, there are other works investigating multi-view consistency for image-to-3D generation. SyncDreamer~\citep{liu2023syncdreamer} proposes a synchronized multi-view diffusion model with a 3D-aware interaction mechanism to correlate the features across different views.
Wonder3D~\citep{long2023wonder3d} presents a multi-view cross-domain attention mechanism to improve the consistency and develops a geometry-aware normal fusion algorithm to extract high-quality surfaces. One-2-3-45~\citep{liu2023one} reconstructs 3D object meshes in a feedforward manner in $45$ seconds based on SDF-based neural surface reconstruction and Zero123~\citep{liu2023zero}.  
}

\textcolor{black}{Specially, a concurrent work~\citep{li2023instant3d} proposes a two-stage pipeline of applying 3D priors from Objaverse \citep{deitke2023objaverse} for ensuring multi-view consistency. They first train a multi-view diffusion model to generate a sparse set of view images for a given text and then optimize a large reconstruction model to directly regress the NeRF from the generated images. Its inference process takes 20 seconds for generating one object, much longer than ours (20ms).  Moverover, our \name{} does not rely on any 3D dataset and only applies 2D diffusion priors from a text-to-image model for 3D object generation. 
}

\begin{figure*}[t]
\centering
\includegraphics[width=0.99\textwidth]{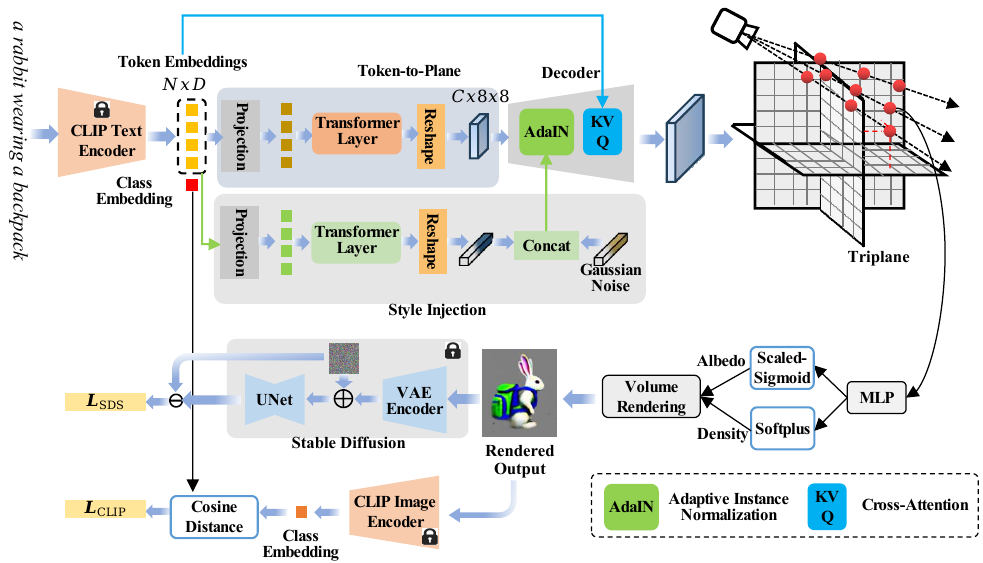}
\caption{\textcolor{black}{Overview of the proposed \name{}, which applies a conditional decoder network to map a text prompt to a corresponding triplane. Three condition mechanisms, \ie cross-attention, style injection, and token-to-plane transformation, are seamlessly combined to bridge text and 3D, tackling the issue of weak supervision from SDS~\citep{poole2022dreamfusion}. Given a random camera pose, a 2D image is rendered from the conditioned triplane through coordinate-based feature sampling, point density and albedo prediction, and differentiable volume rendering. For albedo activation, we propose a scaled-sigmoid function, effectively accelerating the training convergence. During training, the view image is first diffused by adding random noise and then fed into a pretrained UNet conditioned on the text prompt for denoising, which provides the gradient of the SDS loss. We also present an adaptive Perp-Neg algorithm to better solve the Janus problem in our framework. During inference, our \name{} can infer a faithful 3D object from an unseen text prompt in less than one second.}    
}
\label{fig:overview}
\end{figure*}

\section{Methodology}\label{sec:method} 
\subsection{Preliminaries}
\subsubsection{Neural Radiance Field}
NeRF~\citep{mildenhall2020nerf} is an implicit 3D representation widely used in neural inverse rendering.
Given a NeRF, rendering an image for a camera view $c$ involves casting a ray for each pixel of the image; these rays originate from the camera’s center of projection and extend through the pixel’s location on the image plane, reaching out into the 3D world. 
Sampled 3D points $p$ along each ray are then passed through a Multi-Layer Perceptron (MLP), which produces 4
scalar values as output: 
\begin{equation} \label{eq:nerf}
\begin{aligned}
    \tau &= f_\text{softplus} \left(f_\text{MLP}(p;\theta)[1]\right), \\
    \bm{\rho} &= f_\text{sigmoid} \left(f_\text{MLP}(p;\theta)[2:4]\right),
\end{aligned}
\end{equation}
where $\tau \in \mathbb{R}$ is volumetric density, indicating the opacity of the scene geometry at the 3D coordinate. 
$\bm{\rho} \in \mathbb{R}^3$ is the albedo, capturing the intrinsic colors that define the scene.
$f_\text{MLP}$ represents an MLP with parameters $\theta$.
$[i:j]$ denotes extracting channels from $i$ to $j$.
$f_\text{softplus}$ is the Softplus activation function, ensuring positive volumetric density.
$f_\text{sigmoid}$ is the sigmoid function, which confines the albedo to the range of $[0, 1]$.
These density and albedo values on a ray are then composited with volume rendering, producing the RGB value for the pixel.
We use a shading formulation similar to DreamFusion~\citep{poole2022dreamfusion} in rendering.
Carrying out this process for all pixels on the image plane results in the final rendered RGB image denoted as $g_\theta(c)$, which is parameterized by $\theta$ and conditioned on the camera view $c$.

\subsubsection{Score Distillation Sampling (SDS)}\label{sec:sds}
Most existing text-to-3D methods are based on the SDS loss~\citep{poole2022dreamfusion} powered by text-to-image diffusion models~\citep{rombach2022high}.
It facilitates the generation of a NeRF from a text prompt by enforcing that the rendered image from the NeRF at any viewpoint maintains semantic consistency with the given text prompt.

In detail, given a rendered image $\vg_{\theta}(c)$, the SDS loss first introduces random noise $\epsilon\sim \mathcal{N}(\bm{0},\bm{I})$: 
\begin{align}\label{eq:forward}
    \bm{x}_t = \sqrt{\alpha_t} \vg_{\theta}(c) +  \sqrt{1-\alpha_t}\bm{\epsilon},
\end{align}
where $\alpha_{1:T}$ is a predefined decreasing sequence, and the time step $t$ is selected randomly.
Subsequently, a pretrained diffusion UNet $f_\text{diffusion}$ is employed to predict the noise in \equ~\ref{eq:forward} from $\bm{x}_t$, which is proportional to the negative score function (NSF) of the image distribution~\citep{song2021scorebased}:
\begin{align}
&\bm{\epsilon}_{\text{unc}} = f_\text{diffusion}(\bm{x}_{t};t), \nonumber \\
&\bm{\epsilon}_{\text{pos}} = f_\text{diffusion}(\bm{x}_{t};\bm{y},t) - \bm{\epsilon}_{\text{unc}}, \\
&\bm{\epsilon}_{\text{predict}} = \bm{\epsilon}_{\text{unc}} + w_{\text{guidance}} \bm{\epsilon}_{\text{pos}}, \nonumber
\end{align}
where $\bm{\epsilon}_{\text{unc}}$ corresponds to the NSF of the unconditional data distribution, 
and $\bm{\epsilon}_{\text{pos}}$ is proportional to the residual NSF conditioned on the text embedding $\bm{y}$.
$\bm{\epsilon}_{\text{predict}}$ is based on a classifier-free guidance formulation~\citep{ho2022classifier}, indicating the direction for updating the image to match the text description $\bm{y}$, and $w_{\text{guidance}}$ denotes the guidance weight.

Finally, the gradient of the SDS loss with respect to the NeRF parameters $\theta$ can be written as:
\begin{equation}
    \begin{split}
    \nabla_{\theta}\mathcal{L}_{\textnormal{SDS}}(\theta) = \mathbb{E}_{t, \bm{\epsilon}}\left[w(t)(\bm{\epsilon}_{\text{predict}}-\bm{\epsilon})\frac{\partial \vg_{\theta}(c)}{\partial\theta}\right], 
    \end{split}
    \label{eq:sds}
\end{equation}
where $w(t)$ is a weight function depending on $\alpha_{t}$.

\subsection{Conditional NeRF Generation}
By combining NeRF and SDS, existing methods~\citep{poole2022dreamfusion,metzer2022latent,wang2022score} have demonstrated impressive results for text-to-3D generation. 
However, these methods suffer from an important drawback that they need to train a new NeRF for each new text prompt, leading to low efficiency.
To address this issue, we propose to learn a single feedforward network \name{} for multiple text prompts, enabling fast text-to-3D generation.
An overview of \name{} is shown in \fig~\ref{fig:overview}.

In particular, we design a conditional network $f_\text{Instant}(p,\bm{y};\theta)$, which is conditioned on the text feature $\bm{y}$. 
We obtain the text feature with the pretrained text encoder of CLIP~\citep{wang2022clip}.
The conditional network can be formulated as:
\begin{align}\label{eq:instant3d}
f_\text{Instant}(p,\bm{y};\theta) = \hat{f}_\text{MLP}\left(f_\text{sam}\left(p,f_\text{dec}(\bm{y};\theta_1)\right);\theta_2\right).
\end{align}
Inspired by StyleGAN~\citep{karras2019style}, we adopt a decoder architecture $f_\text{dec}$ to transform the text condition $\bm{y}$ into a triplane~\citep{chan2022efficient}.
Then we sample a feature vector for each 3D point $p$ from the triplane using $f_\text{sam}$, which projects $p$ onto each plane, retrieves the corresponding feature vector via bilinear interpolation, and aggregates the three feature vectors via concatenation~\citep{chan2022efficient}.
This point feature is subsequently sent to an MLP $\hat{f}_\text{MLP}$ to produce the density and albedo values.
The parameters of the conditional model $\theta$ include both the decoder parameters $\theta_1$ and MLP parameters $\theta_2$.
Replacing $f_\text{MLP}$ with $f_\text{Instant}$ in \equ~\ref{eq:nerf} enables conditional NeRF prediction, where different $\bm{y}$ yields different NeRF outputs, and the learned model parameters $\theta$ are shared among diverse text prompts. 

The key component in \equ~\ref{eq:instant3d} is the decoder network which plays the role of transferring text information into 3D. 
\textcolor{black}{Developing a decoder within our framework, despite seeming straightforward, presents significant obstacles. A key hurdle is the weak supervision signals from the SDS loss, as outlined in \equ~\ref{eq:sds}, making it difficult to establish a reliable link between the textual conditions and the 3D output. This situation starkly differs from that of text-to-image models like Stable Diffusion~\citep{rombach2022high} and CLIP~\citep{wang2022clip}, which benefit from the strong guidance provided by numerous image-text pairs.}

The weak supervision issue demands a more effective conditional model that can better absorb text information, alleviating the difficulty of bridging text and 3D.
To this end, we introduce an integrated solution combining three condition mechanisms: cross-attention, style injection, and token-to-plane transformation. 

\begin{figure}
    \centering
    \setlength{\tabcolsep}{1pt}
    {\scriptsize
    \begin{tabular}{c c c}
    \includegraphics[width=0.3\columnwidth, height=0.26\columnwidth]{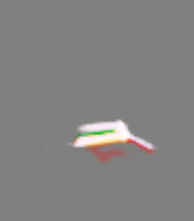}&
    \includegraphics[width=0.3\columnwidth, height=0.26\columnwidth]{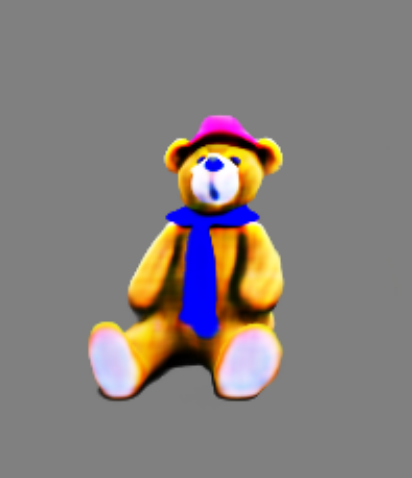}&
    \includegraphics[width=0.3\columnwidth, height=0.26\columnwidth]{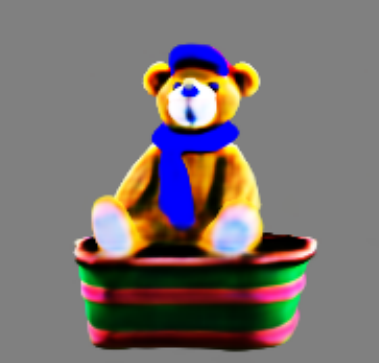}
    \end{tabular}
    \begin{tabular*}{0.9\columnwidth}{P{0.3\columnwidth}P{0.3\columnwidth}P{0.3\columnwidth}@{}}
    \centering
    Cross-Attention &+ Style Injection &+ Token-to-Plane Transformation
    \end{tabular*}
    }
    \caption{Integrating three condition mechanisms produces high-quality results faithful to the text prompt. The prompt here is ``a teddy bear sitting in a basket and wearing a scarf and wearing a baseball cap''.
    }
    \label{fig:comparison_crossatt_style_plane}
\end{figure} 

\subsubsection{Cross-Attention}
\textcolor{black}{Cross-attention is one of the most important strategies for information interaction across various modalities.
Following the prominent success of text-to-image models, we apply cross-attention to inject text descriptions into the 3D representation.
Similar to Stable Diffusion~\citep{rombach2022high}, we employ the multi-head cross-attention module (MHCA) to fuse text embeddings with feature maps of the decoder. Specifically, the query ($Q$) is projected from the feature maps, while the key ($K$) and value ($V$) are projected from text embeddings. In principle, the interacted output of the cross-attention module can be formulated as:
\begin{equation}
\operatorname{Attention}(Q, K, V)=\operatorname{Softmax}\left(\frac{Q K^T}{\sqrt{d}}\right) V,
\end{equation}
where $d$ is the channel dimension of $K$.}

\subsubsection{Style Injection}
\textcolor{black}{Surprisingly, the model with only the cross-attention mechanism completely fails, yielding a meaningless output as shown in \fig~\ref{fig:comparison_crossatt_style_plane}.
\name{} aims to generate 3D objects whose 2D renderings closely align with the synthesized images of Stable Diffusion for the corresponding prompts.
Inspired by the GAN literature~\citep{arjovsky2016towards}, we hypothesize that this optimization failure is mainly due to that there is a significant discrepancy between the 2D renderings from \name{} and the images generated by Stable Diffusion during the initial stage of training. \cite{arjovsky2016towards} highlight that the optimization gradient is unreliable when the generated and target distributions are disjoint. By integrating Gaussian noise into the network, the overlap between the rendered and targeted distributions can be enhanced, thereby leading to more meaningful gradients during training. To this end, we introduce the Adaptive Instance Normalization (AdaIN) to inject random noise into the decoder.}

\textcolor{black}{
In addition to noise, our style injection module also encodes text features into the decoder to improve the controllability of the 3D generation process. 
As shown in \fig~\ref{fig:overview}, it starts with a linear projection layer to more compactly represent the text embeddings. 
Subsequently, we apply a self-attention based Transformer layer to adapt the feature to the style space. 
The output text features are flattened to a vector and concatenated together with random Gaussian noise to generate the style vector. 
Finally, the output vector of the style injection module is embedded into the feature maps of our decoder network with AdaIN in \fig~\ref{fig:overview}.}

\textcolor{black}{Specifically, one linear layer is applied to project the output vector of the style injection module into styles
$\mathbf{l}=\left(\mathbf{l}_s, \mathbf{l}_b\right)$ that control AdaIN operations after each convolution
block of the decoder network. The AdaIN operation can be formulated as
\begin{equation}
\operatorname{AdaIN}\left(\mathbf{m}_i, \mathbf{l}\right)=\mathbf{l}_{s, i} \frac{\mathbf{m}_i-\mu\left(\mathbf{m}_i\right)}{\sigma\left(\mathbf{m}_i\right)}+\mathbf{l}_{b, i},
\end{equation}
where each input feature channel $\mathbf{m}_i$ is normalized independently, and then scaled and biased by the corresponding scalar components from the style $\mathbf{l}$. Thus the dimensionality of $\mathbf{l}$ is twice the number of the input feature channels.}

\begin{figure}[t]
\setlength\abovecaptionskip{2mm}
\centering
\begin{subfigure}{0.485\columnwidth}
    \centering
    \includegraphics[width=1.0\columnwidth]{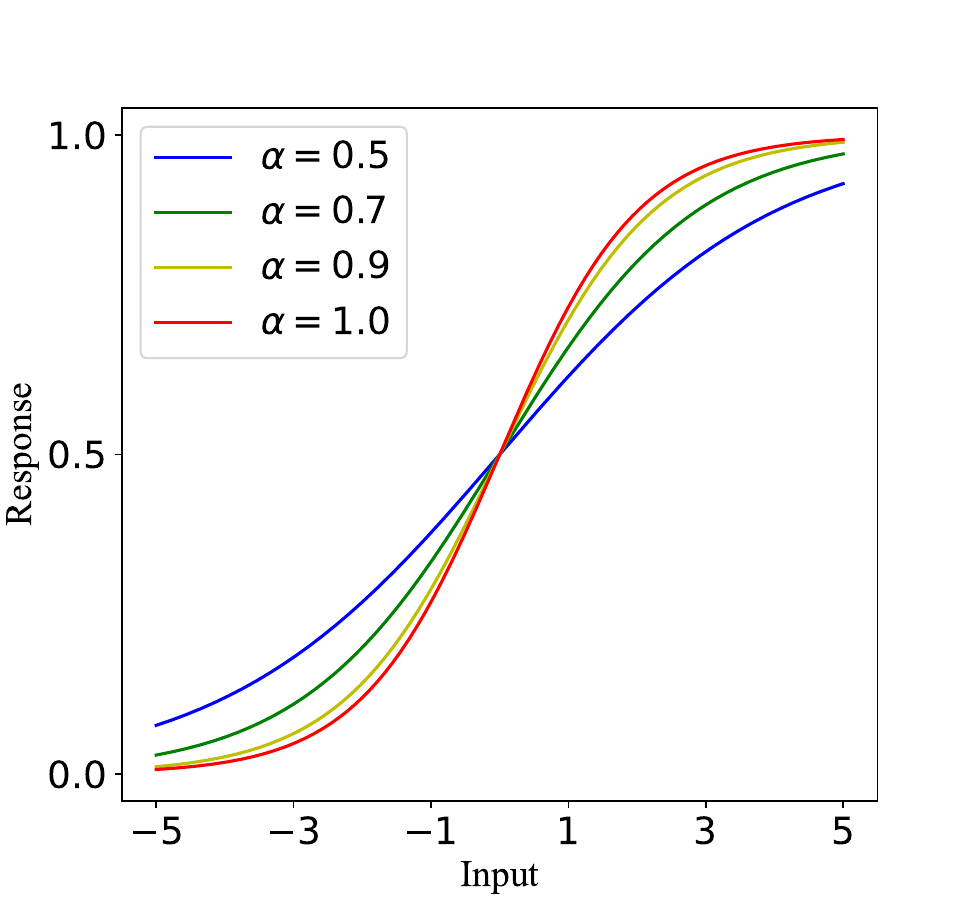}
\end{subfigure}
\hspace{-2pt}
\begin{subfigure}{0.485\columnwidth}
    \centering
    \includegraphics[width=1.0\columnwidth]
    {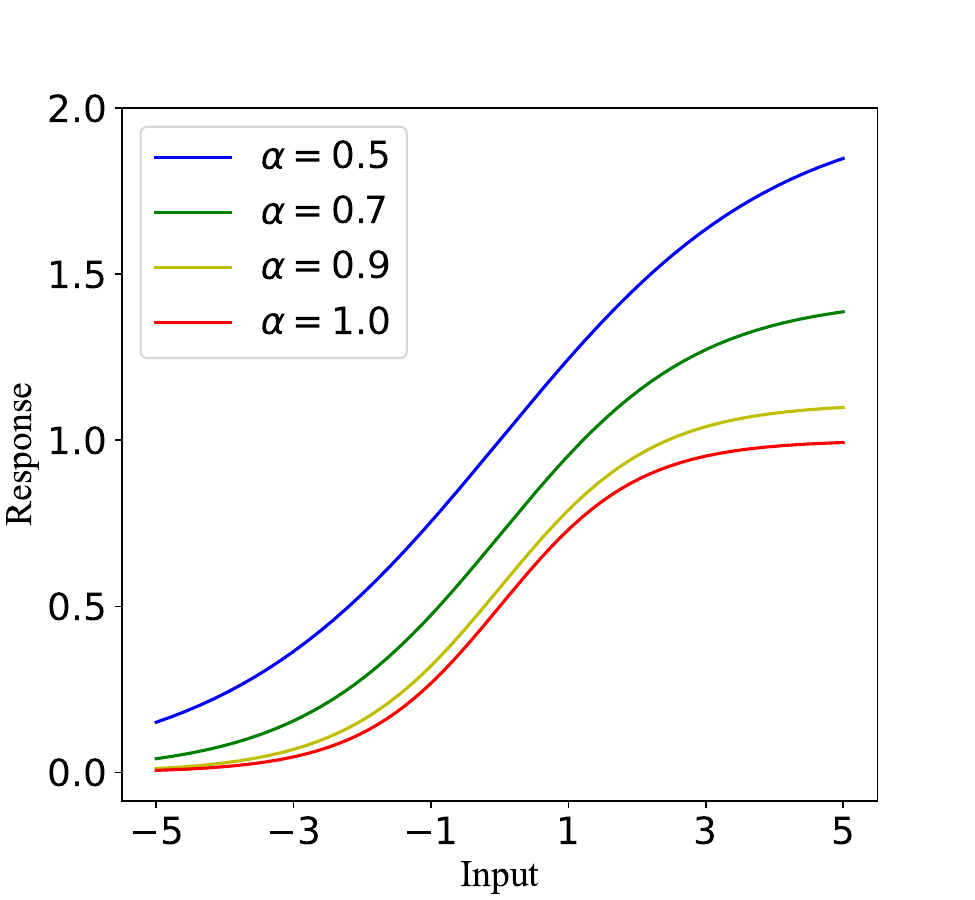}
\end{subfigure}
\caption{Comparison of the original sigmoid function with its stretched variant (left) and the proposed scaled-sigmoid function (right). With $\alpha<1.0$, the scaled-sigmoid function possesses a broader high-gradient region, which accelerates training.}
\vspace{-5mm}
\label{fig:sigmoid}
\end{figure}

\subsubsection{Token-to-Plane Transformation}
Existing decoder architectures typically start with a learnable planar base tensor~\citep{karras2019style,Karras2019stylegan2,Karras2021},
which is kept constant after training.
In contrast, we reformulate this paradigm by predicting the base tensor from the text embedding with a token-to-plane transformation module as shown in \fig~\ref{fig:overview}. 
Consequently, the base tensor evolves from a static entity to a dynamic one, significantly deviating from the common practices of StyleGAN, which can better convey condition information to the decoder.

As illustrated in \fig~\ref{fig:overview}, we employ a self-attention based Transformer layer to transform sequential token embeddings into a 2D feature map. 
Specifically, we first reduce the dimension of tokens with linear projection for more efficient computations. 
Then a multi-head self-attention (MHSA) layer~\citep{vaswani2017attention} is performed to enable global information interaction among tokens. 
We further apply an MLP to refine the token embeddings and then reshape them to a feature map with the shape of $8\times 8$. 
The token-to-plane transformation not only transforms the sequential tokens into a desired planar shape, but also shifts the CLIP embeddings to a feature space suitable for text-to-3D generation. 

As shown in \fig~\ref{fig:comparison_crossatt_style_plane}, the integration of all three condition mechanisms collectively ensures the precise alignment of the generated 3D output with the condition text prompt.

\bfsection{Text Feature}
The text encoder of CLIP~\citep{radford2021learning} generates both \text{class} and \text{token} embeddings for a text prompt. 
The \text{class} embedding usually contains sentence-level semantic information. 
In contrast, the \text{token} embeddings contain rich word-level semantic details of a prompt, which can better represent items in a scene and are commonly used in text-to-image generation \citep{rombach2022high}.  
We aim for the generated 3D objects to reflect all mentioned entities and maintain their inherent relationships, matching the token-level granularity of the text prompt.
Thus, we use the \text{token} embeddings as the text feature to guide our 3D generation, which leads to superior generation results on novel prompts. 

\subsection{Scaled-Sigmoid}\label{sec:scaled-sigmoid}
As shown in \equ ~\ref{eq:nerf}, the sigmoid function is usually used to ensure the albedo prediction falls within the range of $[0,1]$.
However, we observe that applying the conventional sigmoid to albedo causes difficulties in training convergence. 
A deeper analysis reveals that the MLP output may exceed the high-gradient region around zero, leading to gradient vanishing. 
This phenomenon significantly impedes the learning of the 3D representation, resulting in extended training durations and occasional convergence failures. 

To address this issue, we stretch the high-gradient region by multiplying the input with a fractional coefficient $\alpha$:
\begin{align}
    \hat{f}_\text{sigmoid}(x) = f_\text{sigmoid}(\alpha x), \quad \alpha \in (0,1].
\end{align}
The behavior of $\hat{f}_\text{sigmoid}$ with different $\alpha$ is shown in \fig~\ref{fig:sigmoid} (left).
Note that the stretching operation reduces the gradient around $x=0$ by a factor of $\alpha$. 
To counter this effect, we further apply a scaling factor of $\frac{1}{\alpha}$ to the stretched sigmoid:
\begin{align}
    \tilde{f}_\text{sigmoid}(x) = \frac{1}{\alpha} f_\text{sigmoid}(\alpha x), \quad \alpha \in (0,1].
\end{align}
We refer to $\tilde{f}_\text{sigmoid}$ as the scaled-sigmoid, which is visualized in \fig~\ref{fig:sigmoid} (right).
Since $\tilde{f}_\text{sigmoid}$ can surpass the $[0,1]$ bounds,
we employ an annealing strategy: initializing $\alpha$ with a small value (0.5 in our experiment) to accelerate training; over time, we gradually increase $\alpha$ to 1, ensuring the output albedo lies in $[0,1]$.

\subsection{Adaptive Perp-Neg Algorithm}\label{sec:adaptive perp neg}
\bfsection{Previous Perp-Neg Algorithm}
Existing text-to-3D algorithms often face a challenging Janus (multi-head) problem~\citep{metzer2022latent}. 
Rather than generating a coherent 3D output, the learned 3D object tends to exhibit repeated front views at different angles, as the front view has been more prominent in the training data of text-to-image models. 
For instance, when generating a 3D animal, the resulting output often has multiple faces without capturing the natural side and back views of the animal.

To address this issue, \cite{armandpour2023re} propose a Perp-Neg algorithm:
\begin{equation}
\begin{aligned}
    &\bm{\epsilon}_{\text{predict}} = \bm{\epsilon}_{\text{unc}} + w_{\text{guidance}} \left(\bm{\epsilon}_{\text{pos}}  - w_{\text{neg}}{\bm{\epsilon}_{\text{neg}}}^{\perp}\right), \\
    &\bm{\epsilon}_{\text{neg}} = f_\text{diffusion}
        (\bm{x}_{t};\bm{y}_\text{neg},t) - \bm{\epsilon}_{\text{unc}},
    \label{eq:perp-neg}
\end{aligned}
\end{equation}
where $\bm{\epsilon}_{\text{predict}}$ is used in \equ~\ref{eq:sds} to calculate the gradient of the SDS loss. 
$\bm{y}_\text{neg}$ denotes the token embeddings of negative prompts.
For example, when rendering images at the back view, the negative prompts refer to side and front views.
${\bm{\epsilon}_{\text{neg}}}$ is the residual NSF of negative text prompts, penalizing the generation of wrong views at inappropriate angles.
${\perp}$ indicates projection onto the direction perpendicular to $\bm{\epsilon}_{\text{pos}}$.
$w_{\text{neg}}$ represents the concept negation scale, controlling the degree to which \equ~\ref{eq:perp-neg} penalizes the Janus problem. 

\bfsection{Adaptive Adjustment}
Selecting an appropriate value for $w_{\text{neg}}$ is critical for the success of Perp-Neg.
If it is too small, the Janus effect will not be sufficiently mitigated.
If it is too large, it may cause the flat-head issue, where the entire head is squeezed into a flat plane. This is because the $\bm{y}_\text{neg}$ in \equ~\ref{eq:perp-neg} punishes front views when rendering in side angles, and a flat head ensures no face can be seen from a side perspective.
For existing optimization-based methods, it is feasible to manually adjust $w_{\text{neg}}$ for each new text prompt. 
However, as our model is trained on an extensive set of text prompts, it is challenging to find a universally optimum concept negation scale for all training samples. 
To overcome this challenge, we propose an adaptive Perp-Neg algorithm to dynamically adjust the value of $w_{\text{neg}}$ according to the severity of the multi-head problem for different samples.

Specifically, we design the negation scale in our adaptive Perp-Neg as
\begin{align}
w_{\text{neg}}=w_{\text{min}}+ C \cdot \Delta w,
\end{align}
where $w_{\text{min}}$ is the minimum scale, and $\Delta w$ describes the variation range of $w_{\text{neg}}$.
The adaptive parameter $C \in [0,1]$ measures the severity of the multi-head problem.
When the multi-head effect is less pronounced ($C=0$), we apply a small punishment in \equ~\ref{eq:perp-neg} with $w_{\text{neg}}=w_{\text{min}}$.
Conversely, when the multi-head effect is highly severe ($C=1$), the negation scale increases to $w_{\text{min}}+\Delta w$.

To assess the severity of the multi-head problem for a given rendered image $I_v$ at the current viewpoint $v$, we define $C$ as:
\begin{equation}\label{eq:perp-neg-C}
    C = \frac{1}{4n} \sum_{i=1}^n \left[1-\cos(v-v_i)\right] \cdot \left[1+\left\langle \phi(I_{{v}}), \phi(I_{{v_i}})\right\rangle\right], 
\end{equation}
where we randomly select $n$ additional views $\{v_i\}_{i=1}^n$ and measure the similarity between $I_{{v}}$ and each $I_{v_i}$ in feature space.  
The feature extractor $\phi$ is the image encoder of CLIP~\citep{wang2022clip}, and $\left\langle \cdot,\cdot \right\rangle$ represents the cosine similarity.
The rationale for this design is that a higher similarity between different views indicates more severe multi-head effect, and conversely, a lower similarity value among different views suggests a less significant Janus problem.
Besides, we weight the contribution of different view angles in \equ~\ref{eq:perp-neg-C} with $1-\cos(v-v_i)$. This weighting mechanism takes into account that high similarity between views that are farther apart indicates a more severe multi-head effect.

While \equ~\ref{eq:perp-neg-C} offers an effective approach for adjusting the negation scale adaptively, it comes with a notable computational overhead. This is primarily due to the extra cost involved in rendering $\{I_{v_i}\}_{i=1}^n$.
To address this issue, we adopt a simplified approach by setting $n=1$ and utilizing the rendered image from the previous iteration as the additional view. 
To facilitate this process, we maintain a cache that stores image features $\phi(I_{{v}})$ for all training samples. This cache enables us to reuse the feature representations extracted during the preceding training steps. 
The features in the cache are dynamically updated, where the current feature is added while the old feature is removed. 
This approach allows us to implement adaptive Perp-Neg without introducing additional computational costs. 

\subsection{Training Objective}
Except for the SDS loss in \sec~\ref{sec:sds}, the proposed \name{} also incorporates a CLIP loss $\mathcal{L}_{\text{CLIP}}$. 
As shown in \fig~\ref{fig:overview}, we employ the CLIP image encoder to extract image features from the rendered image and use the class embeddings from the CLIP text encoder to represent the text prompt.
We then define the CLIP loss as the cosine distance between the image and text features to align the rendered image more closely with the input prompt.
Remarkably, these two training objectives allow learning a text-to-3D model without requiring any 3D data.

\section{Experiments}
\subsection{Datasets} 
As this work is a pioneering effort for fast text-to-3D generation, we devise three benchmark prompt sets to comprehensively evaluate the proposed framework: Animals, Portraits, and Daily Life.

\bfsection{Animals}
This prompt set is constructed by combining several keywords, where each keyword can be randomly chosen from a predefined set of candidates:
\begin{itemize}[noitemsep,topsep=0pt]

\item \textbf{species}: \textit{wolf, dog, panda, fox, civet, cat, red panda, teddy bear, rabbit,} and \textit{koala}.

\item \textbf{item}: \textit{in a bathtub, on a stone, on books, on a table, on the lawn, in a basket}, and \textit{null}.

\item \textbf{gadget}:  \textit{a tie, a cape, sunglasses, a scarf} and \textit{null}.

\item \textbf{hat}: \textit{beret, beanie, cowboy hat, straw hat, baseball cap, tophat, party hat, sombrero}, and \textit{null}.

\end{itemize}
A sample prompt is structured as ``a \textbf{species} sitting \textbf{item} and wearing \textbf{gadget} and wearing a \textbf{hat}''. The \textit{null} means the keyword is absent from the prompt. In total, the Animals set consists of 3,150 prompts. We randomly allocate 60\% of them for training and the remaining 40\% for testing.

\bfsection{Portraits}
We construct this prompt set by describing portraits of commonly seen figures with various hats and facial expressions. The keywords and their available choices include: 

\begin{itemize}[noitemsep,topsep=0pt]
\item \textbf{figure}: \textit{a white man, a white woman, a boy, a girl, an elderly man, an elderly woman, a black woman, a black man, Obama, Kobe.}

\item \textbf{hat}: \textit{Santa hat, peaked cap, steampunk hat, crown.}

\item \textbf{expressing}: \textit{laughing, crying, grinning, singing, shouting, looking ahead with a very serious expression, opening mouth wide in shock, angry, talking, feeling sad.}
\end{itemize}
These keywords are combined in the manner of ``\textbf{figure} wearing a \textbf{hat} is \textbf{expressing}''. The Portraits set consists of 400 prompts in total, with 60\% for training and 40\% for testing.  

\bfsection{Daily Life}
The above two prompt sets are constructed by combining phrases from a limited range of candidates in a structured manner. 
To demonstrate that our approach can work well in a more general setting where more complex and varied sentences are included, we contribute another dataset named Daily Life by collaborating with ChatGPT \citep{brown2020language}. 
ChatGPT is a large language model designed to converse with humans and can produce high-quality responses when given appropriate context. 
We craft our question as ``\textit{generate a text describing a man or woman's appearance and daily activities. The examples are as follows: a tall man wearing sunglasses is eating a hamburger, a slim woman wearing a baseball cap is reading a book, a beautiful woman wearing a tie is watching a TV}''. 
By repeatedly submitting this question to ChatGPT, we collect more than 17,000 answers as our prompt set. 
These prompts exhibit a wide variety of structures and contain 3,135 unique words in total, demonstrating enhanced complexity and diversity. 
We train our framework on the whole set and perform evaluation by generating additional new prompts via ChatGPT. 

\subsection{Implementation Details}
\bfsection{Camera Setting}
For NeRF rendering, we randomly sample a camera pose in a spherical coordinate system with the polar angle $\gamma \in [25\degree, 110\degree]$, azimuth angle $\phi \in [0\degree, 360\degree]$, and radius from the origin $r\in[3.0, 3.6]$. The camera FOV is randomly selected between $70\degree$ and $80\degree$.

\bfsection{Neural Rendering}
For volumetric rendering, 128 points are sampled along each ray. Among them, 64 points are uniformly sampled, and the remaining 64 are sampled based on the importance distribution along the ray. The resolution of rendered images is $64\times64$ during training and increased to $256\times256$ for testing.

\bfsection{Architecture}
Our decoder network is composed of attention blocks and convolution blocks similar to Stable Diffusion~\citep{rombach2022high}.
More details can be found in the appendix.

\bfsection{Optimization} 
We use the Adam optimizer~\citep{kingma2014adam} with a learning rate of 1e-4, $\beta_1=0.9$, $\beta_2=0.99$ and a zero weight decay. 
All experiments are conducted on NVIDIA A100 GPUs. 
Training on Animals and Daily Life prompt sets is performed on 8 GPUs with a batch size of 96, which takes about 8 hours and 60 hours, respectively. 
Training on Portraits uses 4 GPUs with a batch size of 48, taking about 3.5 hours. 
For inference, given a new text prompt, the generation of the 3D triplane takes about \textcolor{black}{25ms}, and the rendering of one view image at a resolution of $256\times256$ takes about 0.5s.

\begin{figure*}
    \centering
    \setlength{\tabcolsep}{1pt}
    {\scriptsize
    \begin{tabular}{c c c c c c c c}
        \raisebox{0.06\textwidth}{\rotatebox[origin=t]{90}{\scalebox{0.9}{TextMesh}}}& 
        \includegraphics[width=0.13\linewidth, height=0.13\linewidth]{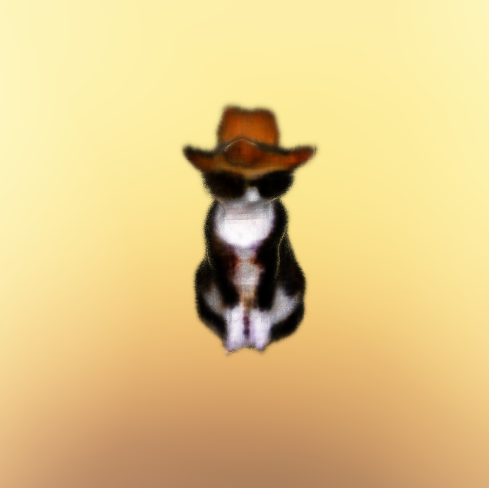}&
        \includegraphics[width=0.13\linewidth, height=0.13\linewidth]{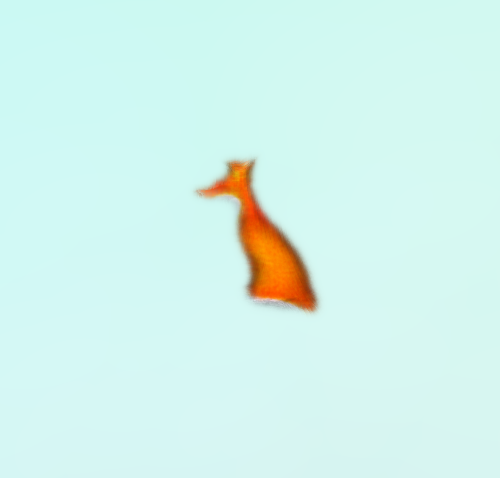}&
        \includegraphics[width=0.13\linewidth, height=0.13\linewidth]{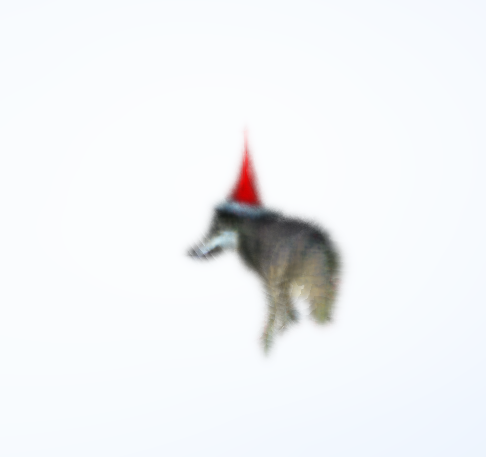}&
        \includegraphics[width=0.13\linewidth, height=0.13\linewidth]{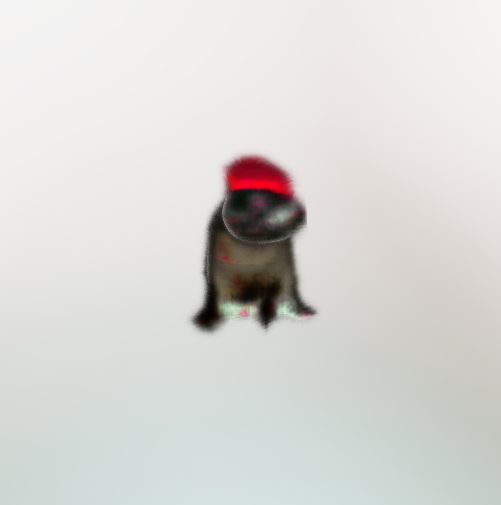}&
        \includegraphics[width=0.13\linewidth, height=0.13\linewidth]{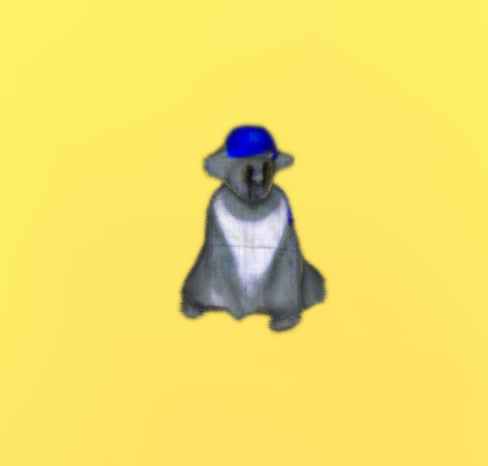}&
        \includegraphics[width=0.13\linewidth, height=0.13\linewidth]{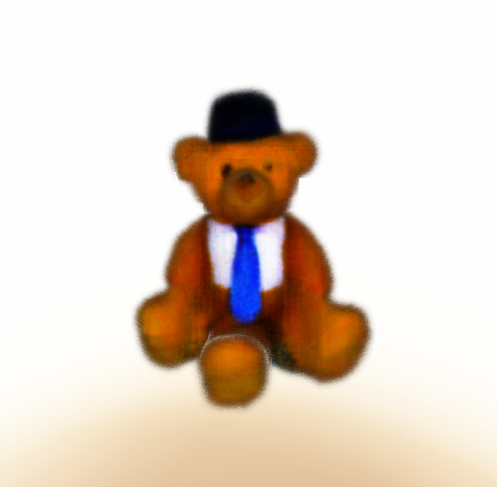}&
        \includegraphics[width=0.13\linewidth, height=0.13\linewidth]{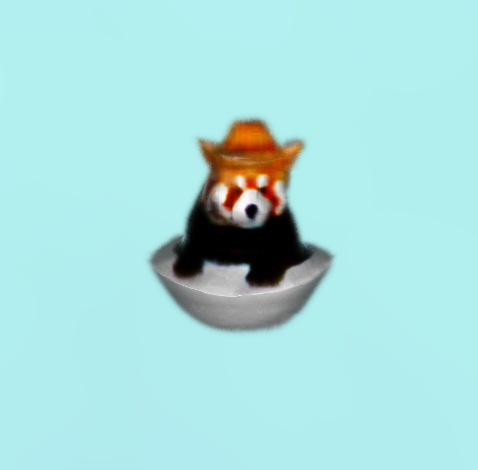}
        \\
        \raisebox{0.06\textwidth}{\rotatebox[origin=t]{90}{\scalebox{0.9}{SJC}}}& 
        \includegraphics[width=0.13\linewidth, height=0.13\linewidth]{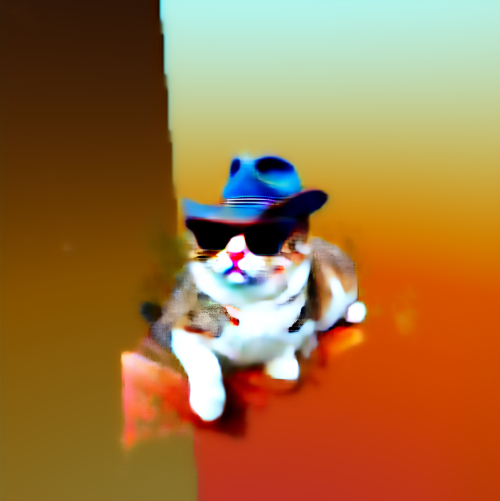}&
        \includegraphics[width=0.13\linewidth, height=0.13\linewidth]{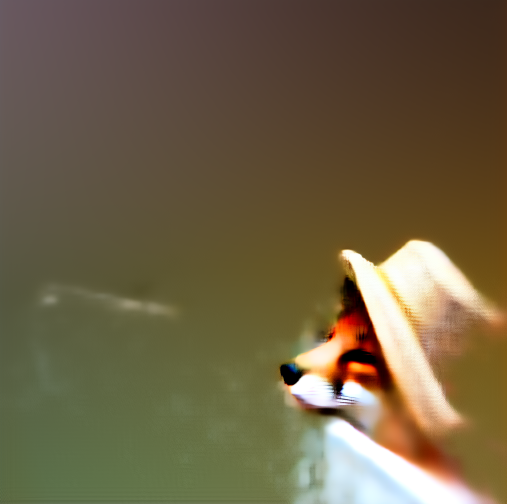}&
        \includegraphics[width=0.13\linewidth, height=0.13\linewidth]{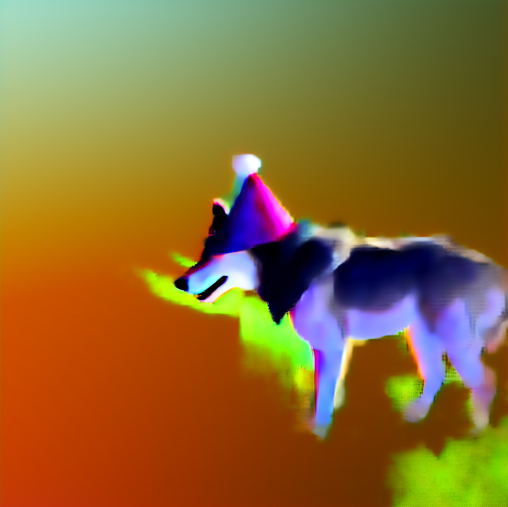}&
        \includegraphics[width=0.13\linewidth, height=0.13\linewidth]{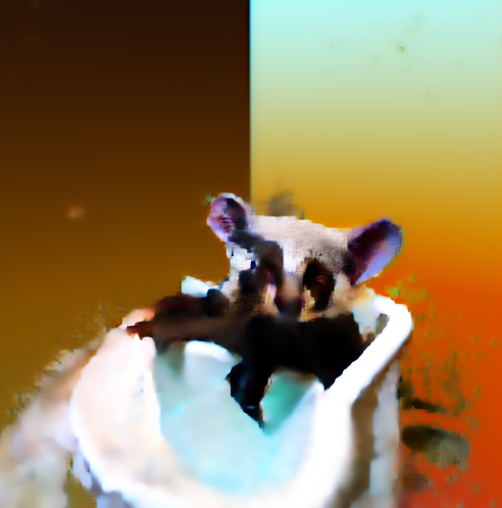}&
        \includegraphics[width=0.13\linewidth, height=0.13\linewidth]{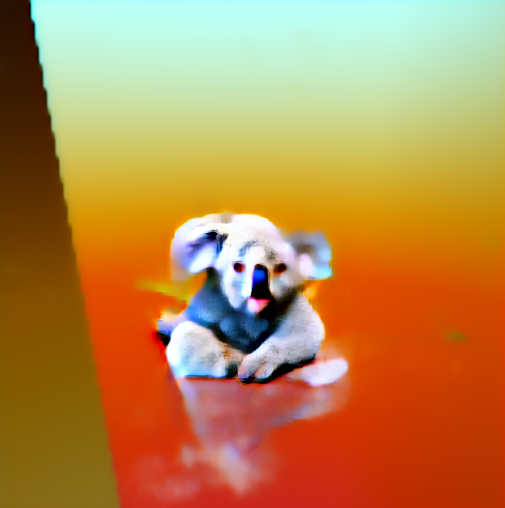}&
        \includegraphics[width=0.13\linewidth, height=0.13\linewidth]{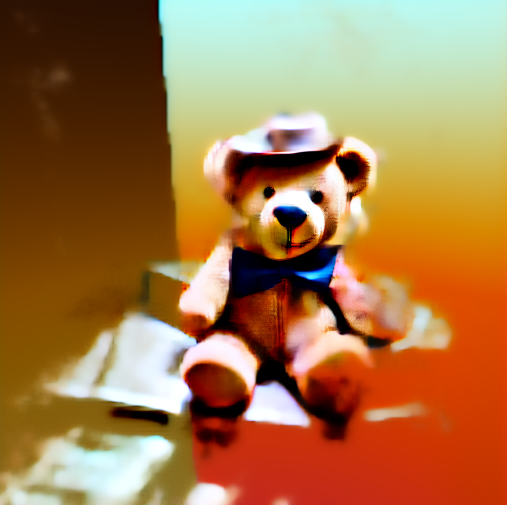}&
        \includegraphics[width=0.13\linewidth, height=0.13\linewidth]{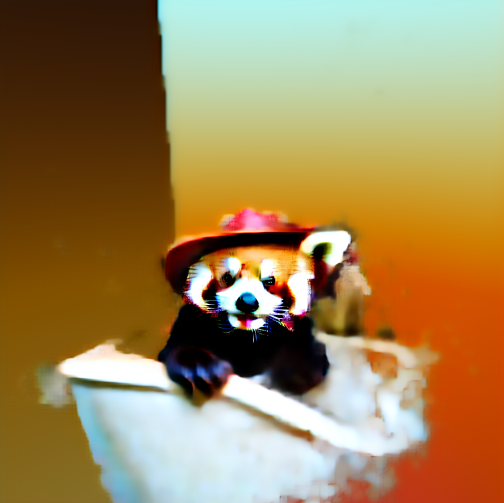}
        \\
        \raisebox{0.06\textwidth}{\rotatebox[origin=t]{90}{\scalebox{0.9}{DreamFusion}}}& 
        \includegraphics[width=0.13\linewidth, height=0.13\linewidth]{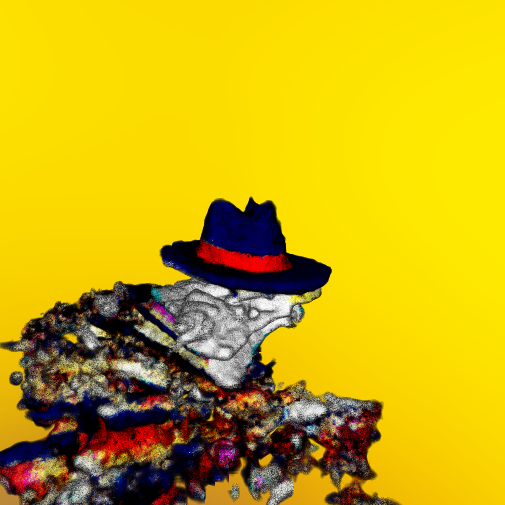}&
        \includegraphics[width=0.13\linewidth, height=0.13\linewidth]{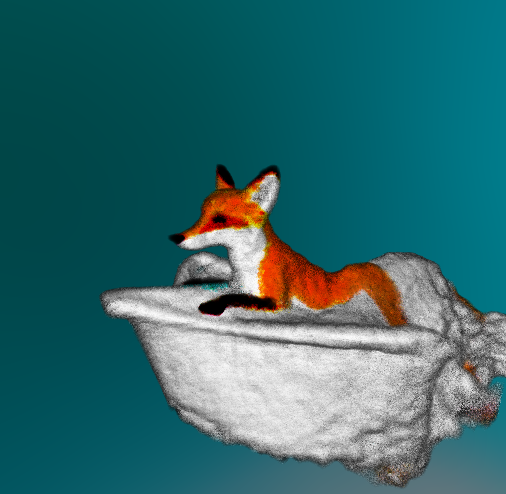}&
        \includegraphics[width=0.13\linewidth, height=0.13\linewidth]{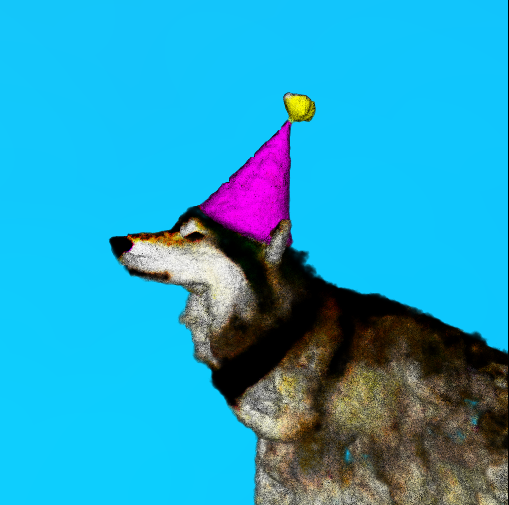}&
        \includegraphics[width=0.13\linewidth, height=0.13\linewidth]{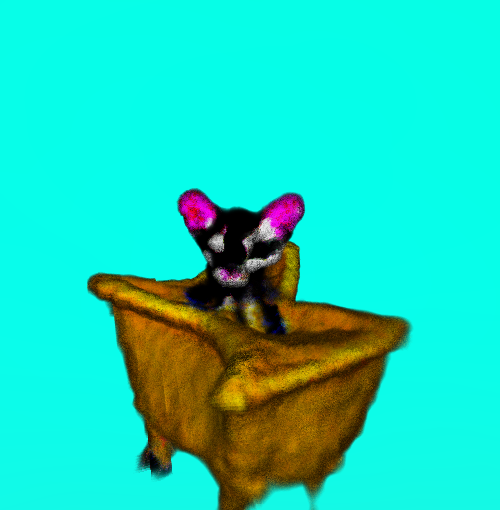}&
        \includegraphics[width=0.13\linewidth, height=0.13\linewidth]{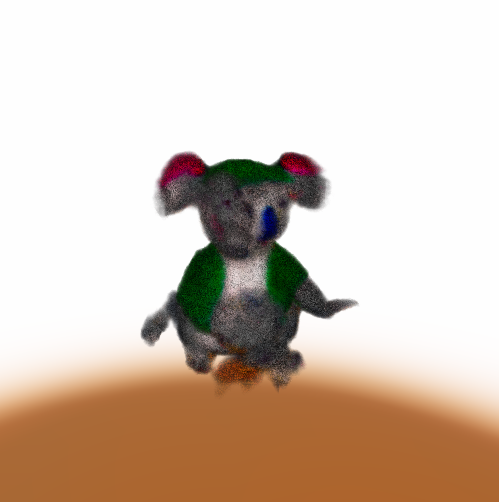}&
        \includegraphics[width=0.13\linewidth, height=0.13\linewidth]{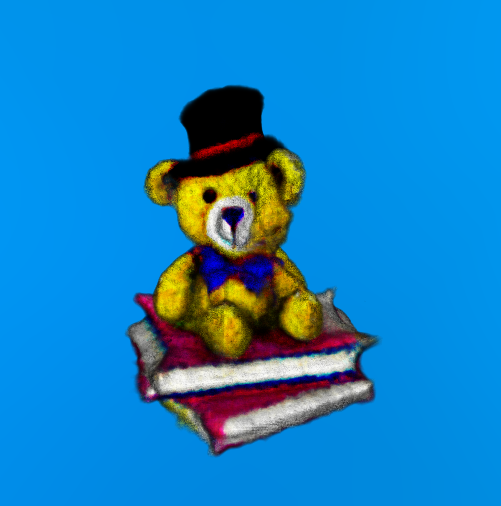}&
        \includegraphics[width=0.13\linewidth, height=0.13\linewidth]{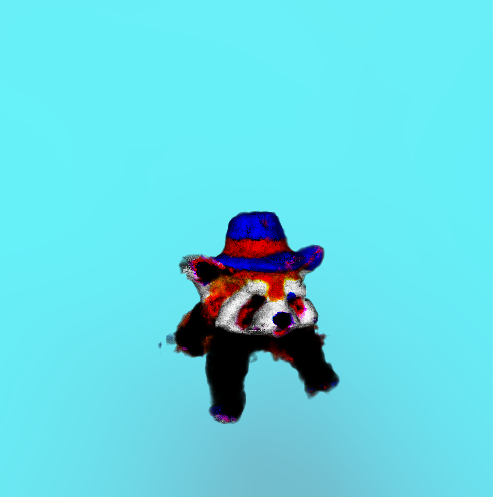}
        \\
        \raisebox{0.06\textwidth}{\rotatebox[origin=t]{90}{\scalebox{0.9}{Latent-NeRF}}}& 
        \includegraphics[width=0.13\linewidth, height=0.13\linewidth]{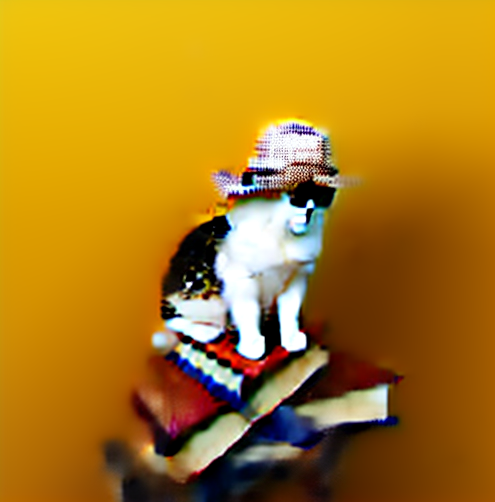}&
        \includegraphics[width=0.13\linewidth, height=0.13\linewidth]{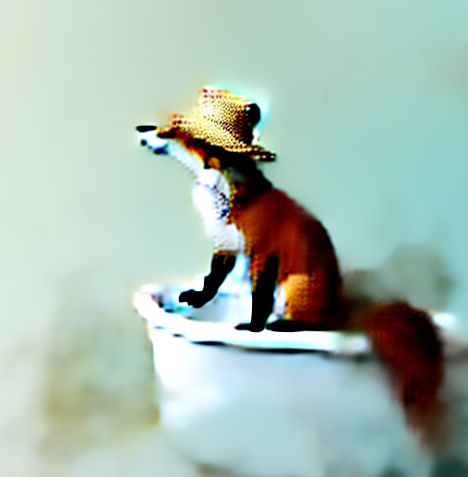}&
        \includegraphics[width=0.13\linewidth, height=0.13\linewidth]{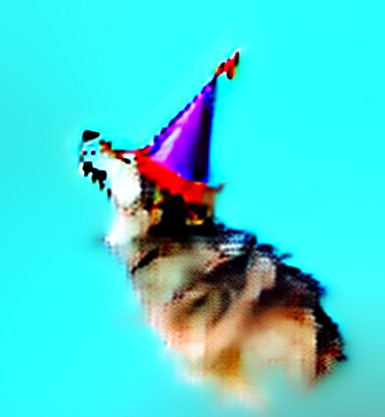}&
        \includegraphics[width=0.13\linewidth, height=0.13\linewidth]{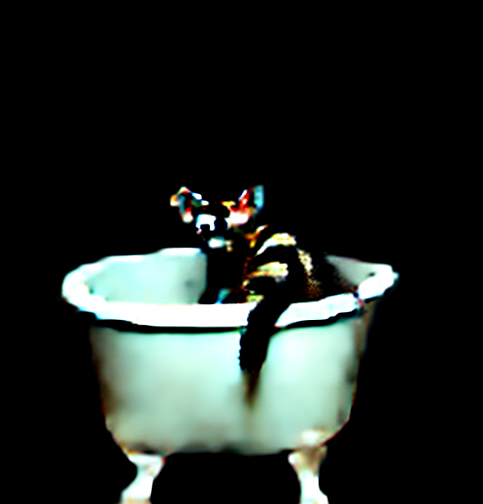}&
        \includegraphics[width=0.13\linewidth, height=0.13\linewidth]{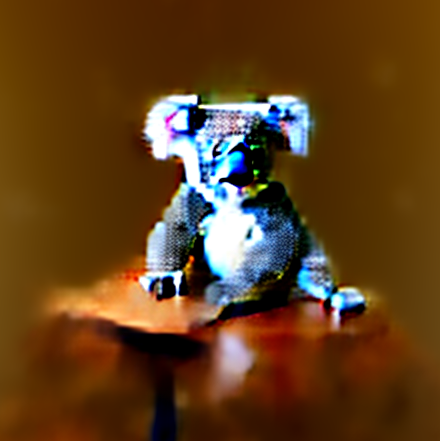}&
        \includegraphics[width=0.13\linewidth, height=0.13\linewidth]{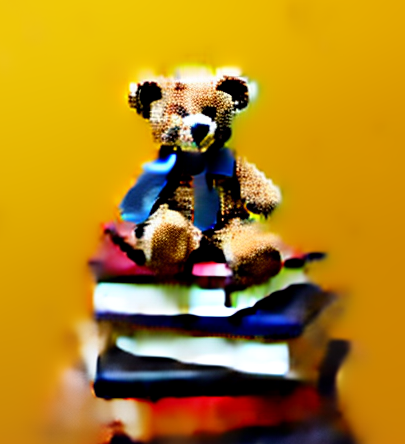}&
        \includegraphics[width=0.13\linewidth, height=0.13\linewidth]{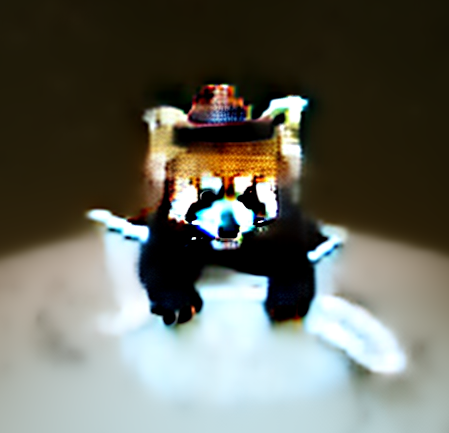}
        \\
        \raisebox{0.06\textwidth}{\rotatebox[origin=t]{90}{\scalebox{0.9}{ProlificDreamer}}}& 
        \includegraphics[width=0.13\linewidth, height=0.13\linewidth]{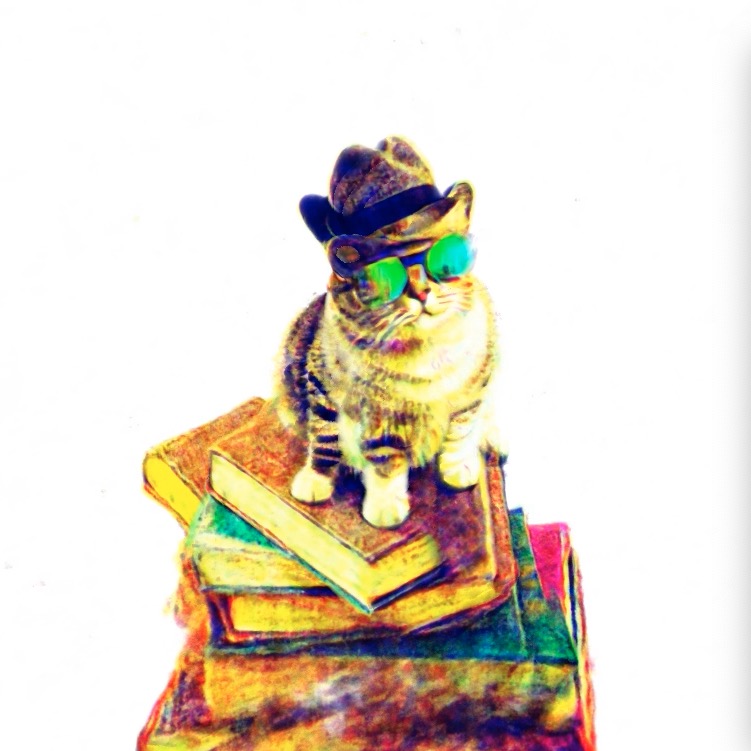}&
        \includegraphics[width=0.13\linewidth, height=0.13\linewidth]{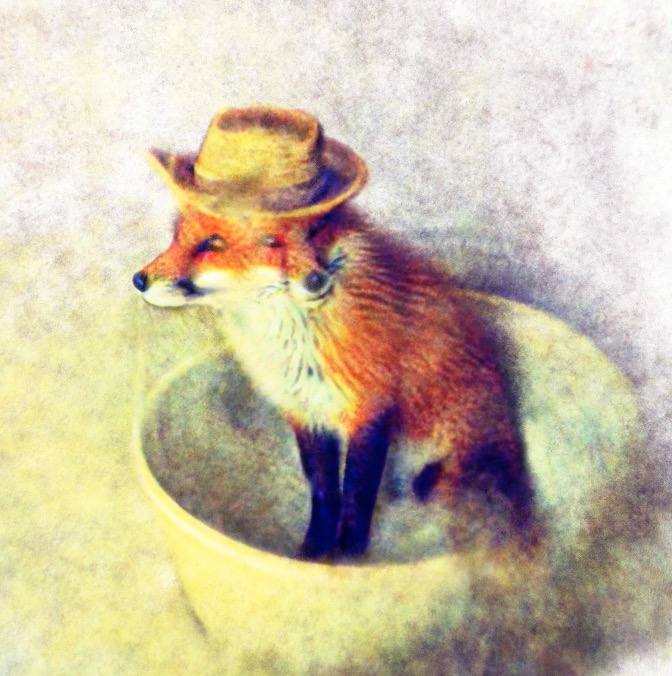}&
        \includegraphics[width=0.13\linewidth, height=0.13\linewidth]{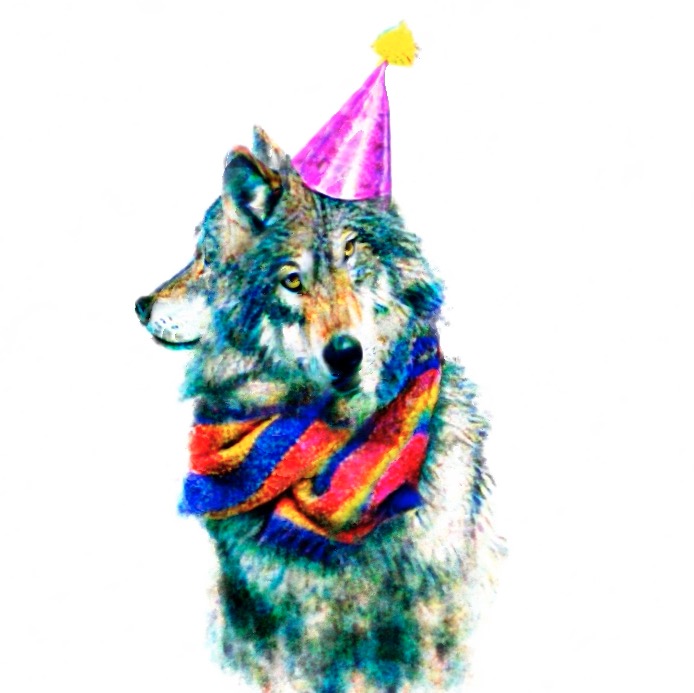}&
        \includegraphics[width=0.13\linewidth, height=0.13\linewidth]{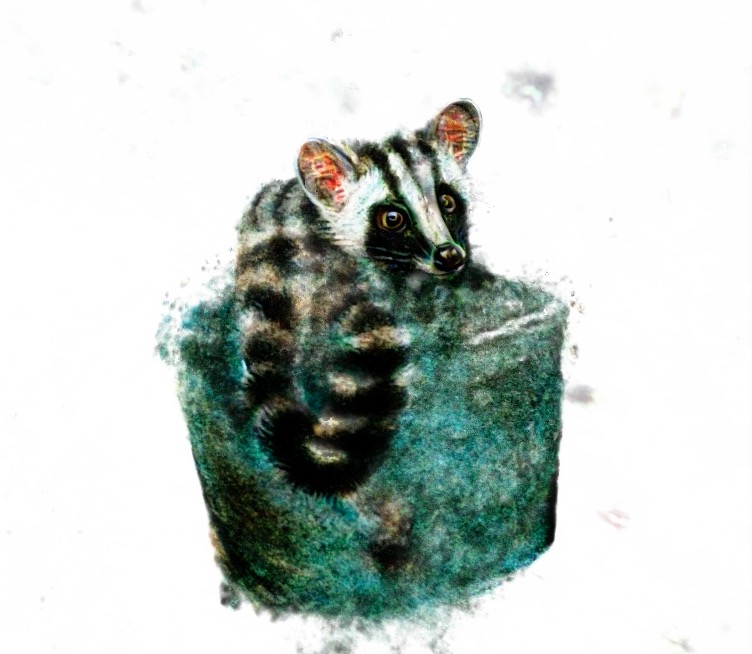}&
        \includegraphics[width=0.13\linewidth, height=0.13\linewidth]{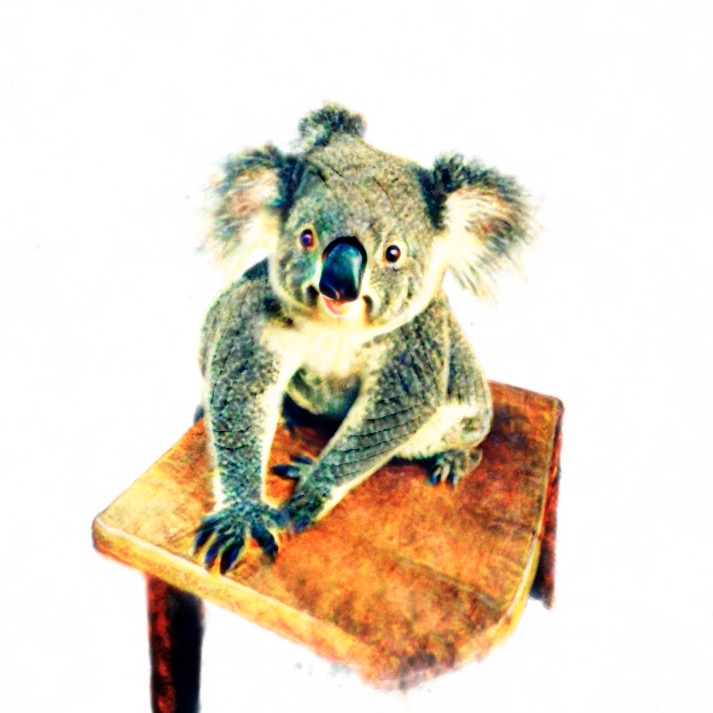}&
        \includegraphics[width=0.13\linewidth, height=0.13\linewidth]{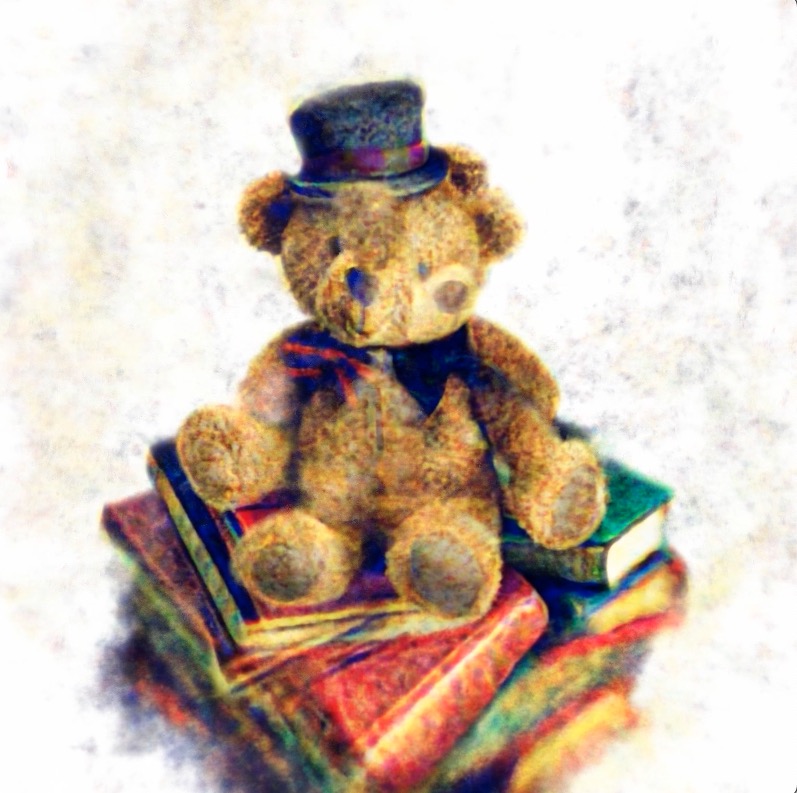}&
        \includegraphics[width=0.13\linewidth, height=0.13\linewidth]{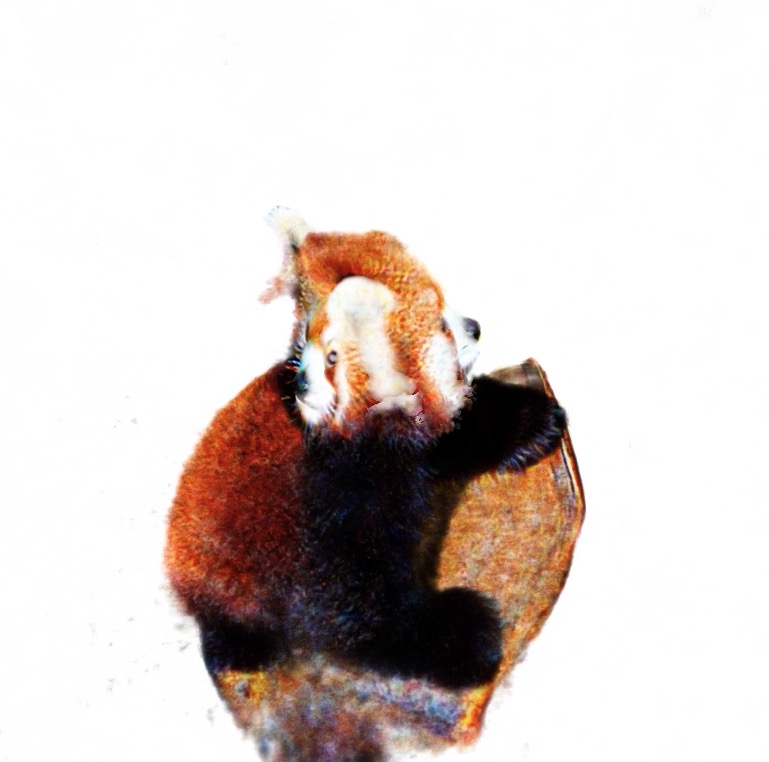}
        \\
        \raisebox{0.06\textwidth}{\rotatebox[origin=t]{90}{\scalebox{0.9}{\name}}}& 
        \includegraphics[width=0.13\linewidth, height=0.13\linewidth]{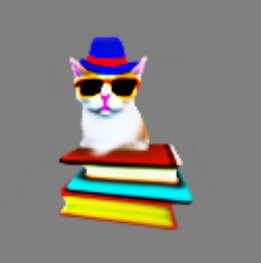}&
        \includegraphics[width=0.13\linewidth, height=0.13\linewidth]{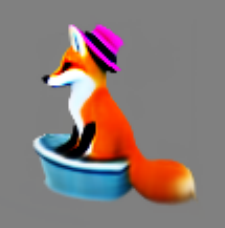}&
        \includegraphics[width=0.13\linewidth, height=0.13\linewidth]{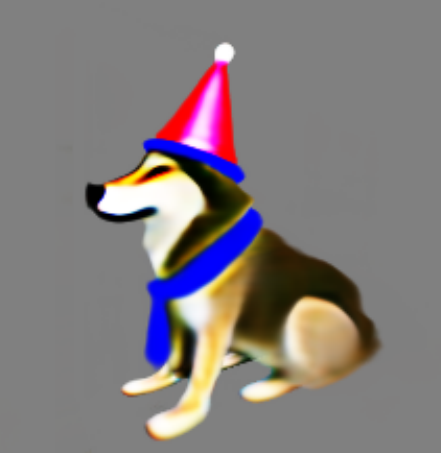}&
        \includegraphics[width=0.13\linewidth, height=0.13\linewidth]{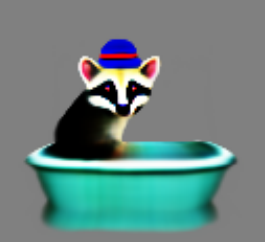}&
        \includegraphics[width=0.13\linewidth, height=0.13\linewidth]{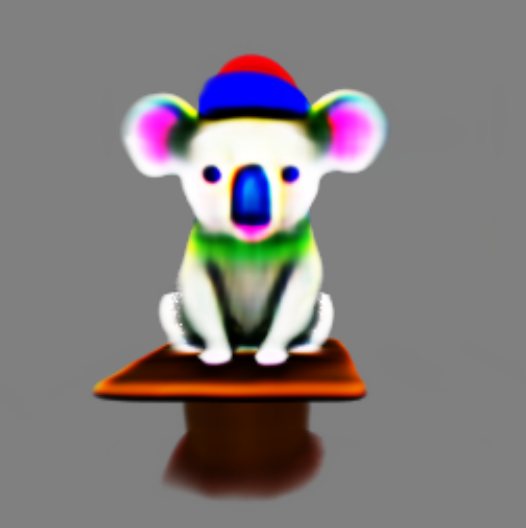}&
        \includegraphics[width=0.13\linewidth, height=0.13\linewidth]{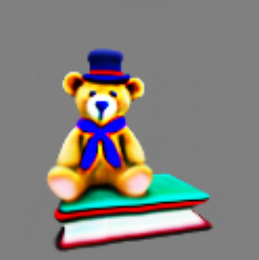}&
        \includegraphics[width=0.13\linewidth, height=0.13\linewidth]{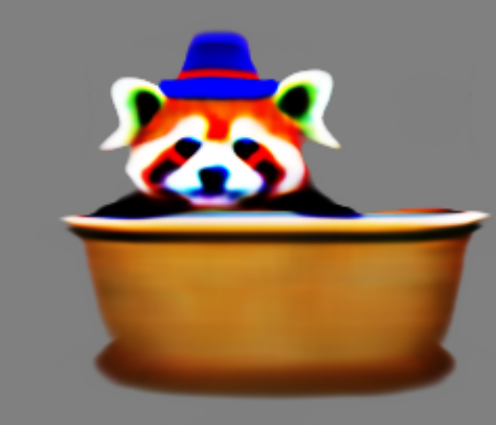}
        \\
        \midrule
        \raisebox{0.06\textwidth}{\rotatebox[origin=t]{90}{\scalebox{0.9}{Point-E}}}& 
        \includegraphics[width=0.13\linewidth, height=0.13\linewidth]{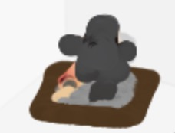}&
        \includegraphics[width=0.13\linewidth, height=0.13\linewidth]{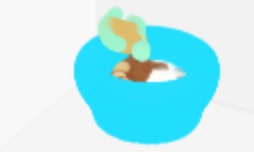}&
        \includegraphics[width=0.13\linewidth, height=0.13\linewidth]{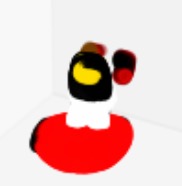}&
        \includegraphics[width=0.13\linewidth, height=0.13\linewidth]{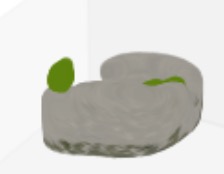}&
        \includegraphics[width=0.13\linewidth, height=0.13\linewidth]{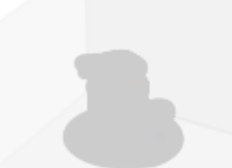}&
        \includegraphics[width=0.13\linewidth, height=0.13\linewidth]{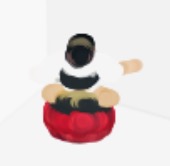}&
        \includegraphics[width=0.13\linewidth, height=0.13\linewidth]{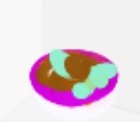}
        \\
    \end{tabular}
    \begin{tabular*}{0.9\linewidth}{P{0.13\linewidth}P{0.13\linewidth}P{0.13\linewidth}P{0.13\linewidth}P{0.13\linewidth}P{0.13\linewidth}P{0.13\linewidth}@{}}
    \centering
    ``a cat sitting on books and wearing sunglasses and wearing a cowboy hat''&
    ``a fox sitting in a bathtub and wearing a straw hat'' & ``a wolf wearing a scarf and wearing a party hat''&``a civet sitting in a bathtub and wearing a beanie''&
    ``a koala sitting on a table and wearing a cape and wearing a baseball cap'' & ``a teddy bear sitting on books and wearing a tie and wearing a tophat'' & ``a red panda sitting in a bathtub and wearing a cowboy hat''
    \end{tabular*}
    }
    
    \caption{\textcolor{black}{
    Qualitative comparison with the state-of-the-art methods including TextMesh \citep{tsalicoglou2023textmesh}, SJC \citep{wang2023score}, DreamFusion \citep{poole2022dreamfusion}, Latent-NeRF \citep{metzer2022latent}, ProlificDreamer \citep{wang2023prolificdreamer}, and Point-E \citep{nichol2022point} on the Animals dataset. 
    The proposed \name{} achieves higher-quality results while being much more efficient than the baselines.}
    }
    \label{fig:comparison_with_sota}
\end{figure*} 

\begin{figure*}
    \centering
    \setlength{\tabcolsep}{1pt}
    {\scriptsize
    \begin{tabular}{c c c c c c}
        \\
        \includegraphics[width=0.13\linewidth, height=0.13\linewidth]{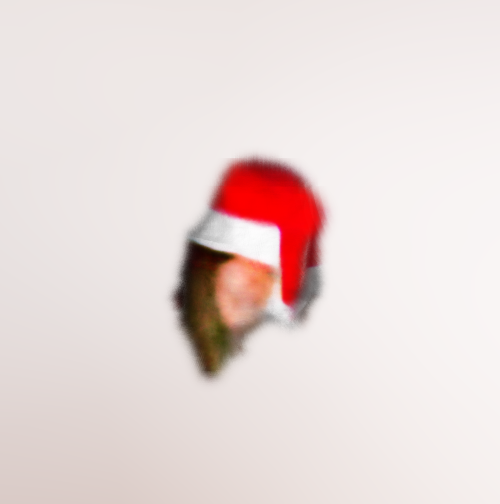}&
        \includegraphics[width=0.13\linewidth, height=0.13\linewidth]{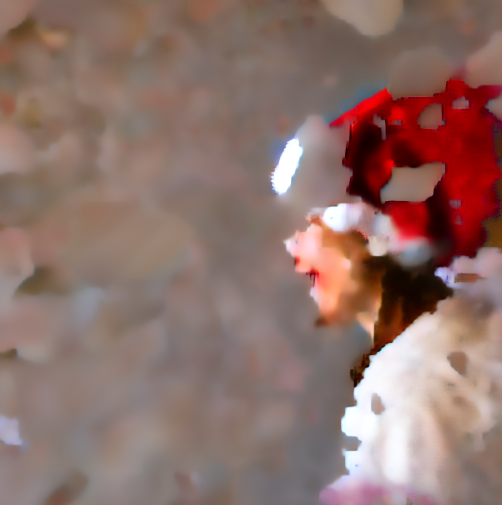}&
        \includegraphics[width=0.13\linewidth, height=0.13\linewidth]{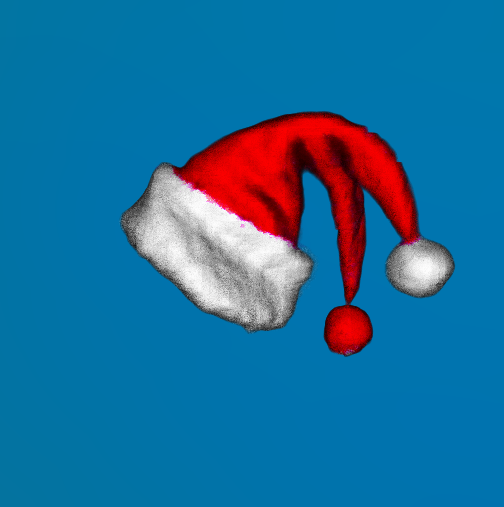}&
        \includegraphics[width=0.13\linewidth, height=0.13\linewidth]{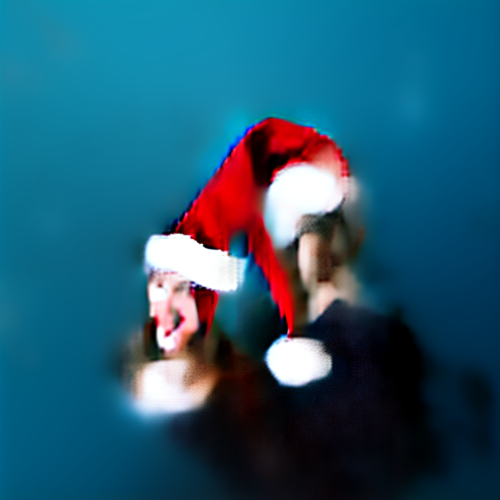}&
        \includegraphics[width=0.13\linewidth, height=0.13\linewidth]{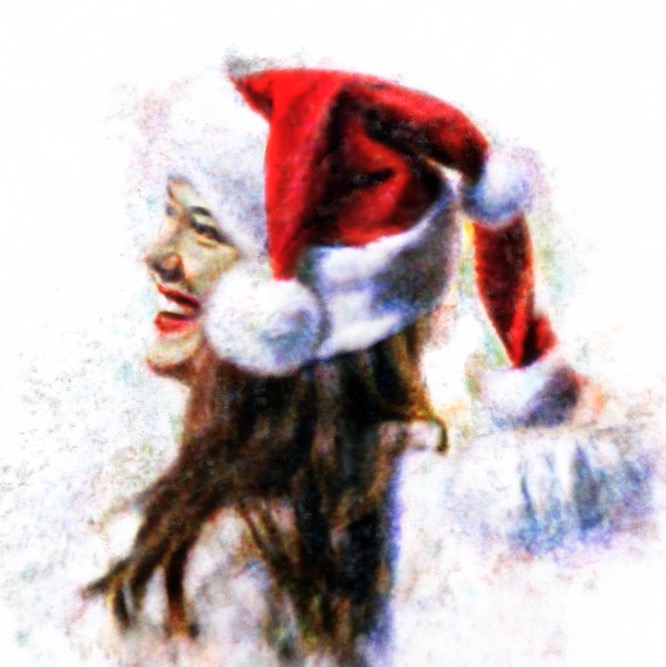}&
        \includegraphics[width=0.13\linewidth, height=0.13\linewidth]{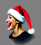}
        \\
        \includegraphics[width=0.13\linewidth, height=0.13\linewidth]{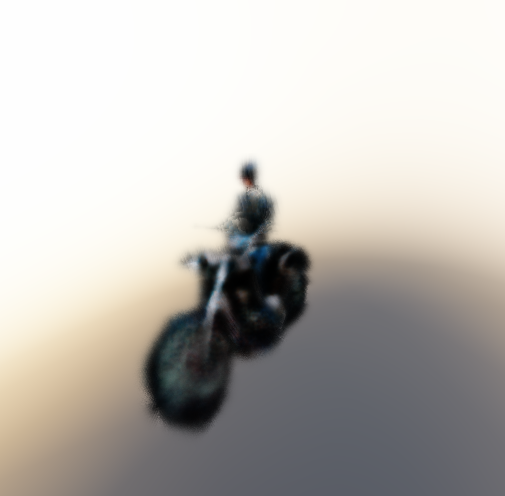}&
        \includegraphics[width=0.13\linewidth, height=0.13\linewidth]{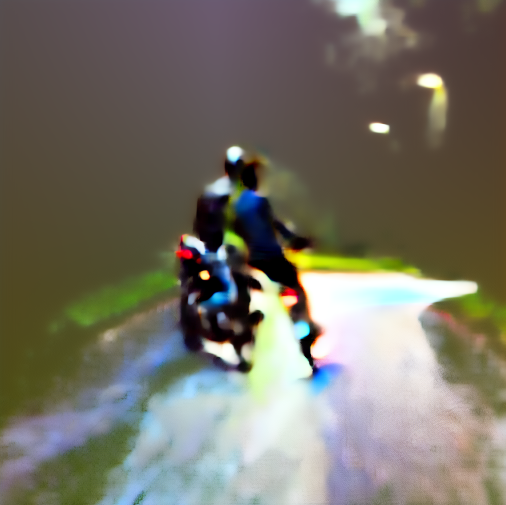}&
        \includegraphics[width=0.13\linewidth, height=0.13\linewidth]{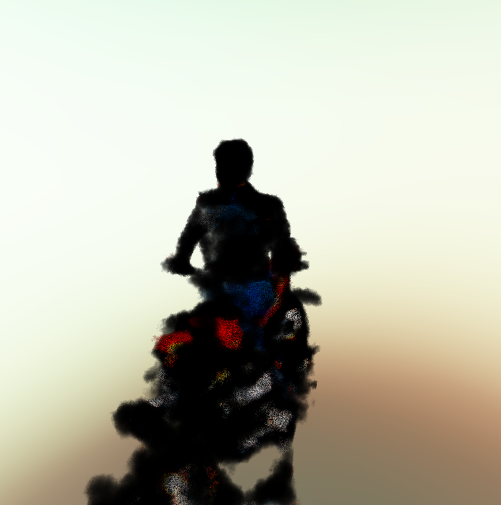}&
        \includegraphics[width=0.13\linewidth, height=0.13\linewidth]{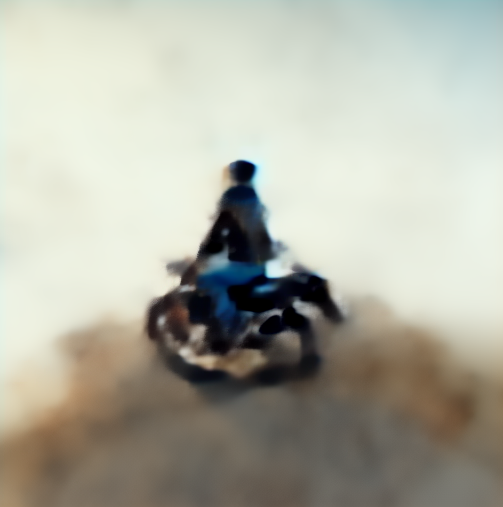}&
        \includegraphics[width=0.13\linewidth, height=0.13\linewidth]{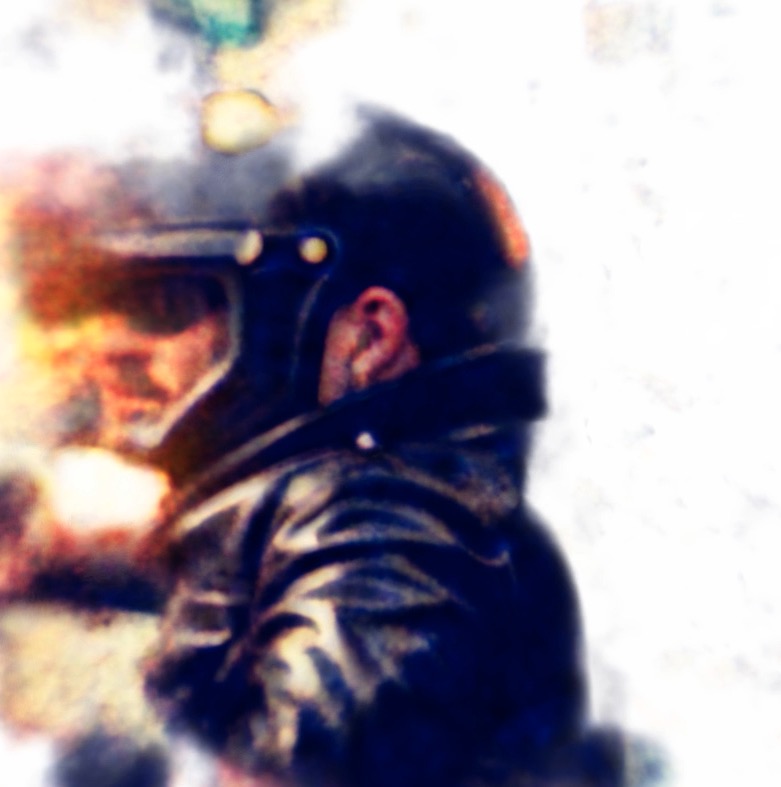}&
        \includegraphics[width=0.13\linewidth, height=0.13\linewidth]{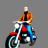}
        \\
        TextMesh & SJC &DreamFusion &Latent-NeRF &ProlificDreamer &\name{}
    \end{tabular}
    }
    \caption{\textcolor{black}{
    Qualitative comparison against the baseline approaches on the Portraits (top) and Daily Life (bottom) datasets. 
    The prompts for the top and bottom are ``a white woman wearing a Santa hat is grinning'' and ``a handsome man wearing a leather jacket is riding a motorcycle'', respectively.}
    }
    \label{fig:comparison_portraits}
\end{figure*}

\begin{figure*}[!t]
\setlength\abovecaptionskip{2mm}
\centering
\begin{subfigure}{0.99\linewidth}
    \centering
    \includegraphics[width=0.99\linewidth, height=0.14\linewidth]{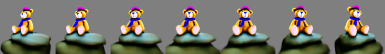}
    \vspace{-1.5mm}
    \caption{``a teddy bear sitting on a stone and wearing a scarf  and wearing a baseball cap''}
\end{subfigure}

\begin{subfigure}{0.99\linewidth}
    \centering
    \includegraphics[width=0.99\linewidth, height=0.14\linewidth]{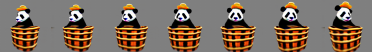}
    \vspace{-1.5mm}
    \caption{``a panda sitting in a basket and wearing a straw hat''}
\end{subfigure}
\caption{Visual results on the Animals set, which are inferred by our \name{} for novel prompts. 
The results demonstrate accurate text-3D alignment and satisfying multi-view consistency.
}
\label{fig:more_results_animals_set1}
\end{figure*}

\begin{figure*}[!t]
\setlength\abovecaptionskip{2mm}
\centering
\begin{subfigure}{0.99\linewidth}
    \centering
    \includegraphics[width=0.99\linewidth, height=0.12\linewidth]{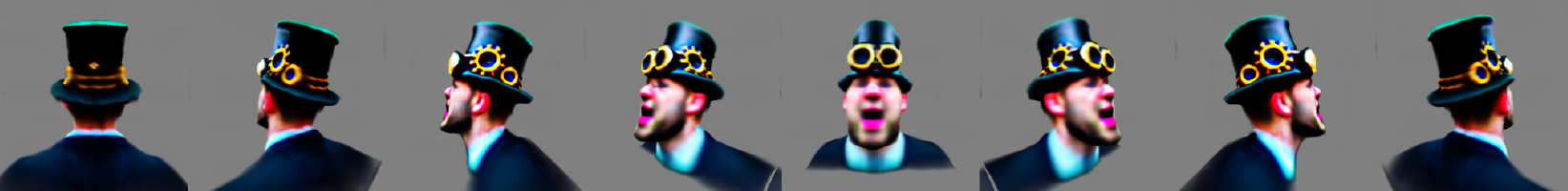}
\end{subfigure}
\begin{subfigure}{0.99\linewidth}
    \centering
    \caption{``a white man wearing a steampunk hat is singing''}
\end{subfigure}

\begin{subfigure}{0.99\linewidth}
    \centering
    \includegraphics[width=0.99\linewidth, height=0.12\linewidth]{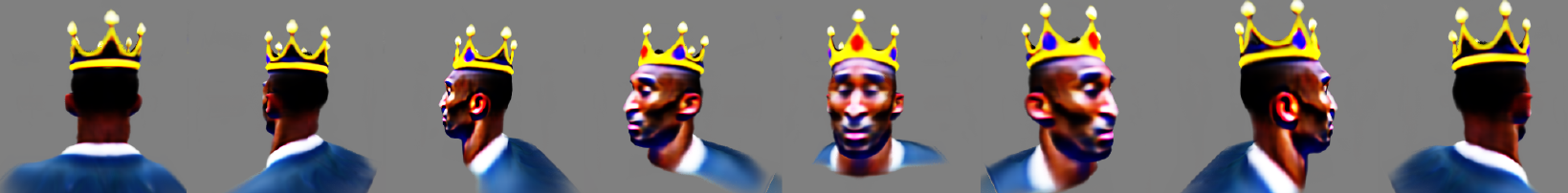}
\end{subfigure}
\begin{subfigure}{0.99\linewidth}
    \centering
    \caption{``Kobe wearing a crown is crying''}
\end{subfigure}

\caption{Visual results on the Portraits set. Our \name{} accurately generates the figures, items, and expressions described in the text prompts, while showing favorable multi-view consistency.
}
\label{fig:portraits1}
\end{figure*}

\begin{figure*}[!t]
\setlength\abovecaptionskip{2mm}
\centering

\begin{subfigure}{0.5\linewidth}
    \centering
    \includegraphics[width=0.99\linewidth, height=0.2\linewidth]{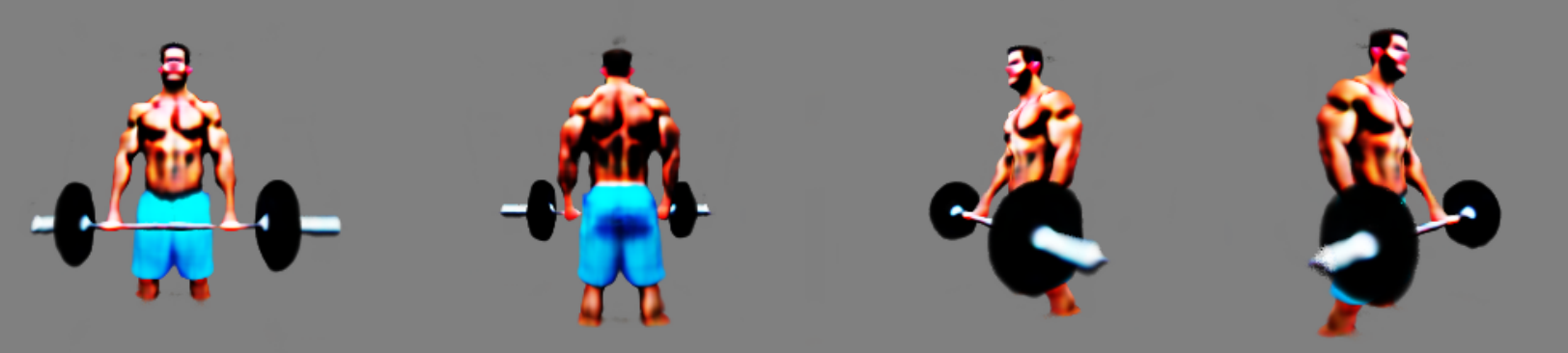}
    \vspace{-1.5mm}
\end{subfigure}
\hspace{-7pt}
\begin{subfigure}{0.5\linewidth}
    \centering
    \includegraphics[width=0.99\linewidth, height=0.2\linewidth]{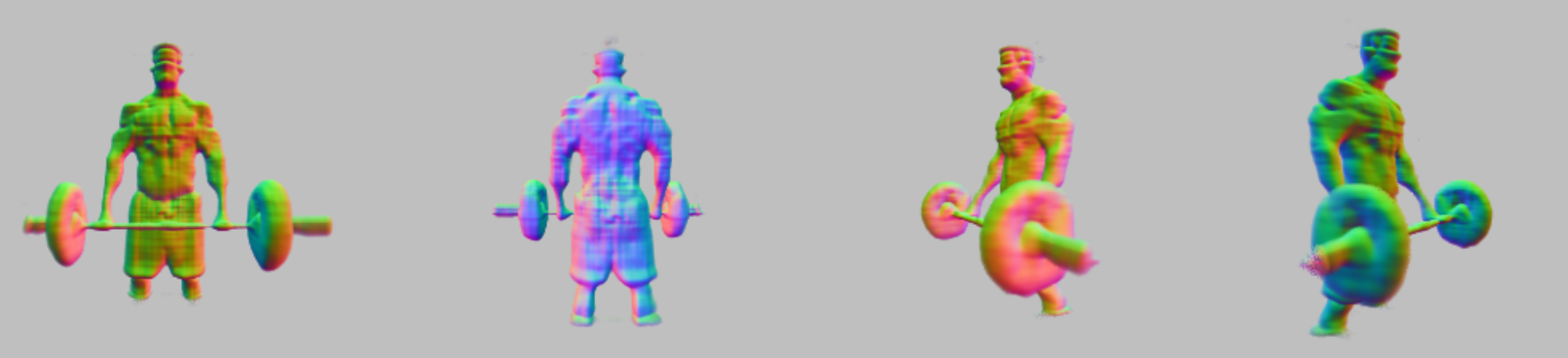}
    \vspace{-1.5mm}
\end{subfigure}
\begin{subfigure}{0.99\linewidth}
    \centering
    \caption{``a muscular man is exercising in a fitness room''}
\end{subfigure}

\begin{subfigure}{0.5\linewidth}
    \centering
    \includegraphics[width=0.99\linewidth, height=0.2\linewidth]{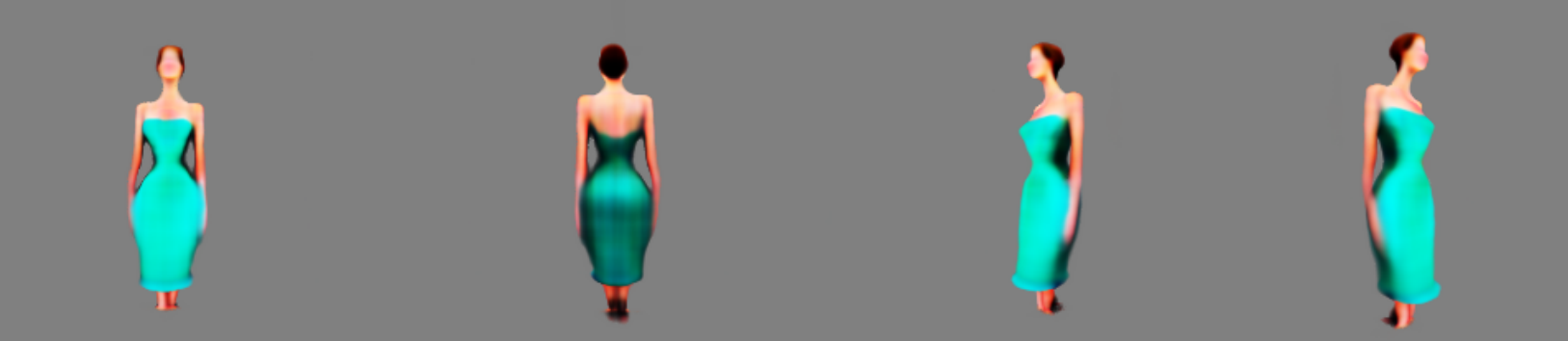}
    \vspace{-1.5mm}
\end{subfigure}
\hspace{-7pt}
\begin{subfigure}{0.5\linewidth}
    \centering
    \includegraphics[width=0.99\linewidth, height=0.2\linewidth]{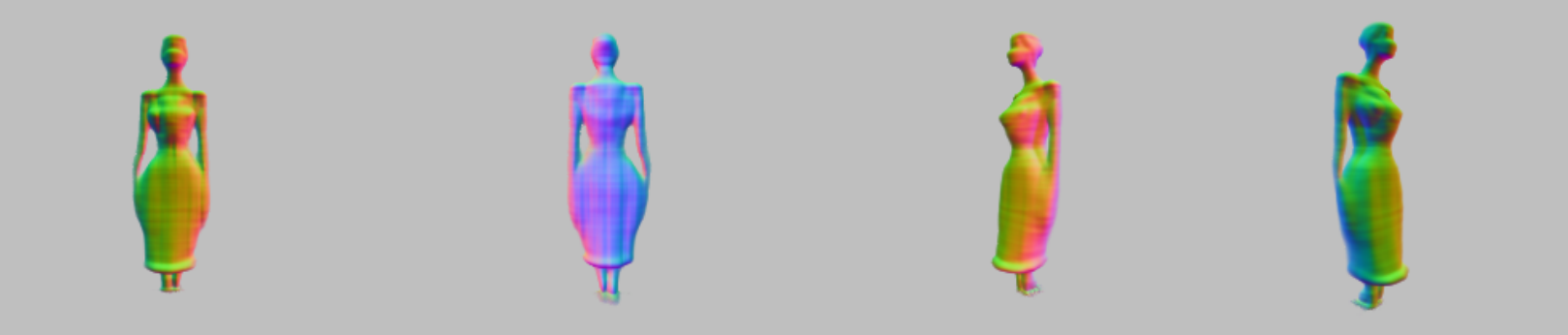}
    \vspace{-1.5mm}
\end{subfigure}
\begin{subfigure}{0.99\linewidth}
    \centering
    \caption{``a slim woman is trying on a dress''}
\end{subfigure}

\begin{subfigure}{0.5\linewidth}
    \centering
    \includegraphics[width=0.99\linewidth, height=0.2\linewidth]{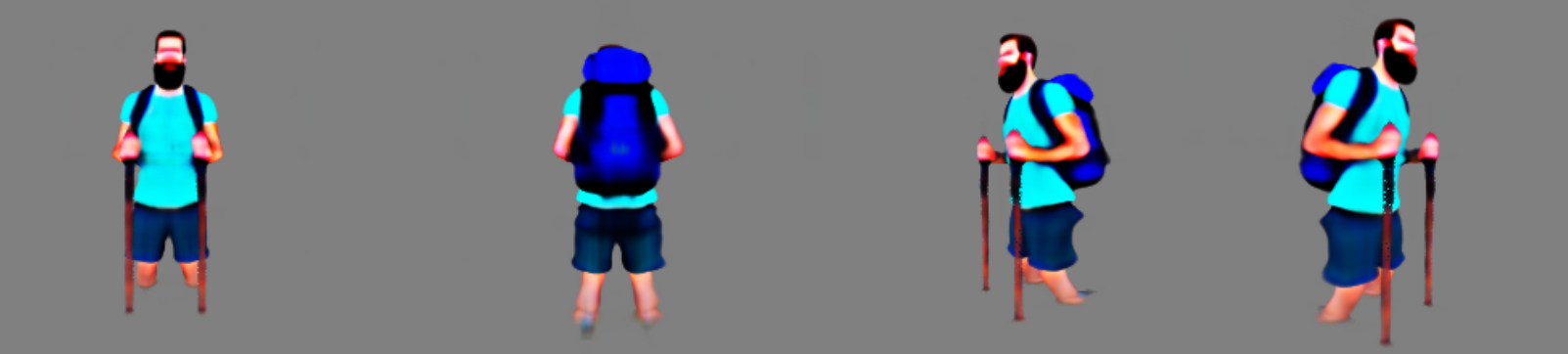}
    \vspace{-1.5mm}
\end{subfigure}
\hspace{-7pt}
\begin{subfigure}{0.5\linewidth}
    \centering
    \includegraphics[width=0.99\linewidth, height=0.2\linewidth]{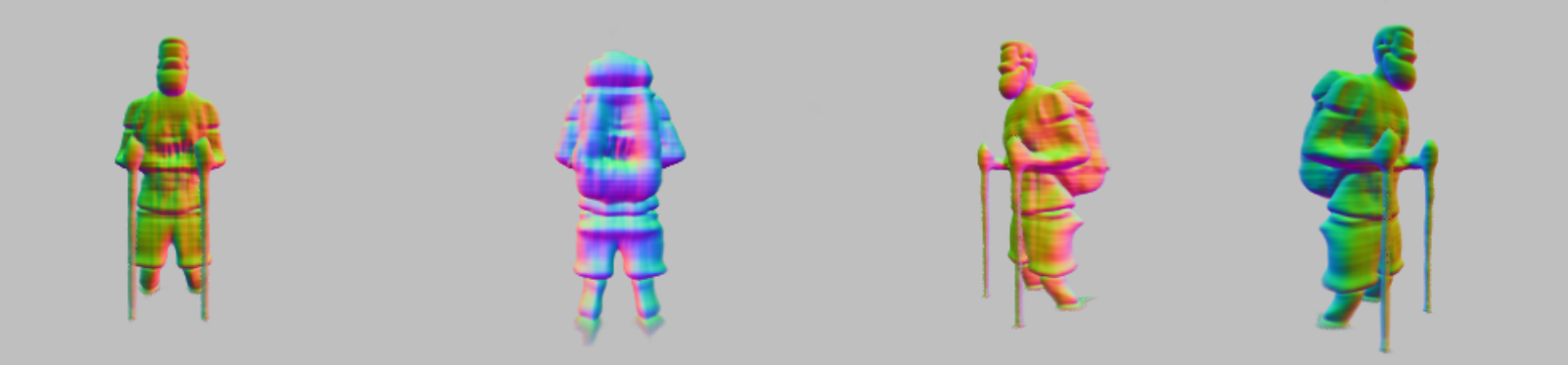}
    \vspace{-1.5mm}
\end{subfigure}
\begin{subfigure}{0.99\linewidth}
    \centering
    \caption{``a bearded man wearing a backpack is climbing a mountain''}
\end{subfigure}

\begin{subfigure}{0.5\linewidth}
    \centering
    \includegraphics[width=0.99\linewidth, height=0.2\linewidth]{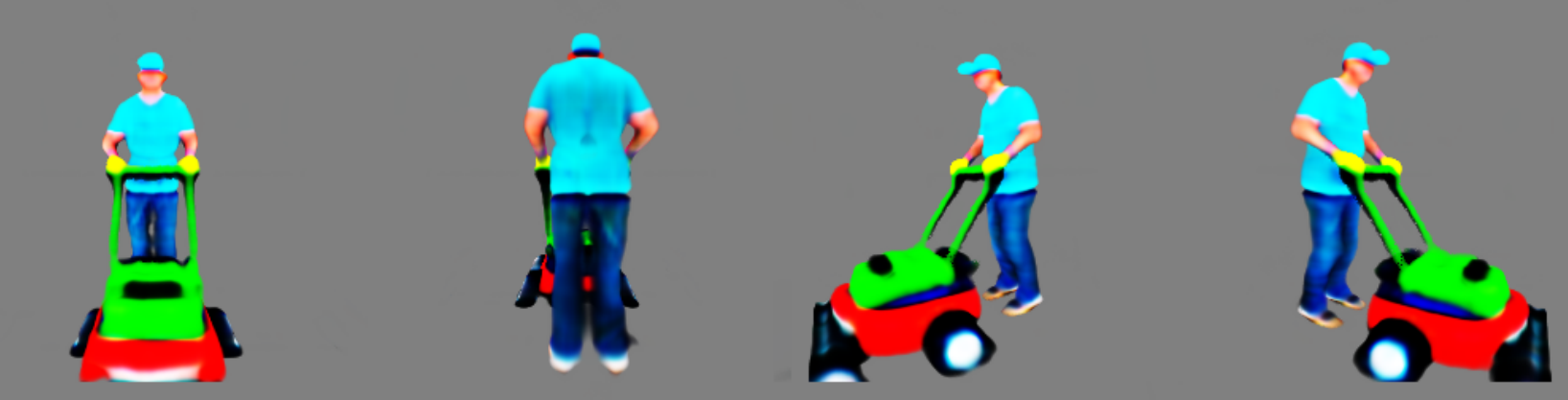}
    \vspace{-1.5mm}
\end{subfigure}
\hspace{-7pt}
\begin{subfigure}{0.5\linewidth}
    \centering
    \includegraphics[width=0.99\linewidth, height=0.2\linewidth]{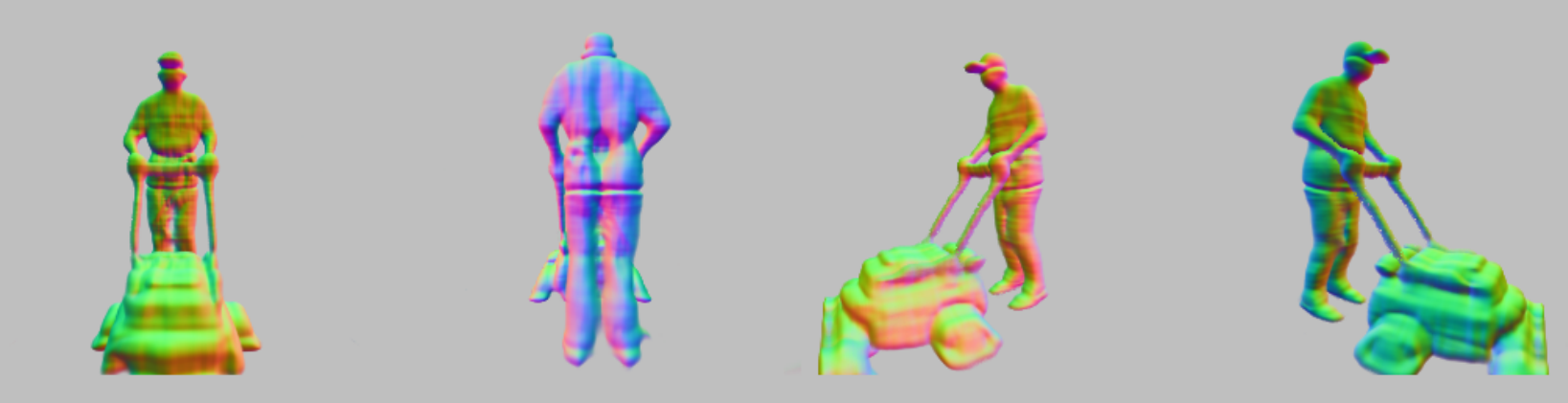}
    \vspace{-1.5mm}
\end{subfigure}
\begin{subfigure}{0.99\linewidth}
    \centering
    \caption{``a man wearing a hat is mowing the lawn''}
\end{subfigure}

\caption{Visual results on the Daily Life set. For each prompt, four views and their normal images are shown, demonstrating that our \name{} is able to generate complex and diverse objects with accurate geometric details for novel prompts. 
}
\label{fig:daily_life1}
\end{figure*}

\subsection{Comparison with the State-of-the-Art Methods}

\bfsection{Qualitative Comparison}
We compare with the state-of-the-art text-to-3D methods including TextMesh \citep{tsalicoglou2023textmesh}, SJC \citep{wang2023score}, DreamFusion \citep{poole2022dreamfusion}, Latent-NeRF \citep{metzer2022latent} and ProlificDreamer \citep{wang2023prolificdreamer}. 
Note that these methods typically take more than an hour to optimize a NeRF for an input prompt, while our framework can generate a 3D object in under a second. 

Meanwhile, since our model is trained on a large set of text inputs, it is able to leverage general 3D priors shared across objects.
Therefore, the generated objects by \name{} have much higher quality than the baselines as illustrated in \fig~\ref{fig:comparison_with_sota} and \ref{fig:comparison_portraits}. 
Besides, our method captures all elements specified in the prompts as well as the relationships between them. 
In contrast, objects produced by the baseline approaches frequently miss important items, and their inter-item relationships can appear disorganized. 

In addition, we provide a comparison with Point-E~\citep{nichol2022point}, which directly trains a 3D diffusion model.
This method is based on a substantially different paradigm and requires millions of 3D-text pairs that are not available to the public.
In contrast, our algorithm does not require any 3D data.
Moreover, due to the limitation of the 3D datasets, especially in terms of scale and diversity, Point-E can only generate simple objects and cannot perform well on more intricate text prompts as shown in \fig~\ref{fig:comparison_with_sota}.
 
Finally, we present more visual examples from the Animals, Portraits, and Daily Life datasets in \fig~\ref{fig:more_results_animals_set1}-\ref{fig:daily_life1}.
The proposed \name{} consistently produces high-quality text-to-3D results with favorable multi-view consistency.
Notably, its performance on the challenging Daily Life dataset (\fig~\ref{fig:daily_life1}) underscores its capability to handle intricate real-world text prompts. 

\begin{figure}[h]
\setlength\abovecaptionskip{2mm}
\centering
\hspace{-2mm}
\begin{tabular}{cc}
\scriptsize
     \includegraphics[width=.47\linewidth]{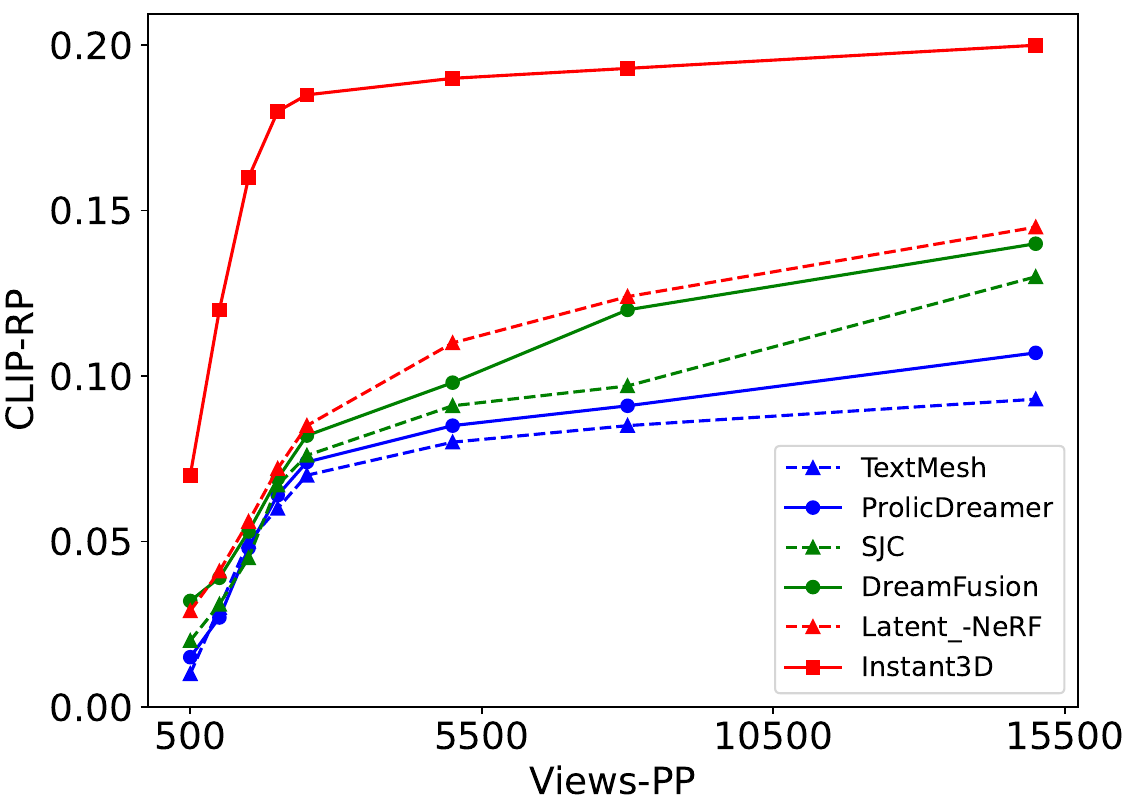} & \includegraphics[width=.5\linewidth]{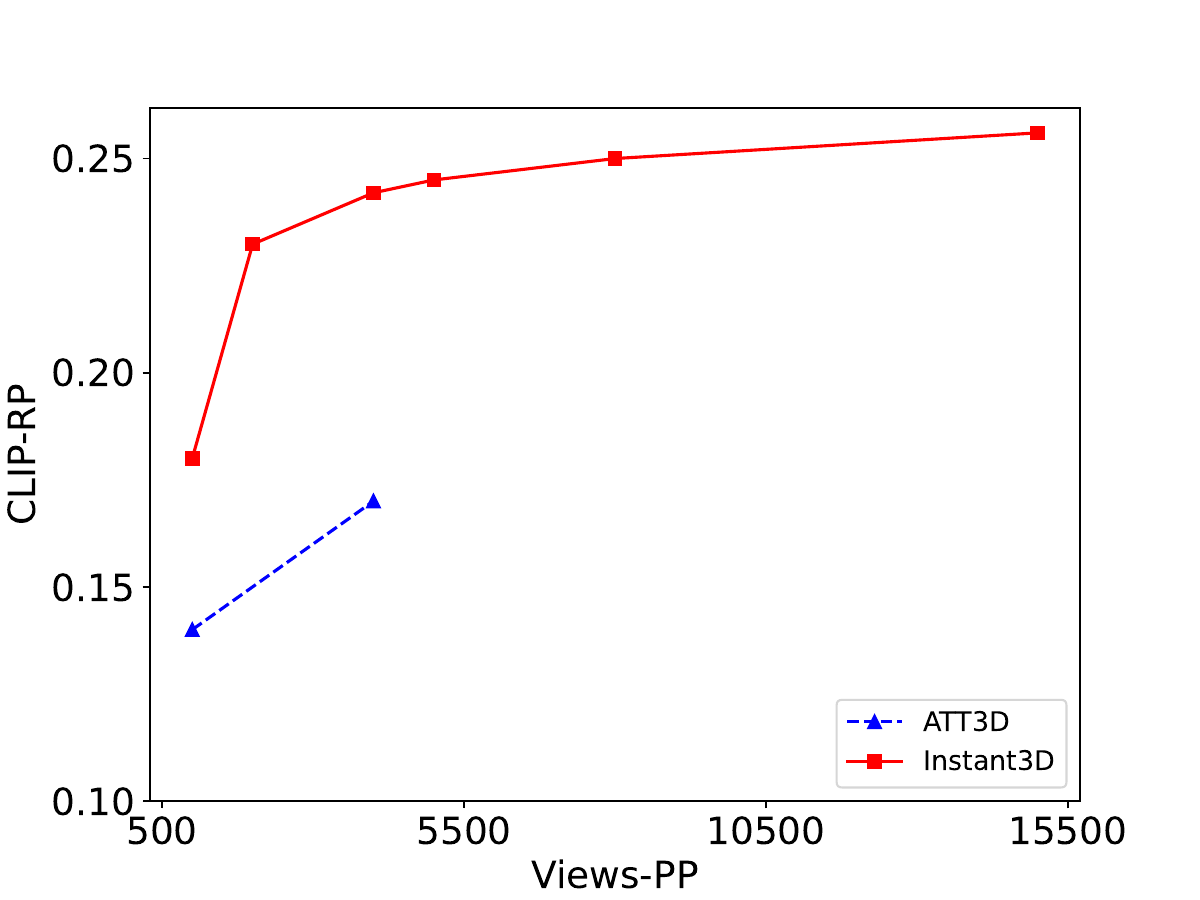} \\
     SOTA & ATT3D
\end{tabular}
\caption{Computation cost comparison against the baseline methods. 
Left: comparison with the state-of-the-art methods on the Animals dataset.
Right: comparison with a concurrent approach ATT3D on their dataset.
Views-PP is the number of rendered images per prompt in training, and CLIP-RP is the retrieval probability computed by CLIP encoders. 
}
\label{fig:comparison_retrieval_rp}
\end{figure}

\bfsection{Computation Costs}
To evaluate the training costs of different methods, we report the number of rendered view images per prompt on average (Views-PP) \vs CLIP retrieval probability (CLIP-RP) on the Animals set. 
Views-PP is computed as:
\begin{equation}
    \centering
    \text{Views-PP} = \frac{\text{Iterations} \times \text{Batch Size}}{\text{Number of Prompts}},
\end{equation}
which essentially measures the computation cost as the number of images rendered for each prompt during training.
CLIP-RP is defined as the average probability of assigning the correct prompt to a rendered image among a set of distraction texts (called the query set).
The similarity between a text and an image is measured by CLIP text and image encoders.
For each prompt, the CLIP-RP is averaged over four view images rendered from distinct camera poses as in DreamFusion~\citep{poole2022dreamfusion}. 
All prompts of the Animals set are used as the query set. 
As shown in \fig~\ref{fig:comparison_retrieval_rp}, the generation quality generally gets better as the training iteration increases, while the improvement speed of the baseline methods is much slower than that of our \name{}. 
Notably, the baselines can only achieve an acceptable CLIP-RP after 10,000 iterations per prompt.
In contrast, our method obtains a much higher CLIP-RP score in only 2,000 iterations per prompt.

\begin{table}
\setlength{\tabcolsep}{1pt}
\centering
\caption{
User study for comparing our method with the baseline approaches in terms of realism and text-3D consistency. 
}
\resizebox{0.99\columnwidth}{!}{
\begin{tabular}{ccccccc} 
\toprule
Method & TextMesh   & SJC & DreamFusion   & Latent-NeRF & ProlificDreamer & \name{} 
 \\ 
\midrule
Realism &4.22\% &4.95\% &6.32\% &7.49\% &1.74\% &75.28\%\\
Consistency &2.35\% &3.91\% &3.69\% &5.10\% &1.33\% &83.62\%\\
\bottomrule
\end{tabular}}
\label{tab:user_study_sota}
\end{table}

\begin{figure*}[h]
    \centering
    \setlength{\tabcolsep}{1pt}
    {\scriptsize
    \begin{tabular}{c c c c c c c }
        \raisebox{0.06\textwidth}{\rotatebox[origin=t]{90}{\scalebox{0.9}{ATT3D}}}& 
        \includegraphics[width=0.13\linewidth, height=0.13\linewidth]{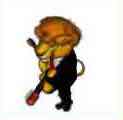}&
        \includegraphics[width=0.13\linewidth, height=0.13\linewidth]{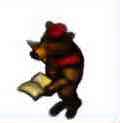}&
        \includegraphics[width=0.13\linewidth, height=0.13\linewidth]{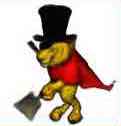}&
        \includegraphics[width=0.13\linewidth, height=0.13\linewidth]{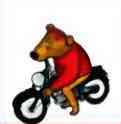}&
        \includegraphics[width=0.13\linewidth, height=0.13\linewidth]{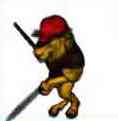}&
        \includegraphics[width=0.13\linewidth, height=0.13\linewidth]{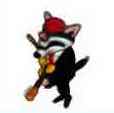}
        \\
        \raisebox{0.06\textwidth}{\rotatebox[origin=t]{90}{\scalebox{0.9}{Instant3D}}}& 
        \includegraphics[width=0.13\linewidth, height=0.13\linewidth]{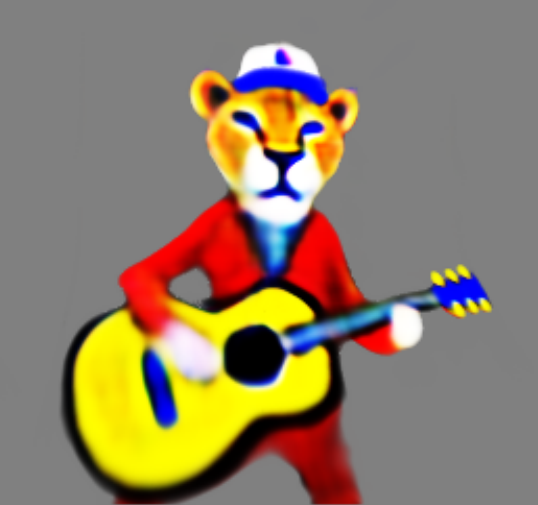}&
        \includegraphics[width=0.13\linewidth, height=0.13\linewidth]{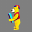}&
        \includegraphics[width=0.13\linewidth, height=0.13\linewidth]{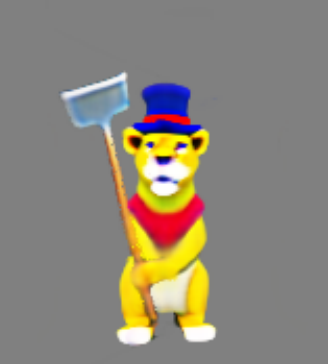}&
        \includegraphics[width=0.13\linewidth, height=0.13\linewidth]{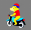}&
        \includegraphics[width=0.13\linewidth, height=0.13\linewidth]{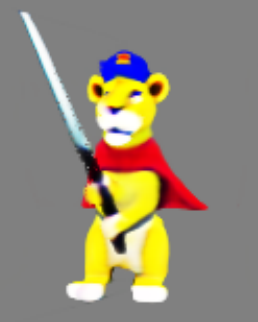}&
        \includegraphics[width=0.13\linewidth, height=0.13\linewidth]{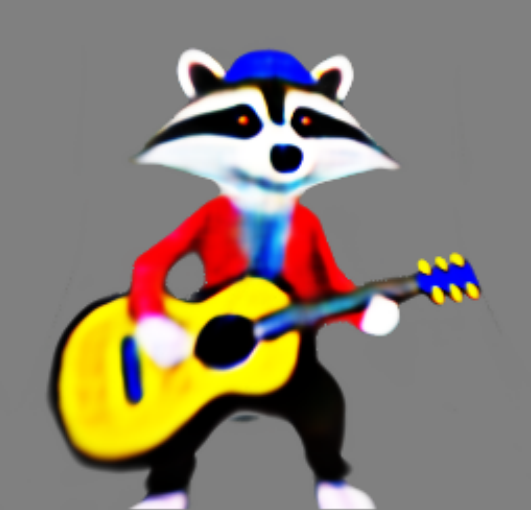}

    \end{tabular}
    \begin{tabular*}{0.9\linewidth}{P{0.05\linewidth}P{0.13\linewidth}P{0.13\linewidth}P{0.13\linewidth}P{0.13\linewidth}P{0.13\linewidth}P{0.13\linewidth}@{}}
    \centering  
    & ``a lion playing the guitar wearing a suit wearing a baseball cap'' & ``a bear holding a book wearing a sweater wearing a baseball cap'' &``a lion holding a shovel wearing a cape wearing a tophat''& ``a bear riding a motorcycle wearing a sweater wearing a baseball cap'' & ``a lion wielding a katana wearing a cape wearing a baseball cap'' & ``a raccoon playing the guitar wearing a suit wearing a baseball cap''
    \end{tabular*}
    }
    
    \caption{Comparison with the concurrent work ATT3D. The text prompts are given in the last row. Our generated 3D objects contain more details and are more distinguishable with better interactions between animals and items. 
    }
    \label{fig:qualitative_comparison_att3d}
\end{figure*}

\bfsection{User Preference Study}
We conduct a user study to evaluate different methods based on subjective perceptual quality assessment.
We show users videos rendered from multiple views of objects, generated by different methods using the same text prompt. 
Users are asked to select the result that most closely matches the description in terms of realism and consistency. 
Each prompt is evaluated by at least $3$ different users. 
As shown in \tab~\ref{tab:user_study_sota}, most users prefer our results to those generated by baselines, highlighting the effectiveness of \name{} in achieving high fidelity and accurate text-3D alignment.

\begin{figure*}
    \centering
    \setlength{\tabcolsep}{1pt}
    {\scriptsize
    \begin{tabular}{c c}
        \raisebox{0.06\textwidth}{\rotatebox[origin=t]{90}{\scalebox{0.9}{w/o Noise}}}& 
        \includegraphics[width=0.9\textwidth, height=0.13\textwidth]{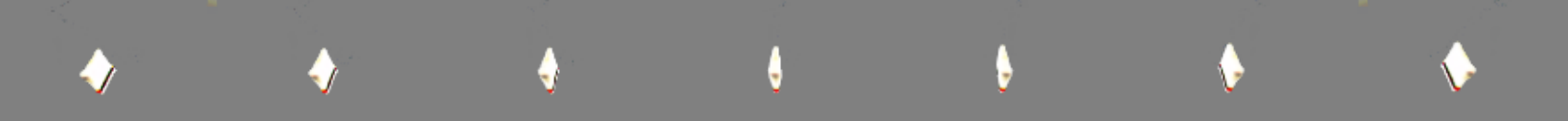}\\
        \raisebox{0.06\textwidth}{\rotatebox[origin=t]{90}{\scalebox{0.65}{w/o Token Embeddings}}}&
        \includegraphics[width=0.9\textwidth, height=0.13\textwidth]{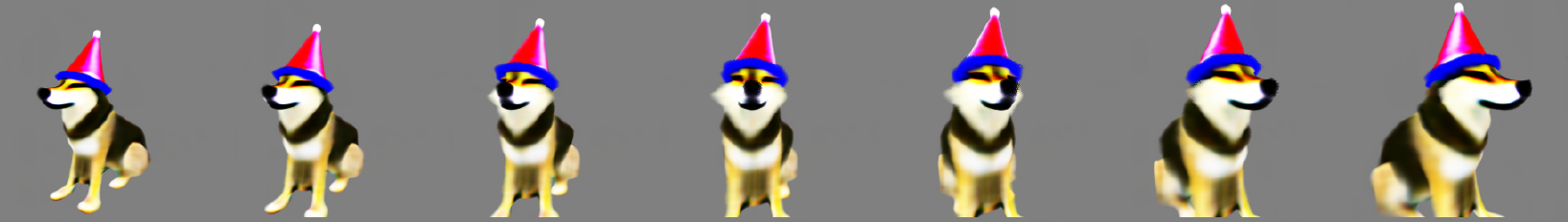}\\
        \raisebox{0.06\textwidth}{\rotatebox[origin=t]{90}{\scalebox{0.9}{Complete}}}&
        \includegraphics[width=0.9\textwidth, height=0.13\textwidth]{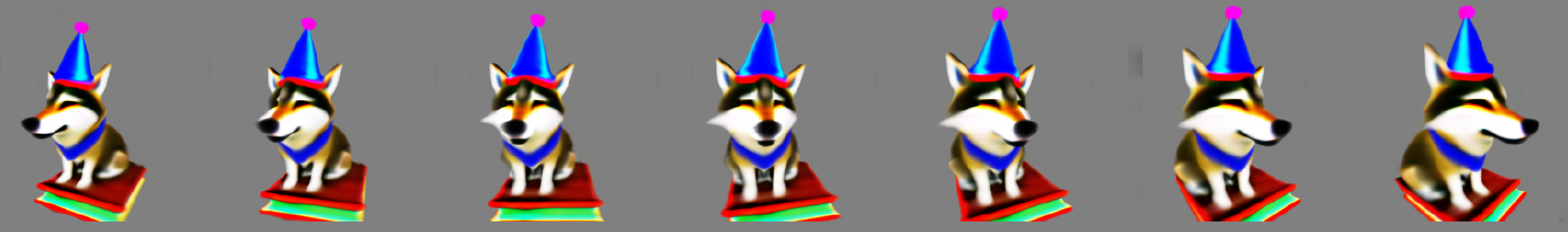}
    \end{tabular}
    }
    \caption{Effectiveness of the proposed style injection module.
    Top: without Gaussian noise; middle: without token embeddings; bottom: full model.
    The prompt is ``a wolf sitting on books and wearing a tie and wearing a party hat''. 
    The comparison shows that the incorporation of Gaussian noise prevents training failure, and the token embeddings contribute to better text-3D consistency.}
    \label{fig:style_mapping_module}
\end{figure*} 

\subsection{Comparison with ATT3D} 
\label{sec:comparison_att3d}     
In this section, we compare our method with a concurrent work ATT3D~\citep{lorraine2023att3d}.
Considering that the code of ATT3D is not publicly available, we retrain and evaluate our model on the dataset of ATT3D composed of 2,400 texts for a better comparison.
Although ATT3D also aims for fast text-to-3D generation similar to this work,
it requires per-prompt fine-tuning after each network inference, leading to lower efficiency.
Moreover, with the techniques proposed in \sec~\ref{sec:method}, our network is more effective than the model of ATT3D, which significantly improves the generation quality.

\bfsection{Qualitative Comparison}
We compare our generated objects with those of ATT3D in \fig~\ref{fig:qualitative_comparison_att3d}. 
The animals generated by ATT3D are blurry and lack essential details, and some described items in the prompts are not distinguishable. 
Moreover, the interactions between animals and their holding items could be disordered, for example, the erroneous orientations of the guitar and katana. 
In contrast, our \name{} consistently produces sharp renderings and ensures plausible interactions between entities.

\bfsection{Computation Costs}
We compare our computation cost with ATT3D by reporting Views-PP \vs CLIP-RP in~\fig~\ref{fig:comparison_retrieval_rp}. 
As the code of ATT3D is not accessible to the public, we directly extract the reported numbers from the original paper~\citep{lorraine2023att3d}. 
Notably, our \name{} requires fewer renderings per prompt during optimization, yet achieves superior text-3D alignment, underscoring its efficiency and effectiveness in high-quality text-to-3D generation.

\bfsection{User Preference Study}
We also conduct user studies to compare ATT3D and our approach based on user preferences.  
We show users videos rendered from multiple views of objects generated by two methods for the same text prompt. 
We ask them to select the result that has better quality in realism and text-3D consistency. 
To enhance the reliability of our study and mitigate variance, each prompt is evaluated by at least $3$ different users. 
As shown in \tab~\ref{tab:user_study_att3d}, the results of our \name{} are clearly preferred over those of the ATT3D in terms of both standards, highlighting the effectiveness of the proposed algorithm.

\begin{table}
\setlength{\tabcolsep}{1.5pt}
\centering
\caption{User preference study for comparing our results against those of ATT3D in terms of image realism and text-3D consistency.}
{
\begin{tabular}{ccccccc} 
\toprule
Method & ATT3D & \name{} 
 \\ 
\midrule
Realism &8.68\% &91.32\% \\
Consistency &13.23\% &86.77\% \\
\bottomrule
\end{tabular}}
\label{tab:user_study_att3d}
\end{table}

\subsection{Ablation Study}
We conduct extensive ablation studies on the Animals set to investigate the effectiveness of the key components in our network, including the style injection module, token-to-plane transformation, scaled-sigmoid function, and the adaptive Perp-Neg algorithm. 

\begin{figure}
    \centering
    \setlength{\tabcolsep}{1pt}
    {\scriptsize
    \begin{tabular}{c c c c}
        \raisebox{0.1\columnwidth}{\rotatebox[origin=t]{90}{\scalebox{0.9}{w/o}}}& 
        \includegraphics[width=0.3\columnwidth, height=0.23\columnwidth]{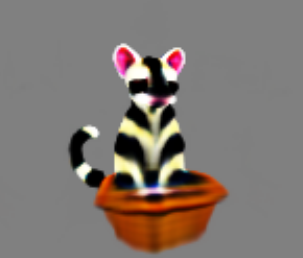}&
        \includegraphics[width=0.3\columnwidth, height=0.23\columnwidth]{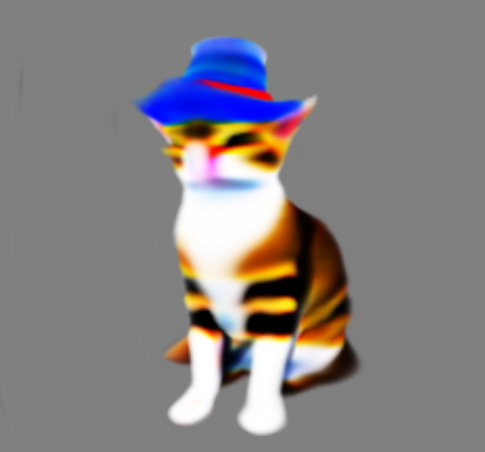}&
        \includegraphics[width=0.3\columnwidth, height=0.23\columnwidth]{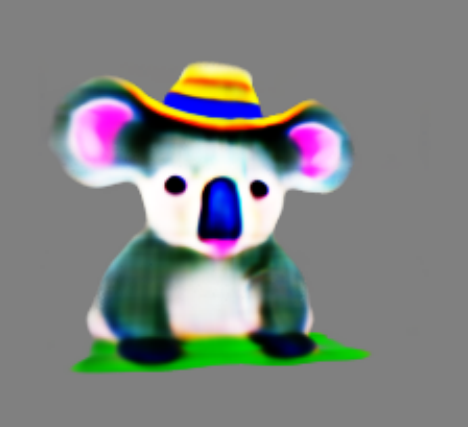}\\
        \raisebox{0.1\columnwidth}{\rotatebox[origin=t]{90}{\scalebox{0.9}{with}}}&
        \includegraphics[width=0.3\columnwidth, height=0.23\columnwidth]{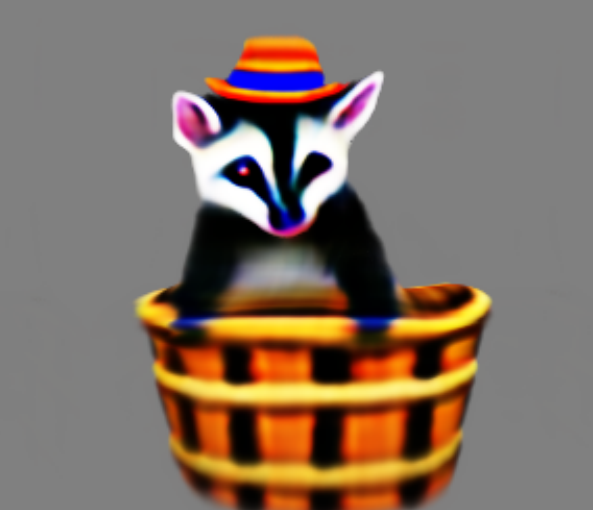}&
        \includegraphics[width=0.3\columnwidth, height=0.23\columnwidth]{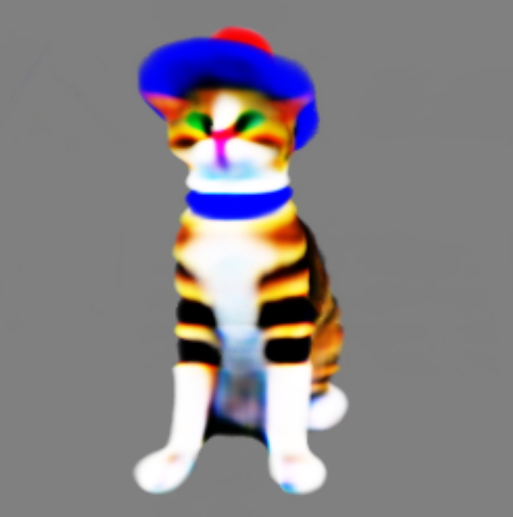}&
        \includegraphics[width=0.3\columnwidth, height=0.23\columnwidth]{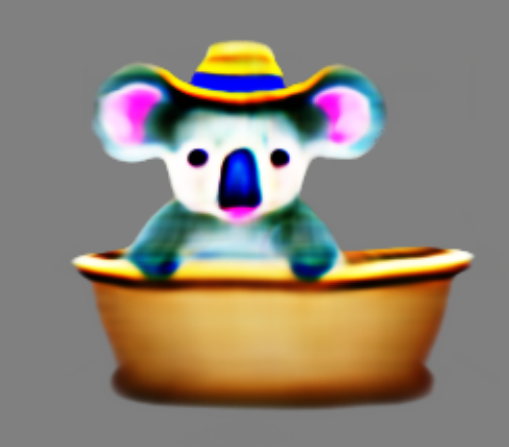}
    \end{tabular}
    \begin{tabular*}{0.9\columnwidth}{P{0.3\columnwidth}P{0.3\columnwidth}P{0.3\columnwidth}@{}}
    \centering
    ``a civet sitting in a basket and wearing a straw hat''&
    ``a cat wearing a tie and wearing a sombrero'' & ``a koala sitting in a bathtub and wearing a sombrero''
    \end{tabular*}
    }
    
    \caption{Effectiveness of the proposed token-to-plane transformation.
    Top: without token-to-plane transformation; bottom: full model.
    }
    \label{fig:ablation_token_to_plane}
\end{figure} 

\begin{figure*}
    \centering
    \setlength{\tabcolsep}{1pt}
    {\scriptsize
    \begin{tabular}{c c c c c c c c}
        \raisebox{0.06\textwidth}{\rotatebox[origin=t]{90}{\scalebox{0.9}{Sigmoid}}}& 
        \includegraphics[width=0.13\linewidth, height=0.13\linewidth]{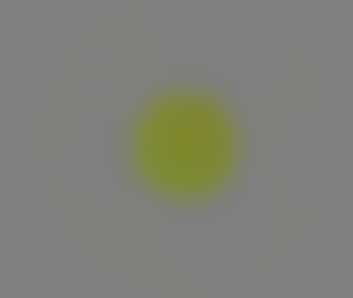}&
        \includegraphics[width=0.13\linewidth, height=0.13\linewidth]{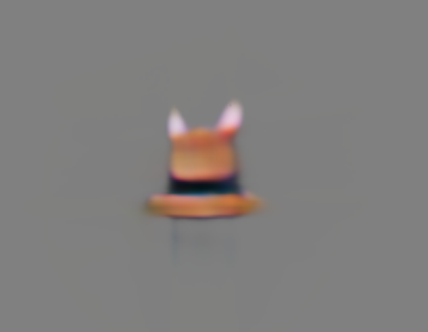}&
        \includegraphics[width=0.13\linewidth, height=0.13\linewidth]{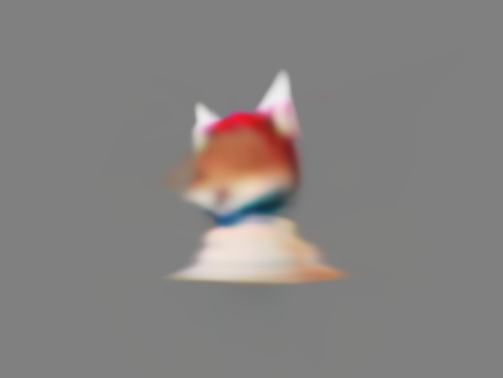}&
        \includegraphics[width=0.13\linewidth, height=0.13\linewidth]{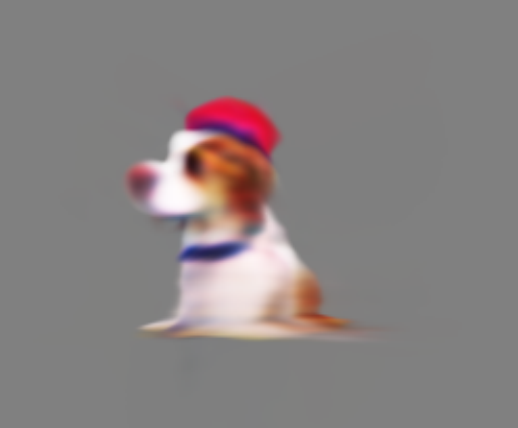}&
        \includegraphics[width=0.13\linewidth, height=0.13\linewidth]{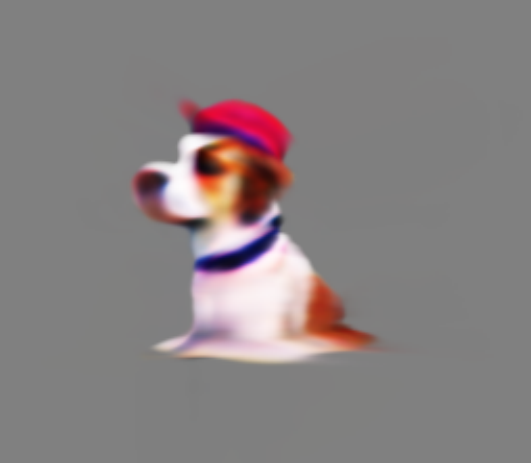}&
        \includegraphics[width=0.13\linewidth, height=0.13\linewidth]{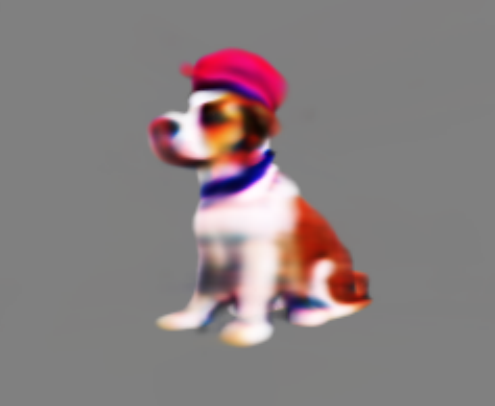}&
        \includegraphics[width=0.13\linewidth, height=0.13\linewidth]{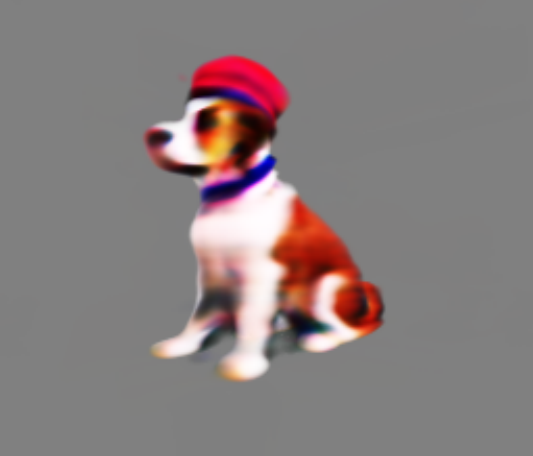}
        \\
        & 0 &3500 &7000 &10500 &14000 &17500 &21000
        \\
        \raisebox{0.06\textwidth}{\rotatebox[origin=t]{90}{\scalebox{0.9}{Scaled-Sigmoid}}}& 
        \includegraphics[width=0.13\linewidth, height=0.13\linewidth]{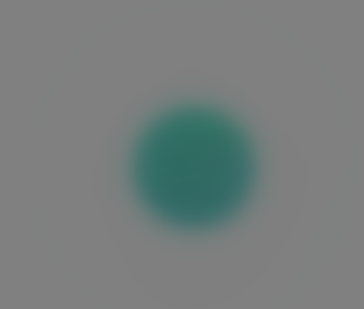}&
        \includegraphics[width=0.13\linewidth, height=0.13\linewidth]{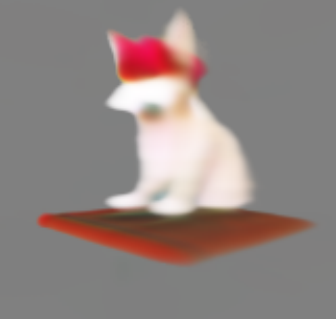}&
        \includegraphics[width=0.13\linewidth, height=0.13\linewidth]{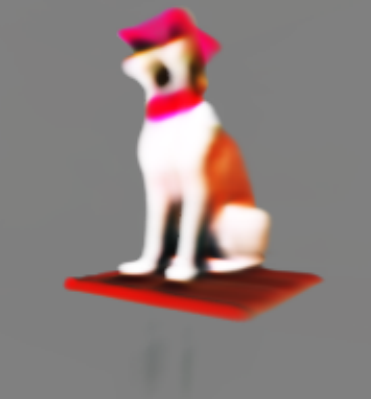}&
        \includegraphics[width=0.13\linewidth, height=0.13\linewidth]{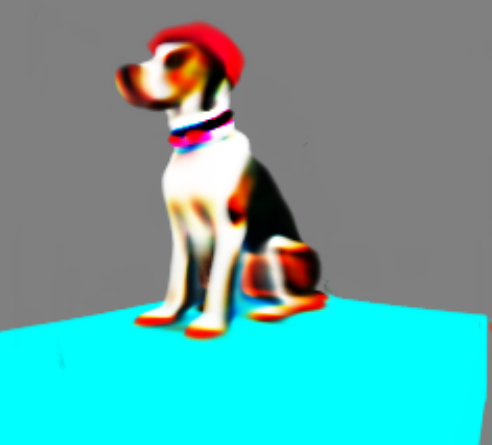}&
        \includegraphics[width=0.13\linewidth, height=0.13\linewidth]{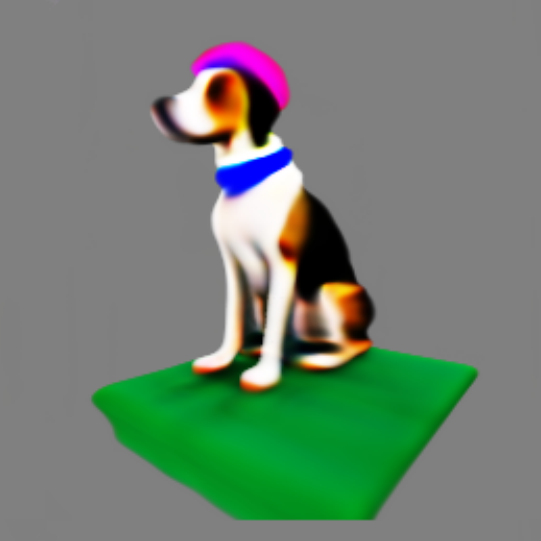}&
        \includegraphics[width=0.13\linewidth, height=0.13\linewidth]{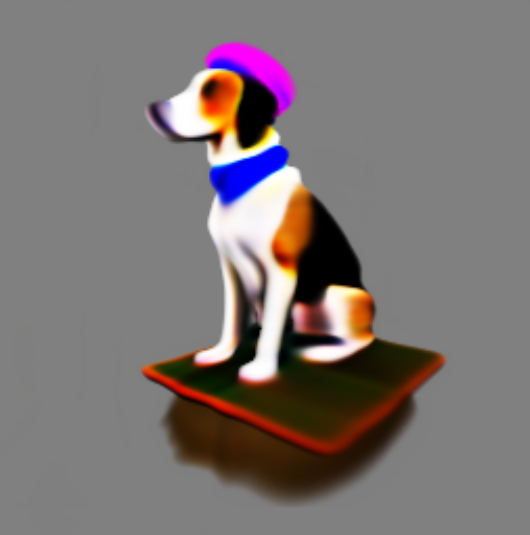}&
        \includegraphics[width=0.13\linewidth, height=0.13\linewidth]{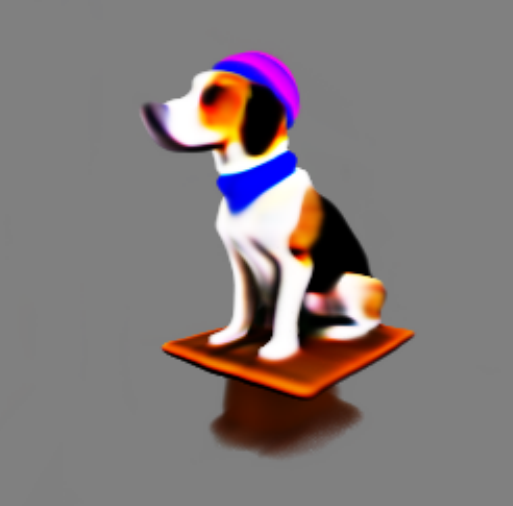}
        \\
        & 0 &350 &700 &1050 &1400 &1750 &2100
        \\
    \end{tabular}
    }
    \caption{
    Evolution process of the generated objects with the conventional sigmoid activation (top) and the scaled-sigmoid activation (bottom) for albedo prediction. The number below each image is the corresponding training iteration in terms of Views-PP. The prompt is ``a dog sitting on a table and wearing a tie and wearing a beanie''. The scaled-sigmoid function significantly accelerates the training convergence and enables better capability in text-3D alignment.
    }
    \label{fig:comparison_scaled_sigmoid}
\end{figure*} 

\begin{figure}
    \centering
    \setlength{\tabcolsep}{1pt}
    {\scriptsize
    \begin{tabular}{c c c c}
        \raisebox{0.1\columnwidth}{\rotatebox[origin=t]{90}{\scalebox{0.9}{Origin}}}& 
        \includegraphics[width=0.3\columnwidth, height=0.23\columnwidth]{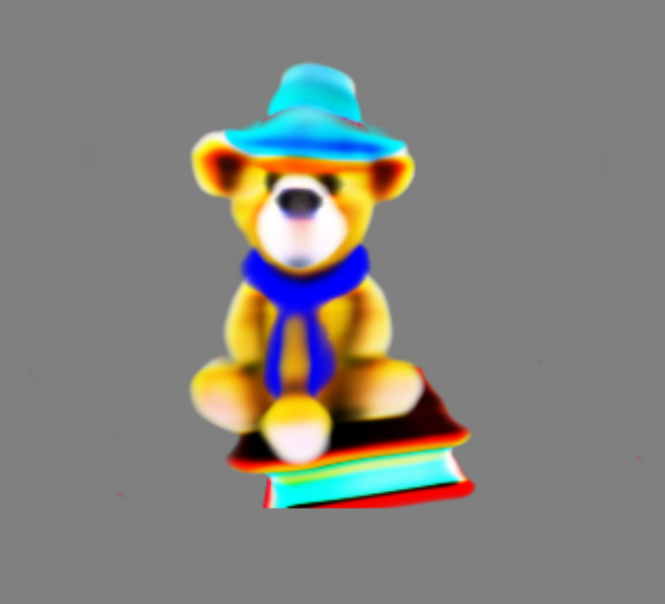}&
        \includegraphics[width=0.3\columnwidth, height=0.23\columnwidth]{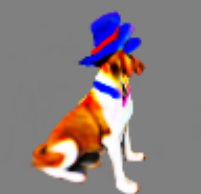}&
        \includegraphics[width=0.3\columnwidth, height=0.23\columnwidth]{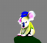}\\
        \raisebox{0.1\columnwidth}{\rotatebox[origin=t]{90}{\scalebox{0.9}{Adaptive}}}&
        \includegraphics[width=0.3\columnwidth, height=0.23\columnwidth]{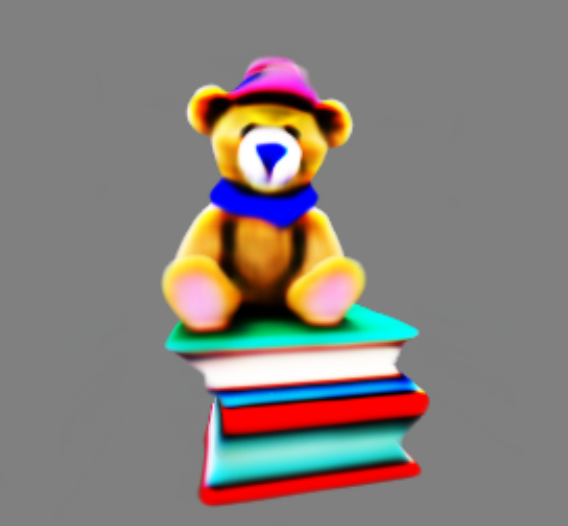}&
        \includegraphics[width=0.3\columnwidth, height=0.23\columnwidth]{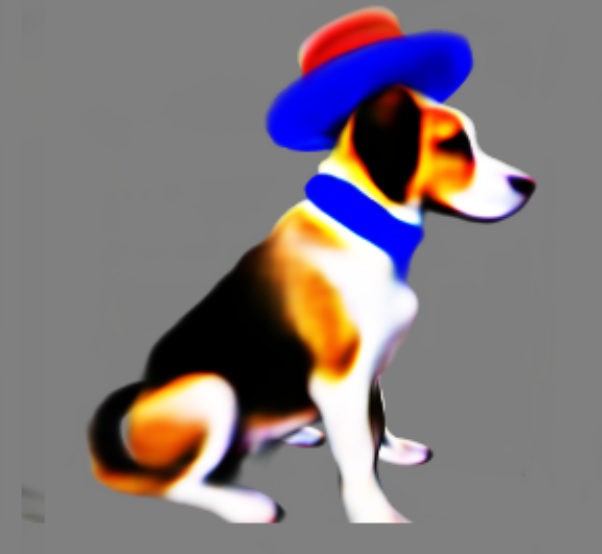}&
        \includegraphics[width=0.3\columnwidth, height=0.23\columnwidth]{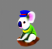}
    \end{tabular}
    }
    \caption{Comparison of the generated objects by using the original Perp-Neg algorithm (top) and our adaptive variant (bottom). In the top row, the teddy bear has three feet, while the dog and koala already display severe flat faces. This suggests that the original algorithm can not simultaneously address Janus problem for multiple objects. In contrast, our adaptive algorithm adjusts the concept negation scales of various objects during the generation process and significantly overcomes the above challenge.  
    }
    \label{fig:adaptive_perp_neg}
\end{figure}

\bfsection{Effectiveness of Style Injection}
To evaluate the importance of our style injection module, we evaluate two ablated versions: one without Gaussian noise and the other excluding token embeddings, whose results are shown in \fig~\ref{fig:style_mapping_module}.

Comparing the top and bottom rows, it is evident that adding Gaussian noise is instrumental in circumventing the training failure issue, which prevents generating null or meaningless outputs. 
The middle and bottom rows underscore the significance of token embeddings in the style injection module, which better inject textual conditions into the triplane generation process, leading to improved text-3D consistency under weak supervision signals.

\bfsection{Effectiveness of Token-to-Plane Transformation}
To understand the effect of our token-to-plane transformation, we replace it with a learnable constant to feed into the decoder. 
As illustrated in \fig~\ref{fig:ablation_token_to_plane} (top), objects generated without token-to-plane transformation cannot be well controlled by the condition text prompts. 
For example, the expected hat is absent in the first column, and the bathtub is missing in the third column. In contrast, our full approach (bottom row) ensures all described items in the prompts are accurately generated and interact in a plausible manner. 

\bfsection{Effectiveness of Scaled-Sigmoid}
In \sec~\ref{sec:scaled-sigmoid}, we introduce a scaled-sigmoid function for NeRF albedo prediction to accelerate the training process.
To verify this design, we provide a comparison between the scaled-sigmoid and the conventional sigmoid function. 
Their corresponding generation processes for the same prompt are shown in \fig~\ref{fig:comparison_scaled_sigmoid}. 
The numbers below images are the training iterations in terms of Views-PP. 
It is evident that the 3D generation driven by the standard sigmoid activation evolves much slower than that of the scaled-sigmoid. 
Notably, even with ten times training iterations of the scaled-sigmoid, the network with the conventional sigmoid function still does not achieve satisfactory convergence.
The evolution process of the generated 3D object also shows that the scaled-sigmoid enables a more significant capability of our model in adjusting the output to match the text description.

\bfsection{Effectiveness of Adaptive Perp-Neg}
As detailed in \sec~\ref{sec:adaptive perp neg}, we propose the adaptive Perp-Neg algorithm to address the Janus problem commonly encountered in tex-to-3D.
To understand the benefits of this algorithm, we provide a comparison against the original Perp-Neg algorithm where $w_{\text{neg}}$ remains fixed throughout training. 
As shown in \fig~\ref{fig:adaptive_perp_neg} (top), it is not feasible to find a universally optimum value for all objects: the Janus problem is not fully resolved in the first column (three feet for the teddy bear), which indicates the $w_{\text{neg}}$ is too small; but for the other two columns, the Janus problem has been overly punished, evidenced by flat faces, which indicates the $w_{\text{neg}}$ is too large.
In \fig~\ref{fig:adaptive_perp_neg} (bottom), as our adaptive Perp-Neg can dynamically adjust $w_{\text{neg}}$ for each object according to the severity of the Janus problem, it effectively overcomes the above challenge and produces higher-quality results.

\section{Discussions}
\bfsection{Relationship with Baseline Methods}
\textcolor{black}{Similar to existing SDS-based text-to-3D models~\citep{poole2022dreamfusion,wang2023prolificdreamer}, we aim to generate one specific object for one given text prompt.
The difference is that we amortize optimization over text prompts by training on many prompts simultaneously with a unified model, rather than optimizing separately for each prompt. This allows us to share computation across a prompt set, resulting in faster training compared to per-prompt optimization as demonstrated in Figure~\ref{fig:comparison_retrieval_rp}.}

\begin{figure*}[h]
    \centering
    \setlength{\tabcolsep}{1pt}
    {\scriptsize
    \begin{tabular}{c c c c c c c}
        \raisebox{0.06\textwidth}{\rotatebox[origin=t]{90}{\scalebox{0.9}{}}}& 
        \includegraphics[width=0.13\linewidth, height=0.13\linewidth]{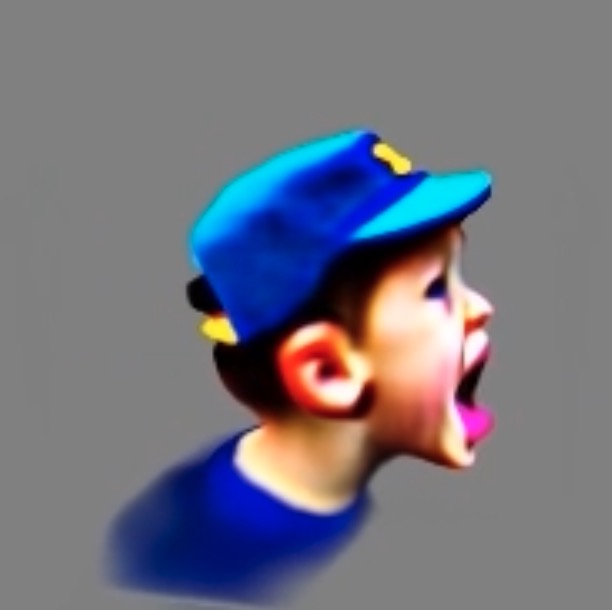}&
        \includegraphics[width=0.13\linewidth, height=0.13\linewidth]{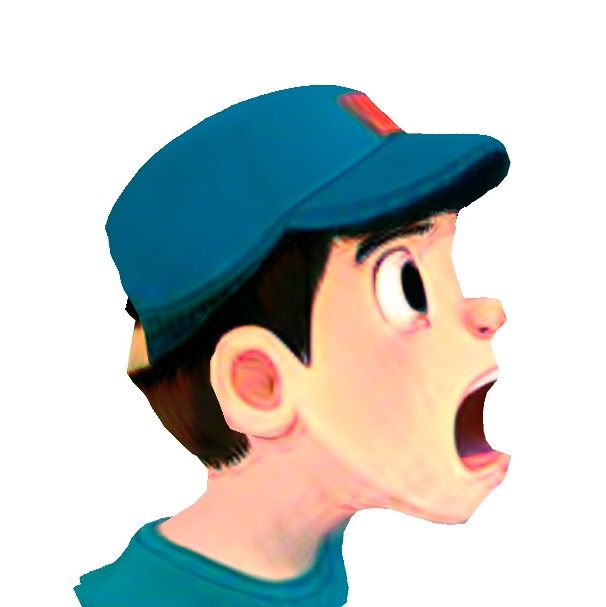}&
        \includegraphics[width=0.13\linewidth, height=0.13\linewidth]{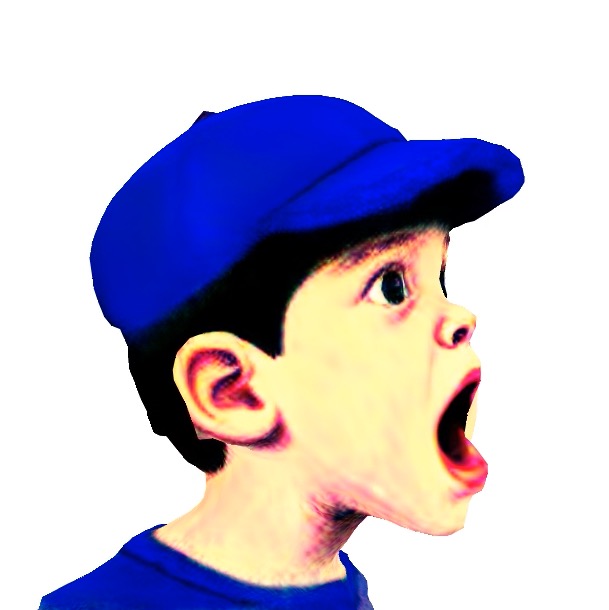}&
        \includegraphics[width=0.13\linewidth, height=0.13\linewidth]{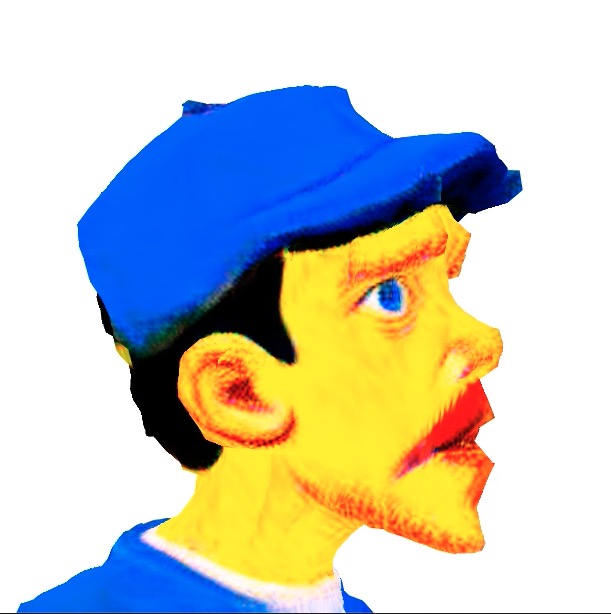}&
        \includegraphics[width=0.13\linewidth, height=0.13\linewidth]{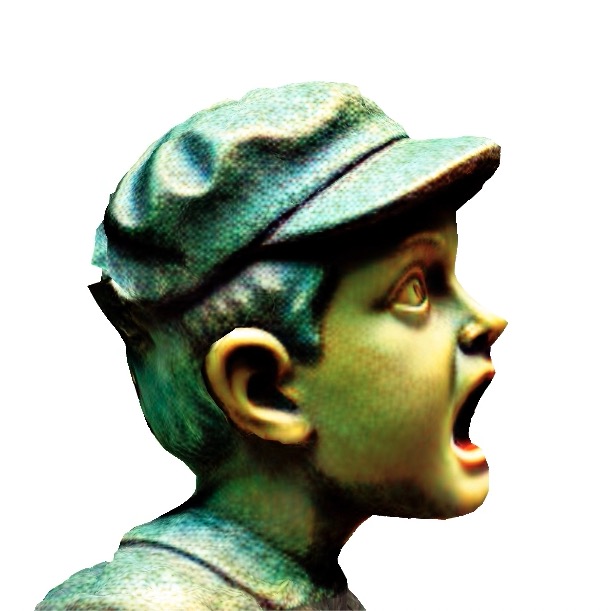}
    \end{tabular}
    \begin{tabular*}{0.9\linewidth}{P{0.12\linewidth}P{0.13\linewidth}P{0.13\linewidth}P{0.13\linewidth}P{0.13\linewidth}P{0.13\linewidth}P{0.13\linewidth}@{}}
    \centering  
    & Origin & ``cartoon style''  &``Monet style'' &``Van Gogh style'' &``Michelangelo style statue of''
    \end{tabular*}
    }
    \caption{\textcolor{black}{It is straightforward to diversify our generated objects by adding more detailed descriptions in an additional post-processing step, such as DMTet~\citep{shen2021deep}. The main prompt is ``a boy wearing a peaked cap is shouting''.}
    }
    \label{fig:stylization}
\end{figure*}

\bfsection{Diversity}
{\color{black}Our model is designed to produce a single, consistent output for a given text prompt, which is a different objective from general 2D generative models that aim to model the entire data distribution and generate different outputs for the same text prompt. 
This task setting is similar to that described in~\citep{lorraine2023att3d}.

While our main goal is not to generate multiple outputs for a single prompt,
it is worth noting that it is straightforward to combine the proposed Instant3D with the post-processing step to achieve more diverse results.
As shown in Figure~\ref{fig:stylization}, we can easily generate text-to-3D results with different styles for the same prompt by including more detailed text descriptions in the refinement stage.
Similar to ProlificDreamer~\citep{wang2023prolificdreamer}, we use DMTet~\citep{shen2021deep} for refinement.}

\begin{figure*}
    \centering
    \setlength{\tabcolsep}{1pt}
    {\scriptsize
    \begin{tabular}{c c c c c c c}
        \raisebox{0.06\textwidth}{\rotatebox[origin=t]{90}{\scalebox{0.9}{}}}& 
        \includegraphics[width=0.13\linewidth, height=0.13\linewidth]{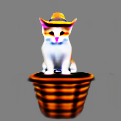}&
        \includegraphics[width=0.13\linewidth, height=0.13\linewidth]{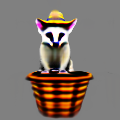}&
        \includegraphics[width=0.13\linewidth, height=0.13\linewidth]{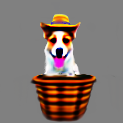}& 
        \includegraphics[width=0.13\linewidth, height=0.13\linewidth]{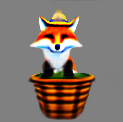}&
        \includegraphics[width=0.13\linewidth, height=0.13\linewidth]{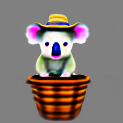}&
        \includegraphics[width=0.13\linewidth, height=0.13\linewidth]{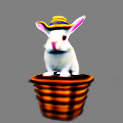}
    \end{tabular}
    \begin{tabular*}{0.9\linewidth}{P{0.05\linewidth}P{0.13\linewidth}P{0.13\linewidth}P{0.13\linewidth}P{0.13\linewidth}P{0.13\linewidth}P{0.13\linewidth}@{}}
    \centering  
    & ``cat'' & ``civet'' &``dog''& ``fox'' & ``koala'' &``rabbit''
    \end{tabular*}
    }
    
    \caption{\textcolor{black}{Fine-grained consistency for changing the animal species. The corresponding text is ``a \textit{species} sitting in a basket and wearing a sombrero''. The remaining items other than species are well maintained. 
    }}
    \label{fig:fine_grained_consistency}
\end{figure*} 

\bfsection{Fine-Grained Consistency}
\textcolor{black}{
Fine-grained consistency between text and 3D is often of interest to the research community and can be useful for applications requiring precise control over the generated 3D models. 
As demonstrated in Figure~\ref{fig:fine_grained_consistency}, our model can effectively change the animal species in the text prompt while maintaining other items in the scene. 
This capability reflects our model's strong understanding of the individual components described in the text, allowing it to accurately modify specific aspects of the 3D object based on the given prompt.
}

\bfsection{Limitations}
\textcolor{black}{As an early exploration of fast text-to-3D generation, we verify the proposed \name{} on three compositional datasets and one open-world prompt set, \ie Daily Life. However, compared with the dataset scale of text-to-image synthesis, these prompt sets are still relatively small. This can be attributed to two factors. On one hand, it is difficult and time-consuming to collect a large dataset with coupled prompts. On the other hand, the computation burden of training a text-to-3D network is thought to be much heavier than that of a text-to-image model due to the higher dimension of 3D objects. These can be solved by more efficient 3D representation methods and powerful computation devices in the future.
}

\bfsection{Negative Societal Impacts}
\textcolor{black}{The fast text-to-3D generation approach proposed in this paper aims to infer a faithful 3D object for a testing prompt in 20ms. We note that it could be potentially applied to unsafe scenarios such as generating violent or sexual content through third-party fine-tuning like other existing methods. However, it actually inherits the data bias from the used text-to-image model, \ie Stable Diffusion \citep{rombach2022high}, whose training text-image pairs should be carefully filtered. Besides, there is also growing interest in focusing on such potential safety problems in the research community.}

\section{Conclusion}  
In this paper, we demonstrate the feasibility of fast text-to-3D generation with a one-pass feedforward network. To overcome the learning difficulties brought by weak supervision from SDS loss, we integrate three condition mechanisms, \ie cross-attention, style injection, and token-to-plane transformation, to better embed text prompts into a decoder network. 
We also present a scaled-sigmoid function for albedo activation, which accelerates the optimization convergence by more than ten times. 
In addition, we propose an adaptive Perp-Neg algorithm, which effectively tackles the Janus problem in our new paradigm without introducing extra computation costs. 
Extensive experiments on four prompt sets show that our \name{} is able to generate faithful 3D objects for novel prompts in less than one second. Moreover, our approach shows much better optimization efficiency. 
We expect this work to inspire further explorations in fast text-to-3D generation.

\section*{Acknowledgement}
Ming Li is funded by the ISEP-IDS PhD scholarship in NUS.
Pan Zhou is supported by the Singapore Ministry of Education (MOE) Academic Research Fund (AcRF) Tier 1 grant. 
Xiangyu Xu is supported by NSFC (62302385) and the computational resources provided by the HPC platform of Xi'an Jiaotong University.
{\bibliographystyle{spbasic}
\bibliography{references}
}

\appendix
\section{Network Architecture}
{Our decoder network is inspired by the UNet architecture of Stable Diffusion~\citep{rombach2022high} whose encoder network is removed herein. Overall, we set the \textit{base channel} as 80 instead of the original 320 and increase the \textit{layers per block} to 10, resulting in a deeper and more powerful decoder network. 
It takes the output of the token-to-plane transformation, with a resolution of $8\times 8$, as input. Its five stages gradually increase the feature resolution to $16\times 16$, $32\times 32$, $64\times 64$, $128\times 128$ and $256\times 256$, respectively. The corresponding \textit{channel multipliers} are 4, 4, 2, 1 and 1. Each of the first three stages consists of ten alternate attention and convolution blocks, while the last two stages are only composed of convolution blocks. One $\times 2$-upsampling layer is inserted after each stage to increase the feature resolution, finally yielding an output triplane with the shape of $3\times 32\times 256\times 256$.} 

\begin{figure*}[h]
\centering
\includegraphics[width=0.9\textwidth]{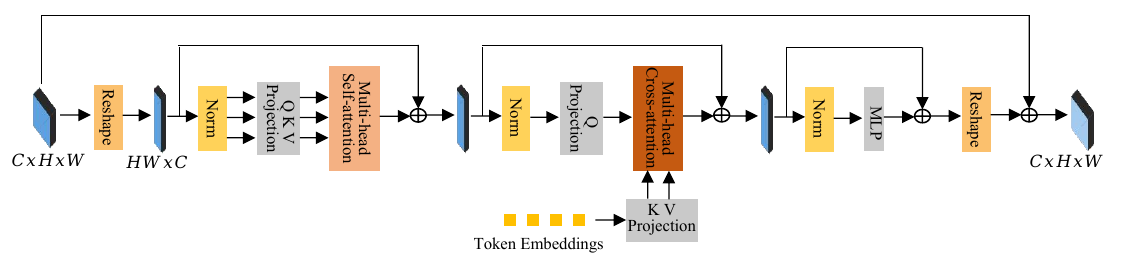}
\caption{Illustration of the attention block, which mainly contains a self-attention mechanism for inner-feature map information interaction and a cross-attention mechanism for text-3D information interchange.
}
\label{fig:attentional_block}
\end{figure*}

\bfsection{Attention Block}
The attention block is shown in \fig~\ref{fig:attentional_block}, which enables inner information interaction by multi-head self-attention and text-3D information interchange through multi-head cross-attention. 

\begin{figure*}[h]
\centering
\includegraphics[width=0.7\textwidth]{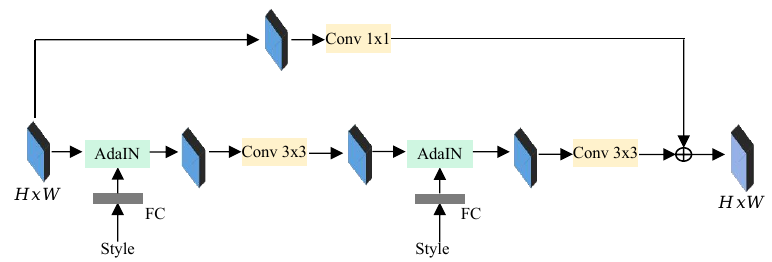}
\caption{Illustration of the convolution block, which employs AdaIN to incorporate the output of the style injection module into the triplane generation. 
}
\label{fig:stylized_conv_blocks}
\end{figure*}

\bfsection{Convolution Block}
The convolution block is illustrated in \fig~\ref{fig:stylized_conv_blocks}, which uses Adaptive Instance Normalization (AdaIN) to absorb the style information from the style injection module. 
Convolution layers are used in this block to introduce spatial inductive bias.

\section{More Visualization Results}
\begin{figure*}[h]
\setlength\abovecaptionskip{2mm}
\centering

\begin{subfigure}{0.99\linewidth}
    \centering
    \includegraphics[width=0.99\linewidth, height=0.14\linewidth]{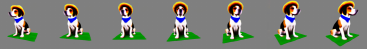}
    \vspace{-1.5mm}
    \caption{``a dog sitting on the lawn and wearing a tie and wearing a sombrero''}
\end{subfigure}
\begin{subfigure}{0.99\linewidth}
    \centering
    \includegraphics[width=0.99\linewidth, height=0.14\linewidth]{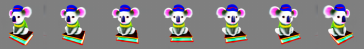}
    \vspace{-1.5mm}
    \caption{``a koala sitting on books and wearing a cape and wearing a baseball cap''}
\end{subfigure}
\begin{subfigure}{0.99\linewidth}
    \centering
    \includegraphics[width=0.99\linewidth, height=0.14\linewidth]{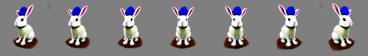}
    \vspace{-1.5mm}
    \caption{``a rabbit sitting on a stone and wearing a beret''}
\end{subfigure}
\begin{subfigure}{0.99\linewidth}
    \centering
    \includegraphics[width=0.99\linewidth, height=0.14\linewidth]{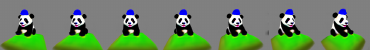}
    \vspace{-1.5mm}
    \caption{``a panda sitting on the lawn and wearing a beanie''}
\end{subfigure}
\begin{subfigure}{0.99\linewidth}
    \centering
    \includegraphics[width=0.99\linewidth, height=0.14\linewidth]{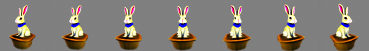}
    \vspace{-1.5mm}
    \caption{``a rabbit sitting in a basket and wearing a tie''}
\end{subfigure}

\begin{subfigure}{0.99\linewidth}
    \centering
    \includegraphics[width=0.99\linewidth, height=0.14\linewidth]{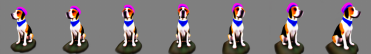}
    \vspace{-1.5mm}
    \caption{``a dog sitting on a stone and wearing a tie and wearing a beanie''}
\end{subfigure}
\caption{More results on the Animals set, which are inferred directly by our \name{} for novel prompts. We visualize continuous view images rendered around the front direction of each object to show the cross-view consistency.}
\label{fig:more_results_animals_set2}
\end{figure*}

\begin{figure*}[!t]
\setlength\abovecaptionskip{2mm}
\centering

\begin{subfigure}{0.99\linewidth}
    \centering
    \includegraphics[width=0.99\linewidth, height=0.12\linewidth]{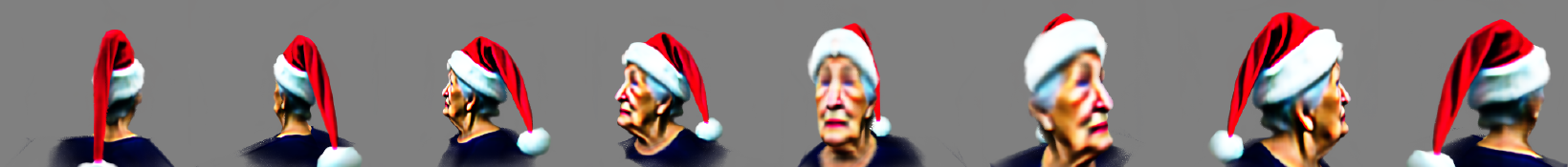}
\end{subfigure}
\begin{subfigure}{0.99\linewidth}
    \centering
    \caption{``an elderly woman wearing a Santa hat is looking ahead with a very serious expression''}
\end{subfigure}

\begin{subfigure}{0.99\linewidth}
    \centering
    \includegraphics[width=0.99\linewidth, height=0.12\linewidth]{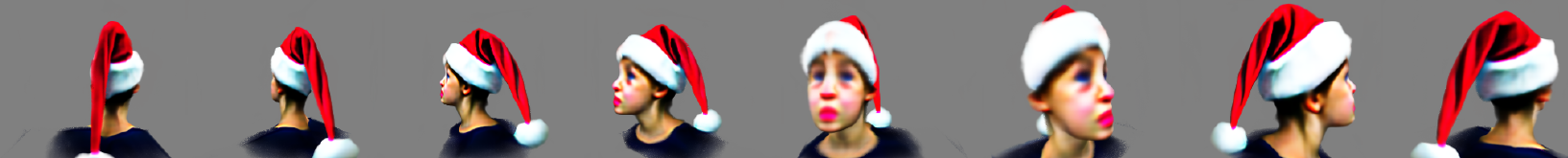}
\end{subfigure}
\begin{subfigure}{0.99\linewidth}
    \centering
    \caption{``a girl wearing a Santa hat is angry''}
\end{subfigure}

\begin{subfigure}{0.99\linewidth}
    \centering
    \includegraphics[width=0.99\linewidth, height=0.12\linewidth]{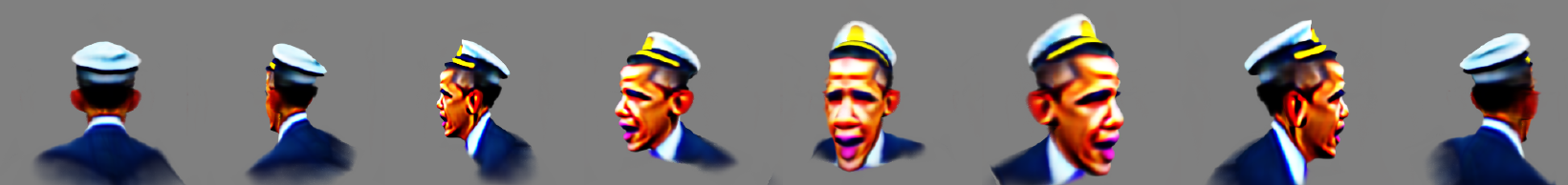}
\end{subfigure}
\begin{subfigure}{0.99\linewidth}
    \centering
    \caption{``Obama wearing a peaked cap is talking''}
\end{subfigure}

\begin{subfigure}{0.99\linewidth}
    \centering
    \includegraphics[width=0.99\linewidth, height=0.12\linewidth]{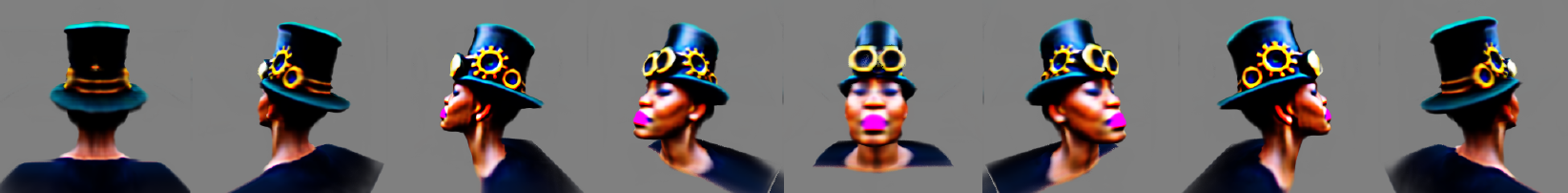}
\end{subfigure}
\begin{subfigure}{0.99\linewidth}
    \centering
    \caption{``a black woman wearing a steampunk hat is feeling sad''}
\end{subfigure}

\begin{subfigure}{0.99\linewidth}
    \centering
    \includegraphics[width=0.99\linewidth, height=0.12\linewidth]{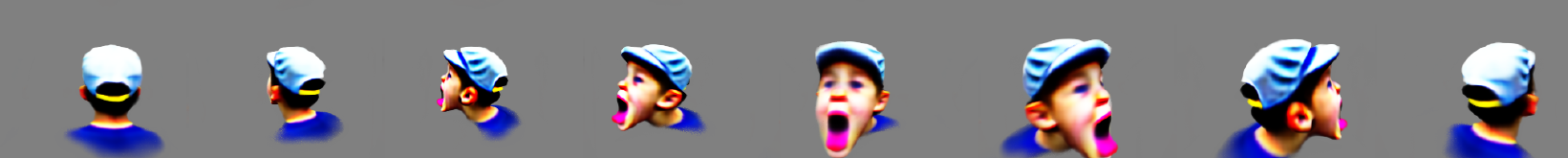}
\end{subfigure}
\begin{subfigure}{0.99\linewidth}
    \centering
    \caption{``a boy wearing a peaked cap is shouting''}
\end{subfigure}

\begin{subfigure}{0.99\linewidth}
    \centering
    \includegraphics[width=0.99\linewidth, height=0.12\linewidth]{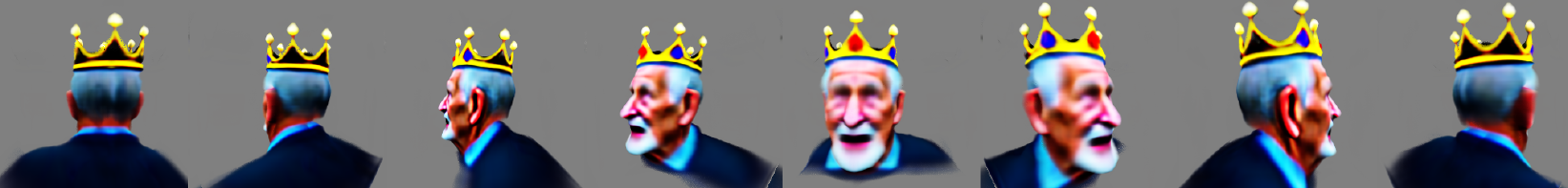}
\end{subfigure}
\begin{subfigure}{0.99\linewidth}
    \centering
    \caption{``an elderly man wearing a crown is opening mouth wide in shock''}
\end{subfigure}

\begin{subfigure}{0.99\linewidth}
    \centering
    \includegraphics[width=0.99\linewidth, height=0.12\linewidth]{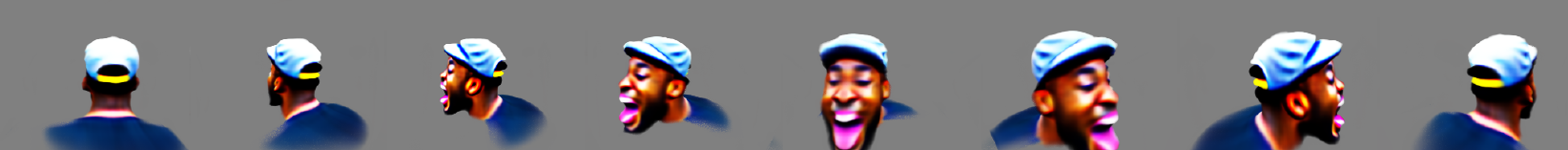}
\end{subfigure}
\begin{subfigure}{0.99\linewidth}
    \centering
    \caption{``a black man wearing a peaked cap is laughing''}
\end{subfigure}

\caption{More results on the Portraits set. The prompts describe various figures wearing different hats and expressing different emotions. We visualize continuous view images rendered from each object.}
\label{fig:portraits2}
\end{figure*}

\begin{figure*}[!t]
\setlength\abovecaptionskip{2mm}
\centering
    
\begin{subfigure}{0.5\linewidth}
    \centering
    \includegraphics[width=0.99\linewidth, height=0.2\linewidth]{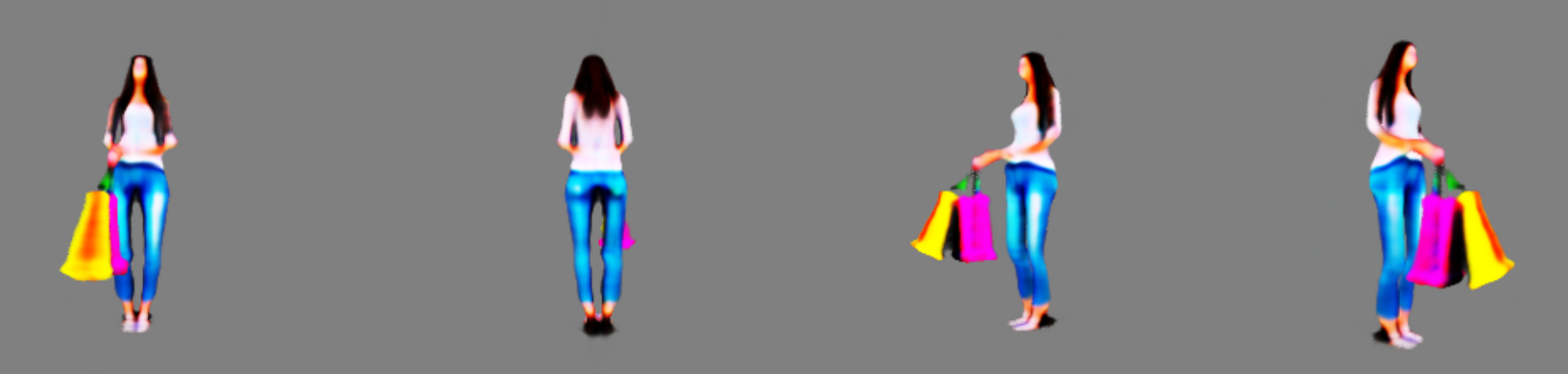}
    \vspace{-1.5mm}
\end{subfigure}
\hspace{-7pt}
\begin{subfigure}{0.5\linewidth}
    \centering
    \includegraphics[width=0.99\linewidth, height=0.2\linewidth]{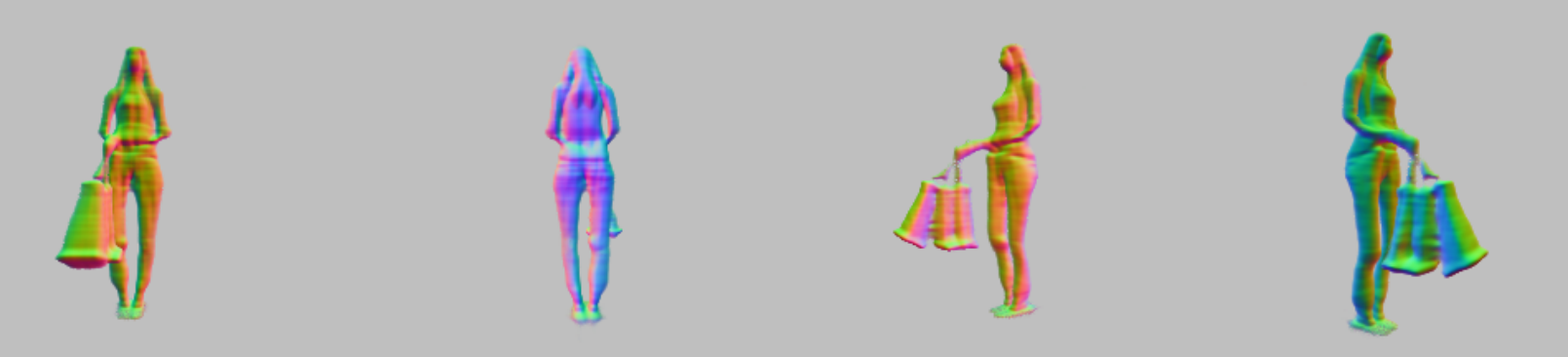}
    \vspace{-1.5mm}
\end{subfigure}
\begin{subfigure}{0.99\linewidth}
    \centering
    \caption{``a woman with long hair wearing jeans is shopping''}
\end{subfigure}
\begin{subfigure}{0.5\linewidth}
    \centering
    \includegraphics[width=0.99\linewidth, height=0.2\linewidth]{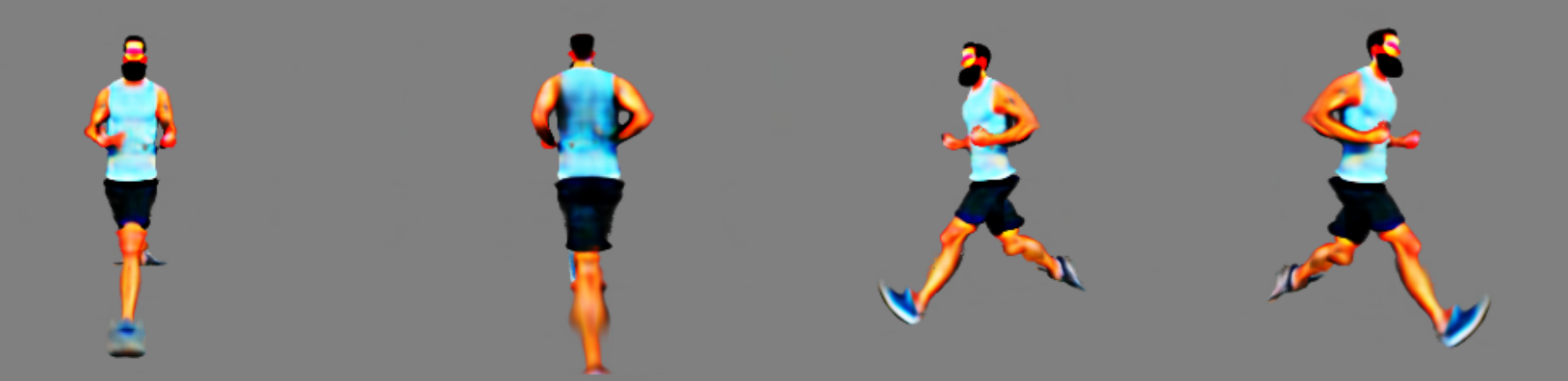}
    \vspace{-1.5mm}
\end{subfigure}
\hspace{-7pt}
\begin{subfigure}{0.5\linewidth}
    \centering
    \includegraphics[width=0.99\linewidth, height=0.2\linewidth]{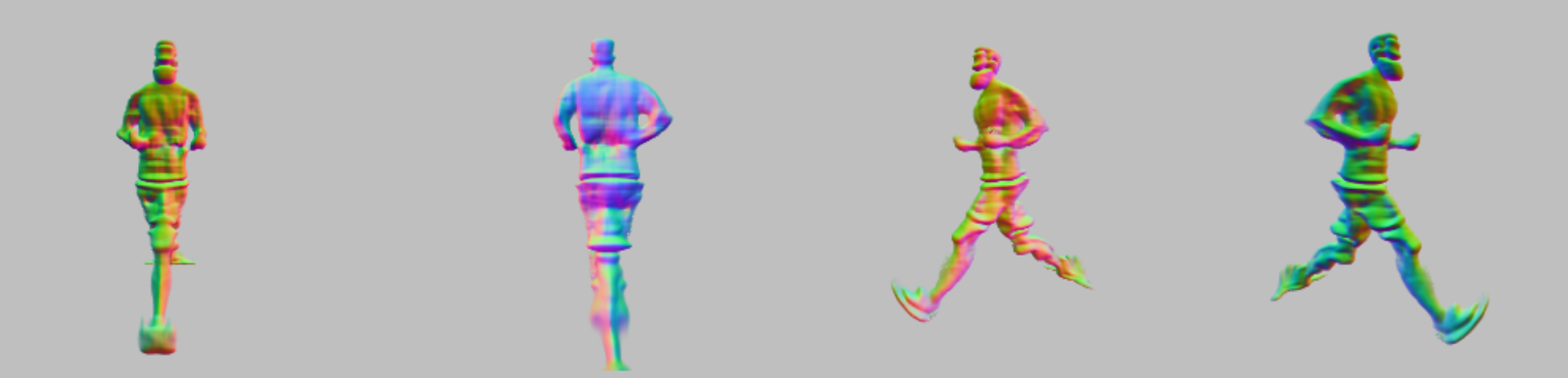}
    \vspace{-1.5mm}
\end{subfigure}
\begin{subfigure}{0.99\linewidth}
    \centering
    \caption{``a strong and tall man with a beard is jogging''}
\end{subfigure}

\begin{subfigure}{0.5\linewidth}
    \centering
    \includegraphics[width=0.99\linewidth, height=0.2\linewidth]{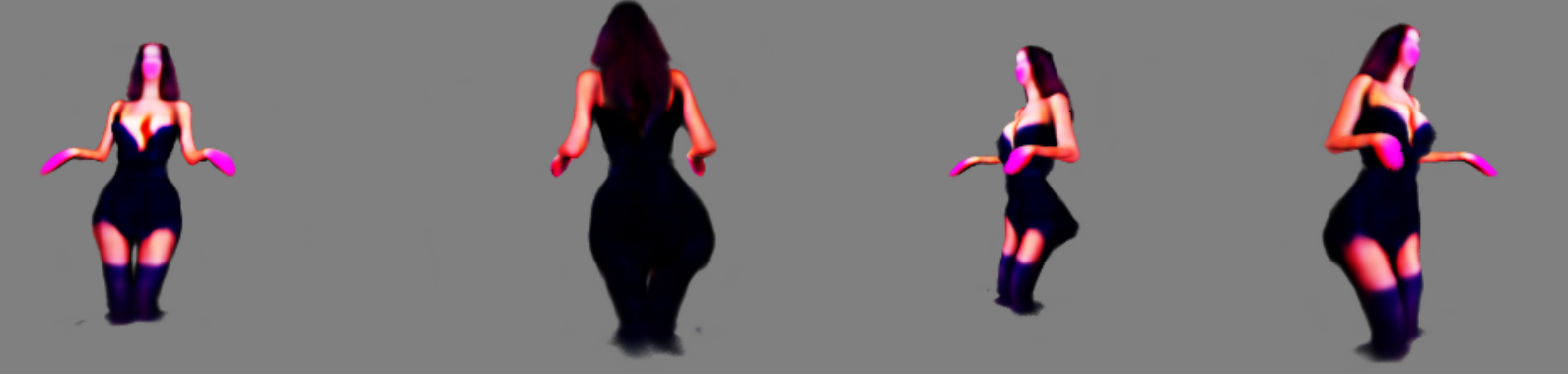}
    \vspace{-1.5mm}
\end{subfigure}
\hspace{-7pt}
\begin{subfigure}{0.5\linewidth}
    \centering
    \includegraphics[width=0.99\linewidth, height=0.2\linewidth]{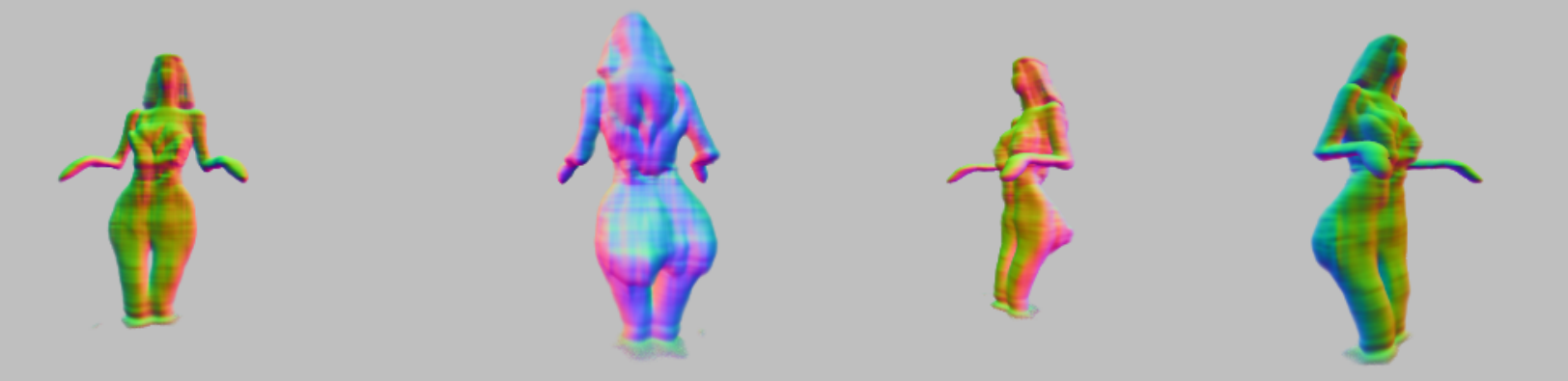}
    \vspace{-1.5mm}
\end{subfigure}
\begin{subfigure}{0.99\linewidth}
    \centering
    \caption{``a trendy woman is dancing to music''}
\end{subfigure}

\begin{subfigure}{0.5\linewidth}
    \centering
    \includegraphics[width=0.99\linewidth, height=0.2\linewidth]{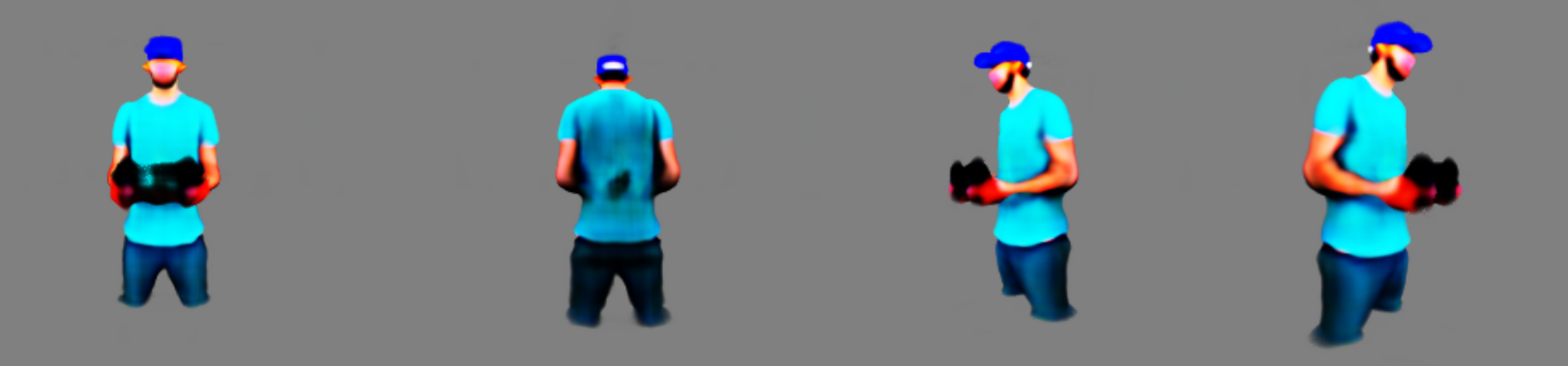}
    \vspace{-1.5mm}
\end{subfigure}
\hspace{-7pt}
\begin{subfigure}{0.5\linewidth}
    \centering
    \includegraphics[width=0.99\linewidth, height=0.2\linewidth]{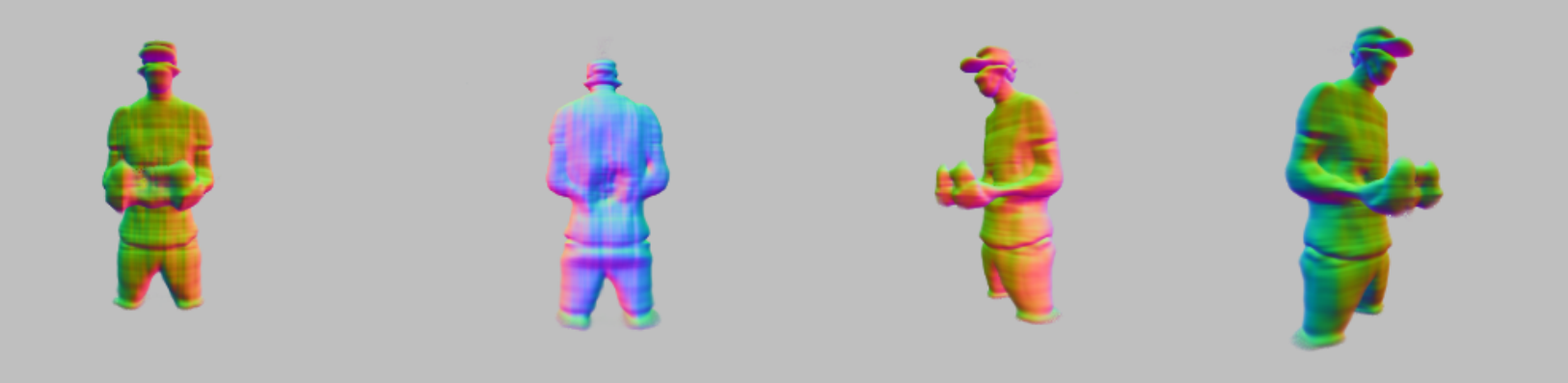}
    \vspace{-1.5mm}
\end{subfigure}
\begin{subfigure}{0.99\linewidth}
    \centering
    \caption{``a man wearing a baseball cap is playing video games''}
\end{subfigure}

\begin{subfigure}{0.5\linewidth}
    \centering
    \includegraphics[width=0.99\linewidth, height=0.2\linewidth]{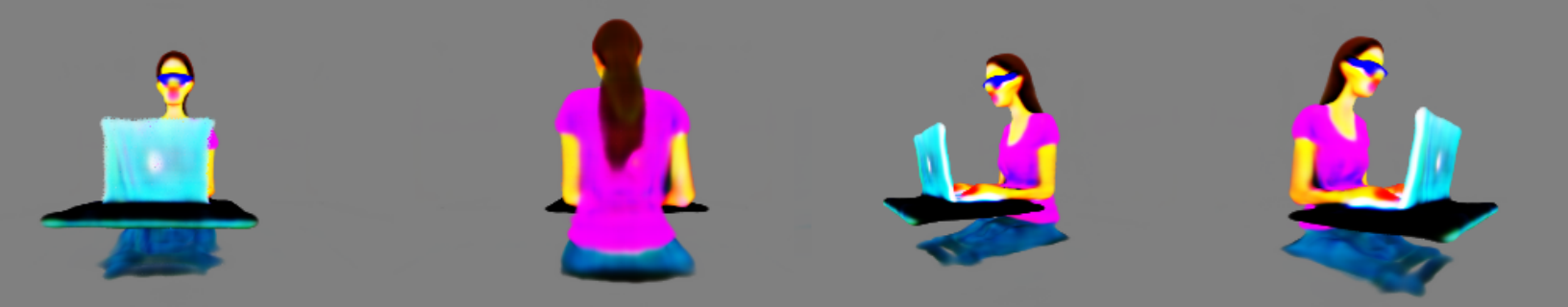}
    \vspace{-1.5mm}
\end{subfigure}
\hspace{-7pt}
\begin{subfigure}{0.5\linewidth}
    \centering
    \includegraphics[width=0.99\linewidth, height=0.2\linewidth]{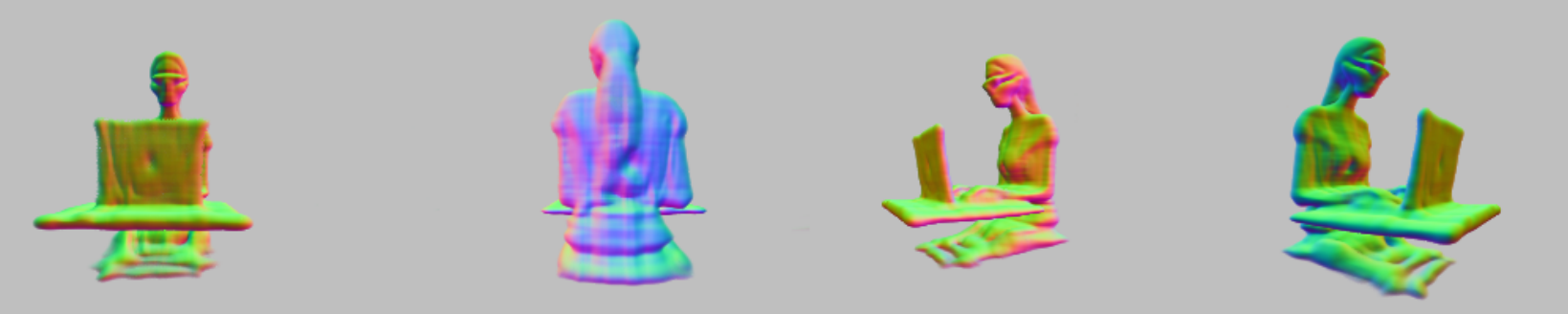}
    \vspace{-1.5mm}
\end{subfigure}
\begin{subfigure}{0.99\linewidth}
    \centering
    \caption{``a woman with glasses is working on a laptop''}
\end{subfigure}

\begin{subfigure}{0.5\linewidth}
    \centering
    \includegraphics[width=0.99\linewidth, height=0.2\linewidth]{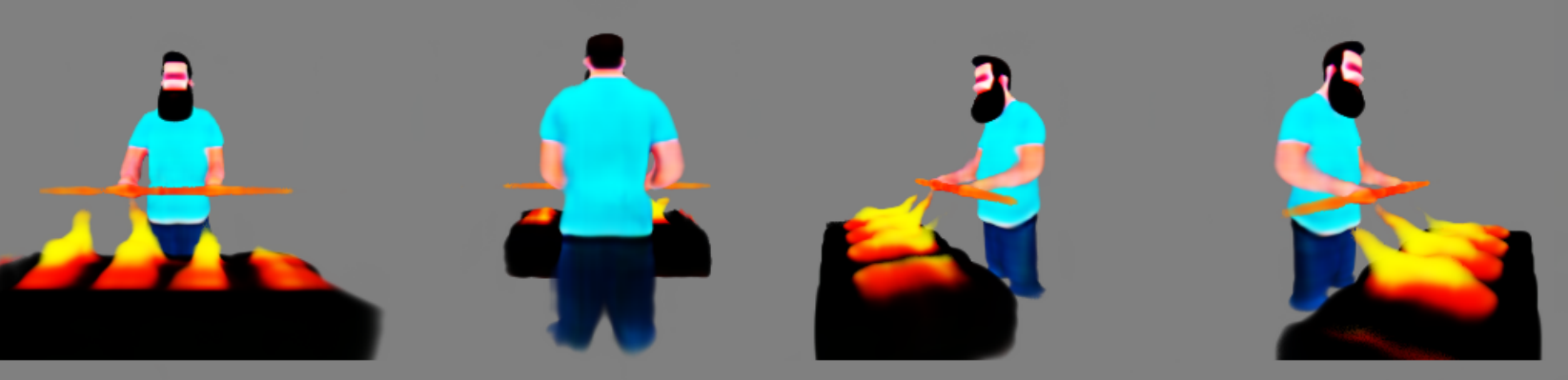}
    \vspace{-1.5mm}
\end{subfigure}
\hspace{-7pt}
\begin{subfigure}{0.5\linewidth}
    \centering
    \includegraphics[width=0.99\linewidth, height=0.2\linewidth]{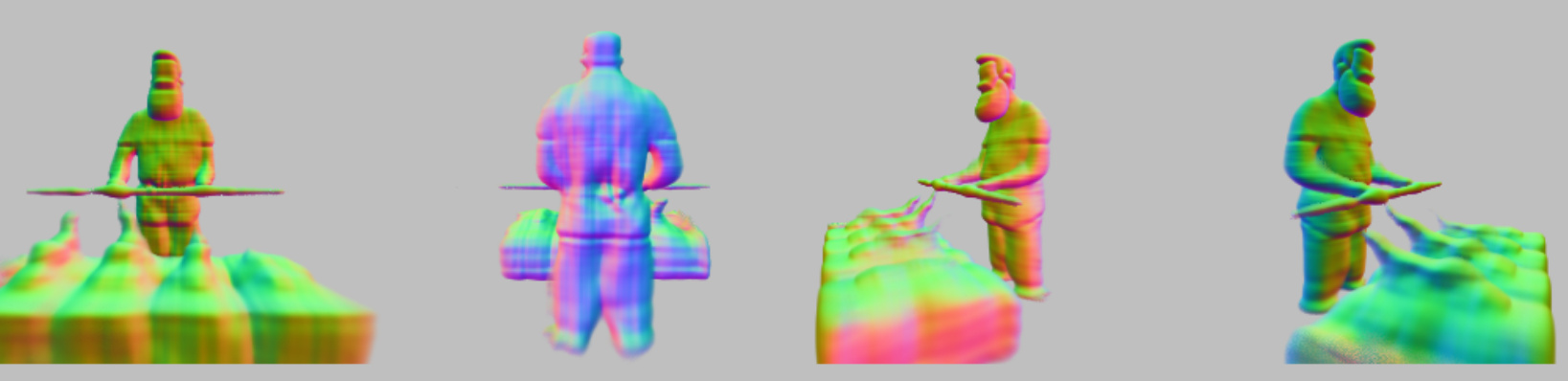}
    \vspace{-1.5mm}
\end{subfigure}
\begin{subfigure}{0.99\linewidth}
    \centering
    \caption{``a stocky man is barbecuing''}
\end{subfigure}
\caption{More generated objects for new testing prompts by our \name{} trained on the Daily Life set. For each prompt, four view images and their normal images are shown. Our framework is able to generate complex and diverse objects for new prompts with high-quality geometric details.}
\label{fig:daily_life2}
\end{figure*}

We show more objects generated by our \name{} trained on the Animals set (\fig~\ref{fig:more_results_animals_set2}), Portraits set (\fig~\ref{fig:portraits2}), and Daily Life set (\fig~\ref{fig:daily_life2}). The results demonstrate that our \name{} successfully generates faithful 3D objects for new prompts.

\end{document}